\PassOptionsToPackage{most}{tcolorbox} 
\PassOptionsToPackage{numbers,sort&compress}{natbib}
\documentclass[11pt]{article}

\usepackage[final]{acl}

\usepackage{amssymb}
\usepackage{booktabs}
\usepackage{tabularx}
\usepackage{makecell}
\usepackage{array}
\usepackage{tabularx}
\usepackage{booktabs}
\usepackage{graphicx}
\usepackage{pifont}
\usepackage[final]{microtype}

\newcolumntype{L}[1]{>{\raggedright\arraybackslash}p{#1}}
\newcolumntype{C}[1]{>{\centering\arraybackslash}p{#1}}
\newcolumntype{R}[1]{>{\raggedleft\arraybackslash}p{#1}}

\providecommand{\cmark}{\ding{51}}
\providecommand{\xmark}{\ding{55}}
\providecommand{\pmark}{$\sim$}

\providecommand{\methodcell}[2]{#1\,{\scriptsize\cite{#2}}}

\usepackage{tabularx,booktabs,array,makecell}

\usepackage{times}
\usepackage{latexsym}

\usepackage[utf8]{inputenc}
\usepackage{amsmath,amssymb}
\usepackage{pifont}
\usepackage{placeins} 
\usepackage{flafter}
\usepackage{booktabs}
\usepackage{multirow}
\usepackage{lipsum}
\usepackage{tabularx}
\usepackage{subcaption}
\usepackage{arydshln}
\usepackage{adjustbox} 
\usepackage{makecell}
\usepackage{wrapfig}
\usepackage{float}

\usepackage{siunitx}
\usepackage[T1]{fontenc}

\usepackage{inconsolata}
\usepackage{dblfloatfix}
\usepackage{graphicx}
\usepackage{enumitem}
\usepackage{cuted}
\usepackage{capt-of} 
\usepackage{graphicx}
\usepackage{dblfloatfix} 
\usepackage{caption}  
\usepackage{fontawesome5} 
\usepackage{hyperref}    
\usepackage{xcolor}       

\makeatletter
\renewcommand\section{\@startsection{section}{1}{\z@}{-2.2ex plus
    -0.5ex minus -.2ex}{1.8ex plus .4ex minus .2ex}{\Large\bfseries\raggedright}}
\renewcommand\subsection{\@startsection{subsection}{2}{\z@}{-2.0ex plus
    -0.5ex minus -.2ex}{1.4ex plus .3ex minus .2ex}{\large\bfseries\raggedright}}
\makeatother

\newcommand{\corrauthormark}{\ensuremath{\ddagger}}

\title{Volumetric Radiology AI in the Era of Multimodal Large Language Models}

\author{
\textbf{Zanting Ye\textsuperscript{1*}},
\textbf{Shengyuan Liu\textsuperscript{2*\textdagger}},
\textbf{Xin Liu\textsuperscript{1}},
\textbf{Chenhui Wang\textsuperscript{3}},
\textbf{Zhisong Wang\textsuperscript{4}},\\
\textbf{Jiashuai Liu\textsuperscript{5}},
\textbf{Zipei Wang\textsuperscript{6}},
\textbf{Cheng Wang\textsuperscript{2}},
\textbf{Wentao Pan\textsuperscript{2}},
\textbf{Mengjie Fang\textsuperscript{6}}, \\
\textbf{Di Dong\textsuperscript{6}},
\textbf{Mohammad Salmanpour\textsuperscript{7}},
\textbf{Arman Rahmim\textsuperscript{7}},
\textbf{Yu Gu\textsuperscript{8}}, \\
\textbf{Yong Xia\textsuperscript{4}},
\textbf{Hongming Shan\textsuperscript{3}}, 
\textbf{Yixuan Yuan\textsuperscript{2\corrauthormark}},
\textbf{Yefeng Zheng\textsuperscript{9\corrauthormark}},
\textbf{Lijun Lu\textsuperscript{1\corrauthormark}}
\\
\\
 \textsuperscript{1}Southern Medical University,
 \textsuperscript{2}The Chinese University of Hong Kong,\\
 \textsuperscript{3}Fudan University,
 \textsuperscript{4}Northwestern Polytechnical University,\\
 \textsuperscript{5}Xi'an Jiaotong University,
 \textsuperscript{6}Institute of Automation, Chinese Academy of Sciences, \\
 \textsuperscript{7}University of British Columbia,
 \textsuperscript{8}Microsoft Research,
 \textsuperscript{9}Westlake University
}

\begin{document}
\maketitle

\begin{center}
    \vspace{-30pt}
    {\small \textbf{\textsuperscript{*}These authors contributed equally to this work.}\quad
    \textbf{\textsuperscript{\textdagger} Project Leader.}\quad
    \textbf{\textsuperscript{\corrauthormark} Corresponding authors.}}\\[-1pt]
    \large
    \faGithub\hspace{6pt}\href{https://github.com/Saint-lsy/Awesome-3D-Radiology-Analysis-FM}{\texttt{\color{black}GitHub Repo}}
\end{center}

\vspace{10pt} 

\begin{abstract}

Advances in multimodal large language models (MLLMs) are extending radiological artificial intelligence (AI) beyond task-specific image analysis toward multimodal understanding and reasoning. Volumetric radiology, however, presents a fundamental representational mismatch: clinical interpretation often requires the integration of full-volume spatial context with acquisition-dependent quantitative information, whereas current MLLMs are commonly conditioned on selected two-dimensional (2D) images, compressed visual representations, or report-derived text. As a result, reliable volumetric radiology AI requires both representations that preserve task-relevant three-dimensional (3D) information and systems that can access, verify, and integrate this information during image interpretation and across clinical workflows.

In this Review, we examine more than 200 publications available through July 2026. Specifically, we organize the literature around volumetric representation and multimodal understanding at the model level, agentic orchestration at the system level, and relate both levels to clinical applications and their evaluation. Within this structure, we first review how volumetric foundation models learn 3D representations, align them with clinical language, and compress them for language-model conditioning. We then assess how agentic systems extend MLLMs beyond single-pass inference through planning, specialized tools, memory, and workflow interaction. Together, these analyses distinguish settings in which selected 2D views or report-mediated reasoning can suffice from those that warrant native volumetric modeling. To support critical assessment across these settings, we introduce a Claim–Design–Validation framework to assess whether technical, workflow, and clinical claims are supported by an appropriate design and a validation setting commensurate with the intended use. Across the reviewed literature, the need for native volumetric modeling and agentic capabilities depends on the spatial, quantitative, contextual, and workflow requirements of the intended task rather than on architectural complexity alone. Accordingly, progress toward clinically credible volumetric radiology AI will require faithful volumetric representation, traceable system behavior, validation commensurate with the intended claim, and clearly defined human oversight in realistic workflows.
\end{abstract}
    
\clearpage

\tableofcontents
\clearpage

\section{Introduction}
\label{sec:intro}

Volumetric radiology presents a distinctive challenge for medical artificial intelligence (AI) because clinically relevant evidence is distributed across three-dimensional (3D) space, acquisition settings, and longitudinal examinations \cite{lee2025multimodal,acosta2022multimodal,blankemeier2026merlin,zhang2023biomedclip,zhang2024generalist,chen2019med3d}. Computed tomography (CT), magnetic resonance imaging (MRI), positron emission tomography (PET), and related modalities capture anatomical structure, pathological extent, physiological characteristics, treatment response, and longitudinal change. Relevant findings may be small or sparsely distributed and may depend on continuity between slices or relationships between anatomical structures. Their interpretation also requires quantitative measurements, prior examinations, and clinical context \cite{kelly2022radiology,doshi2024quantitative,Gai2025ThreeDRad}. The difficulty therefore lies not only in processing large volumetric datasets, but also in maintaining the integrity of the spatial and contextual evidence required for a particular clinical task.

Earlier radiology AI approaches, including classical image-processing, machine-learning, and subsequent deep-learning methods, largely addressed this complexity through task-specific solutions for segmentation, detection, classification, measurement, and report generation \cite{lambin2012radiomics,gillies2016radiomics,kamnitsas2017efficient,cicek20163d}. These systems achieved substantial progress on well-defined tasks, but their representations and outputs were typically optimized for a fixed prediction objective. Foundation models broaden this paradigm by learning reusable representations that can be adapted across datasets, anatomical regions, and downstream applications. Volumetric self-supervised approaches, including Models Genesis~\cite{zhou2021models}, Swin UNETR~\cite{tang2022self}, VoCo~\cite{wu2025large}, and related methods~\cite{chen2019med3d,zhuang2019self,zhu2020rubik}, learn transferable anatomical and spatial priors directly from 3D medical images. 3D vision-language pre-training approaches, including CT-CLIP~\cite{hamamci2026generalist}, Percival~\cite{beeche2025pan}, and other related models, further align volumetric representations with radiological reports and clinical semantics. Together, these developments establish a representational basis for radiology systems that can generalize beyond a single task or label space.

Against this background, radiology is entering the era of multimodal large language models (MLLMs), which are rapidly advancing in multimodal interpretation, clinical reasoning, and tool use \cite{rao2025multimodal,lee2025multimodal,moor2023foundation,blankemeier2026merlin}. However, volumetric imaging highlights a key constraint: radiological evidence is often distributed across the full 3D examination, whereas most MLLMs \cite{zhang2024generalist,li2023llava,xu2025lingshu,chen2024towards,lai2026med,moor2023med} operate on selected two-dimensional (2D) images, compressed representations, or report-derived text. Such inputs may suffice when decisive evidence is confined to selected images or already captured in reports. By contrast, tasks involving lesion extent, anatomical relationships, quantitative burden, phase- or sequence-specific characteristics, or longitudinal change require integration across the volume. Accordingly, the need for native volumetric modeling should be determined by the evidence requirements of the task, specifically whether the available representation preserves the spatial and clinical information necessary to support reliable outputs. These evidence requirements are driving the field toward native 3D modeling, which becomes essential when clinically relevant information is distributed across the volume and cannot be reliably preserved by selected 2D representations. Domain-specific volumetric foundation models \cite{blankemeier2026merlin,jiang2025hulu,lai2026med3d,xin2025med3dvlm,sellergren2026medgemma} are well aligned with the anatomical and modality-specific structure of radiology data, while MLLMs offer flexible language-based interaction and broader reasoning capabilities. Progress in volumetric radiology AI therefore depends on integrating these complementary strengths.

\begin{figure*}[!htbp]
\centering
\includegraphics[width=\textwidth]{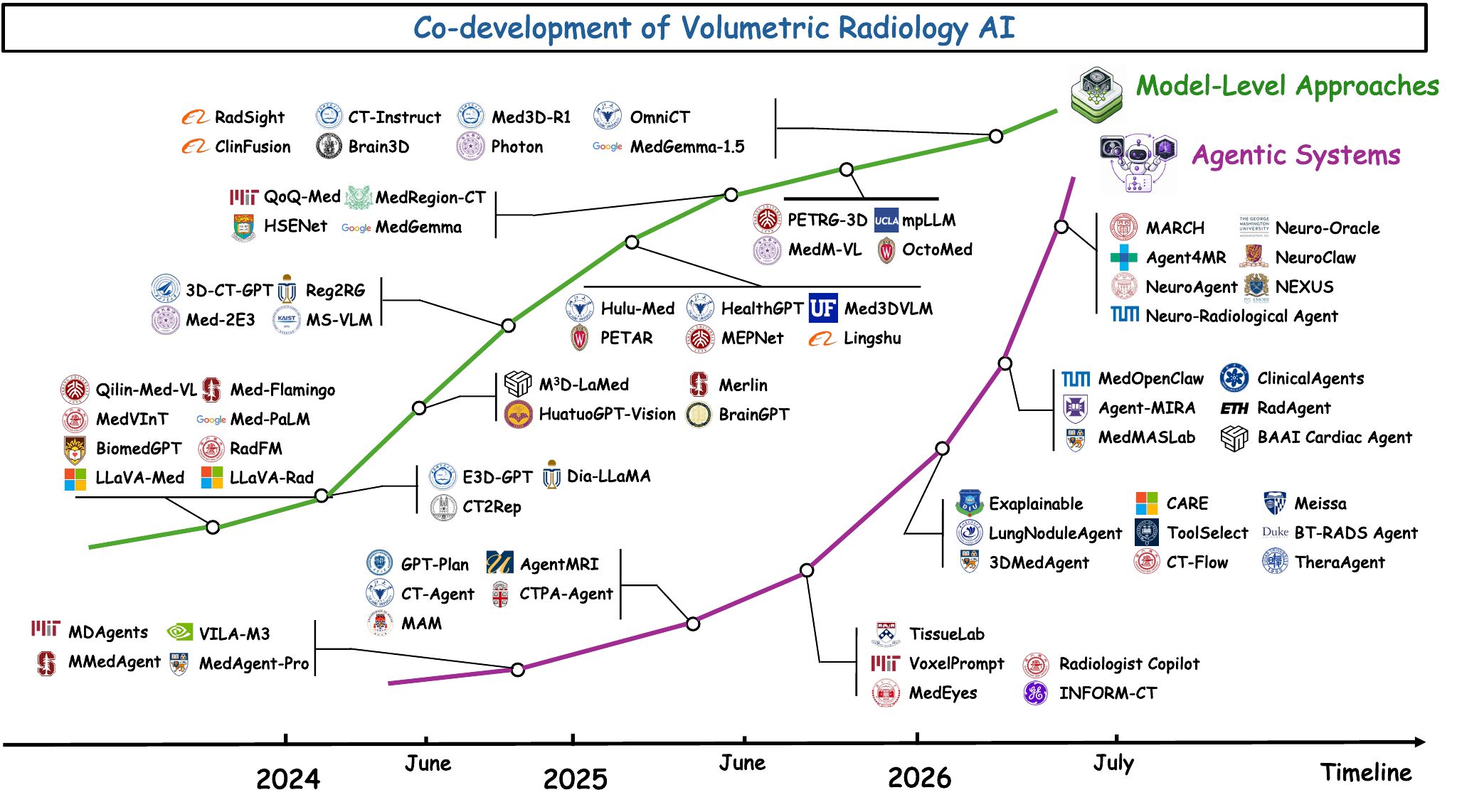}
\caption{\textbf{Co-development of model-level approaches and agentic systems in volumetric radiology AI.} The figure presents two complementary research directions. Model-level approaches, shown in green, include medical MLLMs, volumetric foundation models, and volumetric radiology MLLMs that learn or expose reusable spatial and vision–language representations. Agentic systems, shown in magenta, coordinate these representations through iterative planning, specialized tools, context management, and workflow interaction. }
\label{fig:coevolution}
\end{figure*}

Agentic systems extend this integration from model-level inference to workflow-level interaction. Direct MLLM inference generally produces an output from a predefined visual input and prompt, whereas an agentic system can progressively acquire and evaluate evidence over multiple steps \cite{kim2024mdagents,li2024mmedagent,wang2025medagent}. It may identify relevant series or anatomical regions, invoke specialized tools for image analysis and measurement, retrieve clinical context, compare prior examinations, retain intermediate observations, and revise its conclusions before returning uncertain decisions to human users. Systems such as CT-Agent~\cite{mao2025ct}, 3DMedAgent~\cite{wang20263dmedagent}, RadAgent~\cite{roschewitz2026radagent}, CT-Flow~\cite{gu2026ct}, and Radiologist Copilot~\cite{yu2025radiologist} illustrate this workflow-oriented direction. Such capabilities are particularly relevant to volumetric radiology because clinical interpretation is inherently procedural. Radiologists navigate image volumes, compare phases and sequences, perform measurements, review previous examinations, and integrate imaging evidence with clinical context. Agentic systems should therefore be understood not primarily as autonomous replacements for radiologists, but as mechanisms for targeted inspection, tool-mediated analysis, evidence verification, and human-supervised workflow coordination. Together, volumetric foundation models and agentic systems represent two mutually reinforcing research trajectories: advances in volumetric representation expand the evidence available to agents, while agentic workflows introduce new requirements for representation fidelity, clinical grounding, provenance, and reliability (Figure~\ref{fig:coevolution}).

Building on these technical capabilities, their clinical significance ultimately depends on how they are evaluated against the requirements of their intended tasks. Reporting, diagnosis, prognosis, segmentation, measurement, treatment planning, and longitudinal assessment impose different requirements for input completeness, spatial precision, uncertainty management, and human oversight~\cite{holm2026explainable,lekadir2025future}. Evaluation must therefore align with both the evidence requirements and the operational conditions of the intended application. Fluent report generation does not by itself demonstrate faithful image-based interpretation. Likewise, isolated improvements on classification or question-answering benchmarks do not necessarily establish spatial grounding or clinical usefulness. Similarly, successful tool invocation does not demonstrate that an agent selected the appropriate tool, interpreted its output correctly, or improved the surrounding workflow. These considerations make rigorous evaluation central to trustworthy AI, in which technical capability should be accompanied by traceable evidence paths, explicit provenance and audit trails, and clearly defined human responsibility~\cite{saboury2023artificial}. Clinically credible systems must therefore preserve the volumetric evidence relevant to the task, ground their outputs in imaging and clinical context, and support rather than displace human agency. Their clinical utility should be demonstrated in realistic workflows in which clinicians can examine, challenge, and override AI-supported conclusions while retaining ultimate authority and responsibility for clinical decisions.

\begin{figure}[!t]
\centering
\begingroup
\definecolor{archRed}{HTML}{D83A34}
\definecolor{archOrange}{HTML}{F28C28}
\definecolor{archGold}{HTML}{D8A200}
\definecolor{archBlue}{HTML}{2687C9}
\definecolor{archCyan}{HTML}{14A6C9}
\definecolor{archGreen}{HTML}{6DAA2C}
\definecolor{archPurple}{HTML}{7B3FB2}
\definecolor{archViolet}{HTML}{8A2EB0}

\begin{tikzpicture}[
    x=1mm,
    y=1mm,
    line cap=round,
    line join=round,
    main/.style={
        font=\bfseries\fontsize{7.1}{7.7}\selectfont,
        align=left,
        text width=41mm,
        inner sep=0.8pt
    },
    leaf/.style={
        font=\bfseries\fontsize{7.0}{7.6}\selectfont,
        align=left,
        text width=68mm,
        inner sep=0.4pt
    },
    sectiontag/.style={
        font=\fontsize{6.6}{7.2}\selectfont,
        align=left,
        inner sep=0pt
    },
    spine/.style={line width=2.05pt},
    branch/.style={line width=0.58pt},
    guide/.style={line width=0.36pt}
]

\newcommand{\spinesegment}[3]{%
    \draw[spine, #3] (8,#1) -- (8,#2);
}

\newcommand{\mainnode}[5]{%
    \node[main, anchor=west] (#1) at (18,#2+1.5) {#3};
    \node[sectiontag, anchor=west] at (18,#2-4.7) {\textnormal{(#4)}};
    \draw[spine, #5] (8,#2) -- (15,#2);
}

\newcommand{\leafnode}[3]{%
    \draw[guide, #3] (88,#1) -- (162,#1);
    \node[leaf, anchor=south west] at (91,#1+0.45) {#2};
}

\newcommand{\singlebranch}[3]{%
    \draw[branch, #3] (#1.east) .. controls (64,#2) and (75,#2) .. (88,#2);
}

\newcommand{\hubbranch}[3]{%
    \draw[branch, #3] (#1.east) .. controls (63,#2) and (67,#2) .. (72,#2);
}

\newcommand{\fanleaf}[3]{%
    \draw[branch, #3] (72,#1) .. controls (78,#1) and (80,#2) .. (88,#2);
}

\spinesegment{3}{-8}{archRed}
\spinesegment{-8}{-27}{archOrange}
\spinesegment{-27}{-54}{archBlue}
\spinesegment{-54}{-80}{archCyan}
\spinesegment{-80}{-105}{archGreen}
\spinesegment{-105}{-134}{archPurple}
\spinesegment{-134}{-148}{archViolet}

\mainnode{secintro}{0}{Introduction}{Section 1}{archRed}
\leafnode{0}{Scope, motivation, CDV lens, and contributions}{archRed}
\singlebranch{secintro}{0}{archRed}

\mainnode{secpre}{-16}{Preliminaries}{Section 2}{archOrange}
\leafnode{-10}{Traditional Volumetric Radiology analysis}{archOrange}
\leafnode{-16}{Large language models}{archOrange}
\leafnode{-22}{AI agents in healthcare and radiology}{archOrange}
\hubbranch{secpre}{-16}{archOrange}
\fanleaf{-16}{-10}{archOrange}
\fanleaf{-16}{-16}{archOrange}
\fanleaf{-16}{-22}{archOrange}

\mainnode{secfound}{-40}{Foundation Models}{Section 3}{archBlue}
\leafnode{-32}{Volumetric self-supervised representation learning}{archBlue}
\leafnode{-40}{Vision--language alignment}{archBlue}
\leafnode{-48}{3D/hybrid MLLM interpretation}{archBlue}
\hubbranch{secfound}{-40}{archBlue}
\fanleaf{-40}{-32}{archBlue}
\fanleaf{-40}{-40}{archBlue}
\fanleaf{-40}{-48}{archBlue}

\mainnode{secagent}{-68}{Agentic Systems}{Section 4}{archCyan}
\leafnode{-59}{Reasoning and planning}{archCyan}
\leafnode{-65}{Tool-augmented perception and grounded action}{archCyan}
\leafnode{-71}{Memory and dynamic context management}{archCyan}
\leafnode{-77}{Workflow interaction and multi-agent collaboration}{archCyan}
\hubbranch{secagent}{-68}{archCyan}
\fanleaf{-68}{-59}{archCyan}
\fanleaf{-68}{-65}{archCyan}
\fanleaf{-68}{-71}{archCyan}
\fanleaf{-68}{-77}{archCyan}

\mainnode{secapp}{-93}{Clinical Applications and Evaluation}{Section 5}{archGreen}
\leafnode{-84}{Diagnostic interpretation and reporting}{archGreen}
\leafnode{-90}{Prognosis and treatment response}{archGreen}
\leafnode{-96}{Image-derived planning and clinical decision support}{archGreen}
\leafnode{-102}{Cross-task evaluation and validation}{archGreen}
\hubbranch{secapp}{-93}{archGreen}
\fanleaf{-93}{-84}{archGreen}
\fanleaf{-93}{-90}{archGreen}
\fanleaf{-93}{-96}{archGreen}
\fanleaf{-93}{-102}{archGreen}

\mainnode{secdisc}{-120}{Discussion and Future Directions}{Section 6}{archPurple}
\leafnode{-108}{Volumetric and acquisition-aware intelligence}{archPurple}
\leafnode{-114}{Workflow-level clinical intelligence}{archPurple}
\leafnode{-120}{Self-evolving and embodied intelligence}{archPurple}
\leafnode{-126}{Traceable clinical-grade validation}{archPurple}
\leafnode{-132}{Scope and limitations}{archPurple}
\hubbranch{secdisc}{-120}{archPurple}
\fanleaf{-120}{-108}{archPurple}
\fanleaf{-120}{-114}{archPurple}
\fanleaf{-120}{-120}{archPurple}
\fanleaf{-120}{-126}{archPurple}
\fanleaf{-120}{-132}{archPurple}

\mainnode{secconc}{-144}{Conclusion}{Section 7}{archViolet}
\leafnode{-144}{Final synthesis and outlook}{archViolet}
\singlebranch{secconc}{-144}{archViolet}

\end{tikzpicture}
\endgroup
\caption{
\textbf{Organization of this review.} The review connects model-level volumetric representation and multimodal interpretation with system-level agentic orchestration, and relates both to clinical applications and evaluation. Volumetric foundation models determine what anatomical, quantitative, and clinical information is represented. MLLMs and agentic systems determine how that information is queried, supplemented, verified, and used within clinical and technical workflows. Clinical applications define the intended claims, evidence requirements, and evaluation conditions under which these capabilities should be interpreted.
}
\label{fig:architecture}
\end{figure}
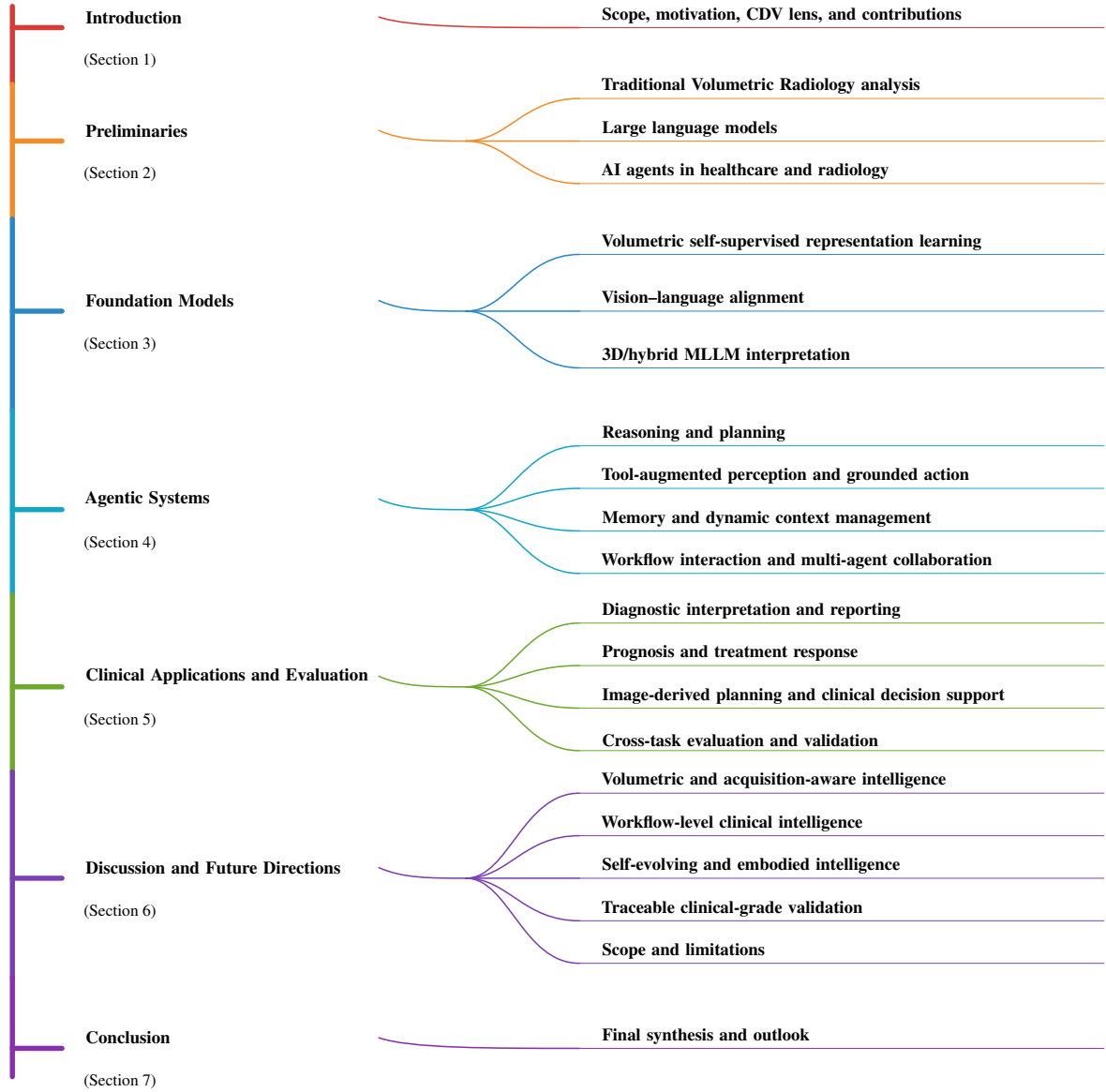

\begin{figure}[!t]
\centering
\includegraphics[width=0.98\textwidth]{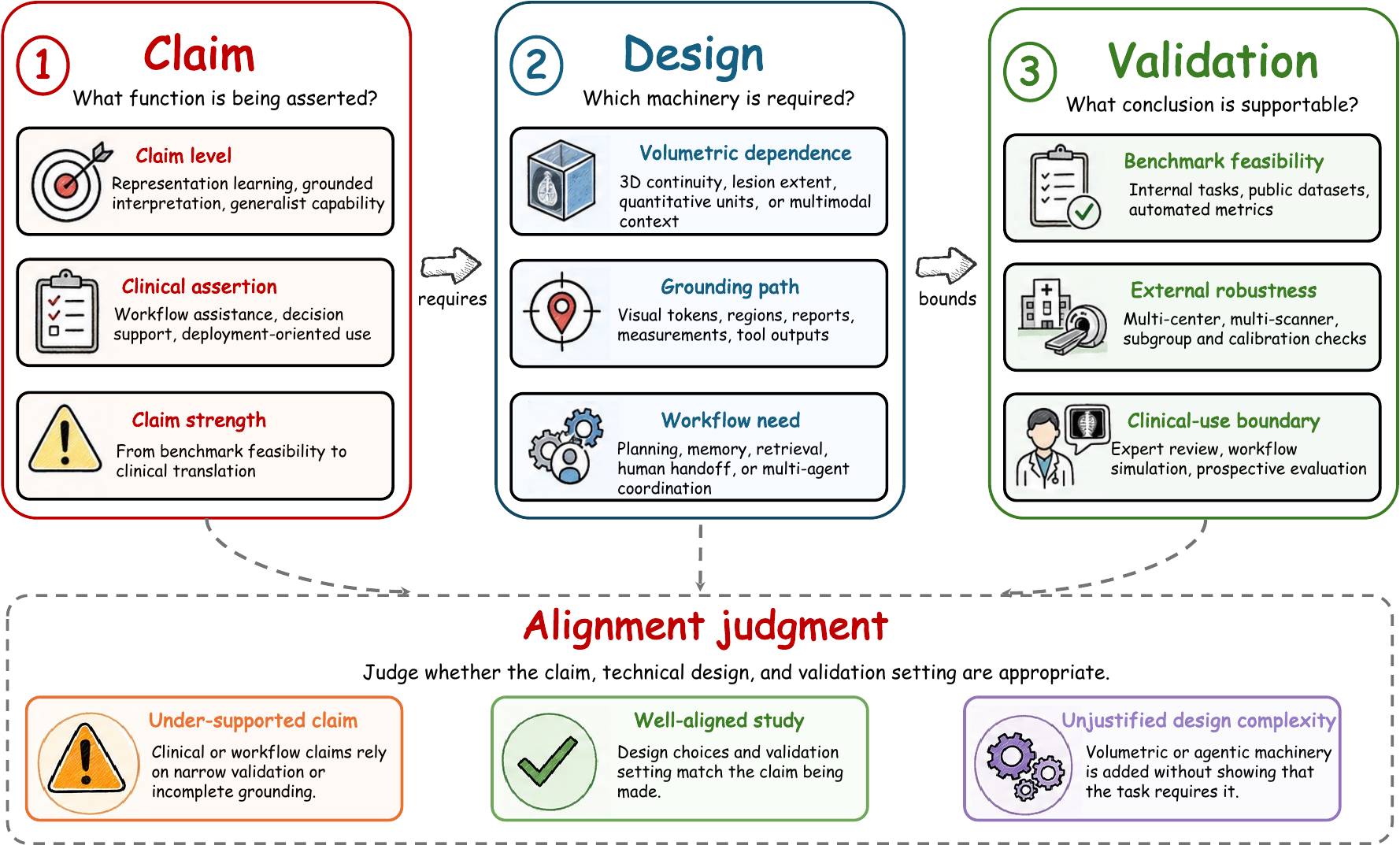}
\caption{\textbf{The Claim–Design–Validation alignment framework.} The framework evaluates the alignment among a study’s stated claim, technical design, and validation setting. The claim defines the intended capability or clinical use; the design specifies how the necessary evidence is represented and used; and the validation setting determines the strength and scope of the conclusions that can be supported.}
\label{fig:cdv}
\end{figure}

Recent reviews have examined generalist medical AI and foundation models~\cite{moor2023foundation,acosta2022multimodal}, medical MLLMs~\cite{xiao2025comprehensive,xiao2026medical}, 3D vision-language modeling~\cite{lee2025multimodal,wu2025vision}, and agentic radiology~\cite{bluethgen2025agenticradiology}. These perspectives, however, are typically examined at different levels of the system and are rarely connected through the question of how volumetric evidence is represented and preserved. In particular, existing surveys have not systematically traced how native volumetric information is transformed into multimodal representations, exposed to language models, retrieved or verified by agents, and ultimately used to support clinical outputs. As a result, the relationship between representation fidelity, downstream reasoning, workflow interaction, and clinical evaluation remains insufficiently characterized. This review draws on more than 200 publications available through July 2026 and organizes the literature around volumetric foundation models and volumetric radiology MLLMs at the model level, agentic radiology systems at the system level, and clinical applications and evaluation resources. Primary emphasis is placed on work published since 2022, with selected earlier studies included to establish the field’s methodological foundations. We place volumetric evidence at the center of this synthesis and examine how its preservation, accessibility, and use are shaped by model design and system behavior. Specifically, we consider when 2D or report-mediated reasoning is sufficient, when native volumetric modeling is necessary, and how language-based and agentic systems can retrieve, verify, and act on spatial evidence while preserving clinical fidelity. We use the term ``volumetric radiology AI'' to encompass representation learning from 3D medical images, alignment between volumetric and clinical-language information, MLLM-based interpretation, tool-mediated agentic systems, and their evaluation within radiological tasks and workflows. As summarized in Figure~\ref{fig:architecture}, we organize the review using a model–system framework linked to clinical applications and evaluation: how volumetric evidence is represented, how it is accessed and acted upon, and how the resulting capabilities are evaluated in clinically relevant settings.

To support a critical rather than purely descriptive synthesis, we use a concise Claim--Design--Validation alignment framework to appraise the studies surveyed in this review (Figure~\ref{fig:cdv}). The framework considers whether a study's stated claim is supported by an appropriate technical design and by a validation setting that reflects the intended use. For volumetric radiology AI, this assessment includes whether the task requires native volumetric modeling, multimodal grounding, tool-mediated analysis, memory, planning, or human handoff, and whether the reported evaluation is sufficient to support the corresponding technical, workflow, or clinical claim. The framework therefore complements the evidence-centered perspective of this review by assessing the alignment among task requirements, system capabilities, and validation conditions.

\begin{figure}[p]
\centering
\makebox[\textwidth][c]{%
\hspace*{-5.7mm}%
\resizebox{0.98\textwidth}{!}{%
\input{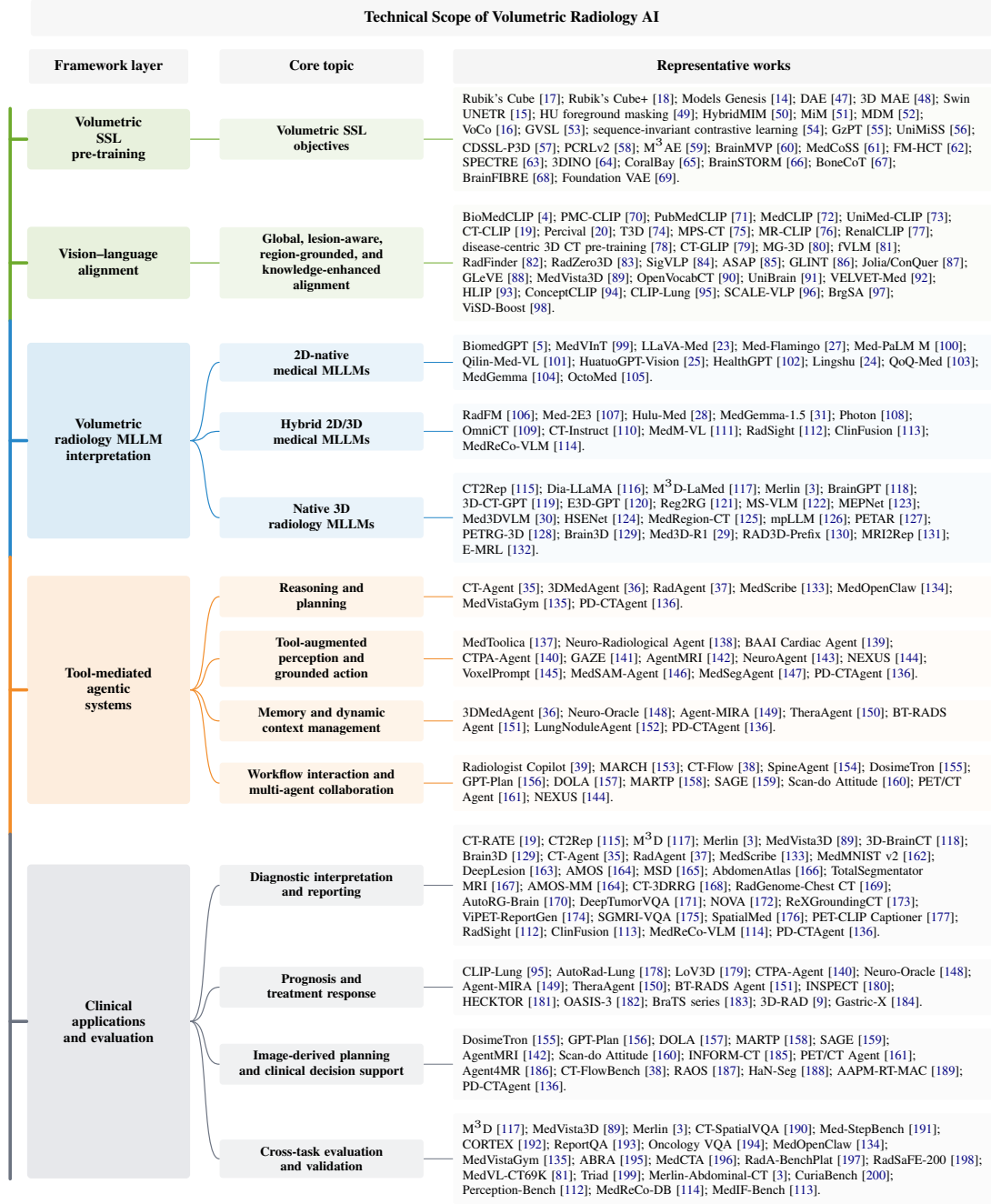}%
}%
\hspace*{5.7mm}%
}
\caption{
\textbf{Technical scope of this review.}
The reviewed literature spans volumetric self-supervised pre-training, vision–language alignment, volumetric radiology MLLMs across 2D-native or selected-view, slice-sequence, hybrid 2D/3D, and native 3D designs, agentic radiology systems, and clinical applications and evaluation. Representative methods and resources are organized according to the corresponding sections and tables.
}
\label{fig:technical_scope}
\end{figure}

The contribution of this review lies in connecting research that has previously been discussed from largely separate model-level, system-level, and application/evaluation perspectives. Figure~\ref{fig:technical_scope} summarizes this technical scope and maps the major research directions to the corresponding sections and tables. We first examine volumetric self-supervised representation learning, vision–language alignment, and volumetric radiology MLLMs across 2D-native or selected-view, slice-sequence, hybrid 2D/3D, and native 3D designs. We then consider how agentic systems extend or complement these models through iterative evidence acquisition, specialized tool use, context management, and workflow interaction. Finally, we relate these technical capabilities to clinical applications and evaluation, with particular attention to whether systems preserve task-relevant volumetric and quantitative evidence, ground their outputs in identifiable image and clinical context, maintain traceable provenance, and demonstrate utility under realistic conditions. This perspective provides a structured basis for assessing how model-level representation and system-level orchestration can contribute to clinically credible radiology AI.


\section{Preliminaries}

This section provides a high-level technical background for the rest of the review. Guided by this field-level model–system–application progression, we trace how volumetric radiology analysis has evolved from task-specific image-analysis pipelines to language-centered multimodal interfaces and, increasingly, to agentic systems that coordinate models, tools, memory, and clinical workflows. The goal is not to duplicate the detailed surveys in later sections, but to establish the conceptual transitions that have driven the emergence of volumetric foundation models and agentic systems.

\subsection{Traditional Volumetric Radiology Analysis Methods}

Early computational paradigm for volumetric radiology treated 3D examinations across diverse radiological and nuclear medicine modalities as structured volumes from which clinically meaningful measurements could be extracted. A typical pipeline consisted of image acquisition and reconstruction, registration or normalization, region-of-interest delineation, handcrafted feature extraction, and a downstream statistical or machine-learning model~\cite{salmanpour2026handcrafted}. Radiomics~\cite{lambin2012radiomics,gillies2016radiomics} made this paradigm explicit by converting volumetric images into high-dimensional quantitative descriptors of intensity, shape, texture, and spatial heterogeneity, and by linking these descriptors to diagnosis, prognosis, or treatment response. This represented an important shift from qualitative visual interpretation toward the development of image-derived quantitative biomarkers.

However, traditional feature-engineering pipelines had limited robustness. Their performance depended heavily on segmentation quality, scanner protocol, reconstruction kernel, voxel spacing, feature standardization, and the chosen classifier. Because the representations were usually handcrafted for a specific organ, lesion, or endpoint, transferability across institutions, modalities, and tasks was limited. In addition, most systems produced fixed, task-specific outputs rather than interactive, context-aware support: they could perform classification, segmentation, or retrieval, but they could not naturally discuss uncertainty, compare findings across prior examinations, or adapt their workflow to a clinician's question.

Deep learning changed the representation layer by replacing handcrafted descriptors with learned features. In volumetric radiology, this transition was especially important because isolated 2D slices may fail to capture information contained in the volumetric context. Early 3D convolutional systems such as DeepMedic~\cite{kamnitsas2017efficient} used multi-scale 3D CNNs to combine local detail with contextual information for brain lesion segmentation. Encoder-decoder architectures then became the dominant template for dense volumetric prediction. 3D U-Net~\cite{cicek20163d} extended the U-Net design to sparse-to-dense volumetric segmentation, while V-Net~\cite{milletari2016vnet} introduced a fully convolutional 3D architecture and Dice-based optimization for imbalanced medical segmentation. Later, nnU-Net~\cite{isensee2021nnu} demonstrated that robust biomedical segmentation depends not only on network design but also on systematic self-configuration of preprocessing, architecture, training, and post-processing across datasets. Transformer-based 3D models such as UNETR~\cite{hatamizadeh2022unetr}, and Swin UNETR~\cite{tang2022self} further expanded the receptive field and made long-range anatomical context easier to model.

Despite these advances, supervised 3D deep networks remained largely task-specific. They usually required curated labels, produced outputs within a fixed prediction space, and were optimized for one clinical objective at a time. Their capabilities did not transfer automatically across tasks: a segmentation model did not automatically become a report generator; a classifier did not automatically expose a reasoning trace; and a volumetric backbone trained on one dataset did not necessarily transfer to another modality or institution. This limitation motivates the next stage of the field: foundation-style representation learning, vision-language alignment, and multimodal models that can reuse volumetric perception across tasks.

\subsection{Large Language Models}

Large language models constitute the second background pillar for modern volumetric radiology systems. The Transformer~\cite{vaswani2017attention} architecture enabled highly parallel sequence processing with attention-based modeling of long-range dependency. GPT-3~\cite{brown2020language} further demonstrated that large-scale autoregressive language modeling can support strong in-context learning. Instruction tuning and reinforcement learning from human feedback, represented by InstructGPT~\cite{ouyang2022training}, improved the usability of LLMs in interactive settings by aligning next-token prediction with user intent and dialogue behavior. Prompting methods such as chain-of-thought reasoning~\cite{wei2022chain} demonstrated that LLMs can generate explicit intermediate reasoning steps, while ReAct-style prompting~\cite{yao2023react} connected reasoning with action selection and tool use.

For radiology, the value of LLMs does not lie in replacing image interpretation with text. Rather, their value lies in supporting clinical language understanding, report generation, question answering, explanations of differential diagnoses, knowledge retrieval, and reasoning over contextual information. These capabilities are important because radiology interpretation extends beyond visual pattern recognition. A comprehensive diagnostic process often requires the integration of imaging findings with anatomical knowledge, prior reports, clinical history, guideline thresholds, uncertainty estimation, and downstream clinical decisions~\cite{salmanpour2026clinically,salmanpour2026radiological,jouzdani2026towards,gorji2026radiological}.

At the same time, LLMs expose a key mismatch. Their native input is text, whereas volumetric radiology information is dense, spatial, quantitative, with clinically important evidence often sparsely distributed. If a CT volume is reduced to a few sentences or a small set of slices before reaching the LLM, subtle findings, longitudinal correspondence, and lesion-level information may be lost. Therefore, the role of the LLM in volumetric radiology should be understood as a reasoning and communication layer built upon reliable spatial grounding rather than as a replacement for volumetric perception. This motivates the subsequent discussion of representation learning, vision-language alignment, volumetric radiology MLLM architectures, and agentic radiology workflows.

\subsection{AI Agents in Healthcare and Radiology}

AI agents extend the capabilities of LLMs and MLLMs from answer generators into goal-directed systems. At minimum, an agent maintains a task state, plans intermediate steps, invokes tools, observes results, updates memory, and decides when to answer or defer to human review. This framing is particularly natural in medicine because clinical reasoning is distributed across heterogeneous evidence sources: images, reports, laboratory values, prior examinations, guidelines, measurements, and specialist tools. General medical agents such as MDAgents~\cite{kim2024mdagents}, MMedAgent~\cite{li2024mmedagent}, and MedAgent-Pro~\cite{wang2025medagent} have therefore explored adaptive collaboration among LLMs for medical decision-making, tool selection for multimodal medical tasks, and evidence-based diagnostic workflows.

Radiology agents differ from general medical agents because the relevant evidence is spatially structured and image-grounded. A useful radiology agent should not only produce a plausible conclusion, but also inspect the correct image regions, invoke the appropriate measurement or segmentation tools, preserve anatomical provenance, compare with prior examinations when needed, and expose sufficient intermediate observations and tool outputs for clinical review. Early radiology-oriented agents already show this shift. 
MedRAX~\cite{fallahpour2025medrax} integrates chest X-ray analysis tools with an MLLM-based reasoning workflow, while CT-Agent~\cite{mao2025ct} and 3DMedAgent~\cite{wang20263dmedagent} extend agentic reasoning to volumetric CT question answering and 3D medical analysis. RadAgent~\cite{roschewitz2026radagent} and CT-Flow~\cite{gu2026ct} further emphasize stepwise CT interpretation and workflow orchestration. Radiologist Copilot~\cite{yu2025radiologist} and the Neuro-Radiological Agent~\cite{erdur2026agenticlargelanguagemodels} highlight the role of human-in-the-loop interaction and tool-mediated clinical assistance.


The distinctive value of agentic systems in volumetric radiology lies in controlled access to image-grounded evidence and coordination across workflow steps. When an intended task depends on through-plane continuity, the system must be able to access the corresponding full-volume evidence. Compared with generic clinical agents, radiology agents must link language outputs to image regions, measurements, series, and tool provenance. This review therefore treats agentic radiology as a system-level extension of, or complement to, volumetric foundation models and volumetric radiology MLLMs: representation learning provides perceptual priors, vision–language alignment provides language-addressable semantics, MLLMs provide interactive interpretation, and agentic systems organize these capabilities within traceable, human-supervised workflows.


\section{Foundation Models in Volumetric Radiology}
\label{sec:foundation}

The rapid advancement of artificial intelligence in medical imaging has ushered in a new era of foundation models for volumetric radiology analysis. In this review, we use foundation models to refer to models pretrained on broad volumetric, visual, textual, or multimodal medical data whose representations or interfaces can be adapted to multiple downstream tasks rather than optimized for a single fixed label space. Under this definition, volumetric self-supervised encoders, vision-language pretrained models, and 3D or hybrid MLLMs are connected by a shared role: they provide reusable perceptual, semantic, or instruction-following capabilities for more than one clinical task. Unlike 2D natural images, 3D radiological data pose unique challenges due to their complex anatomical structures, volumetric nature, and the scarcity of high-quality expert annotations. Before volumetric radiology models can support generation, dialogue, or multi-step understanding, they must first acquire reliable volumetric representations, connect these representations to clinical semantics, and expose the resulting visual information to language models in a form that supports reasoning and communication.

\begin{figure*}[t]
    \centering
    \includegraphics[width=0.98\textwidth]{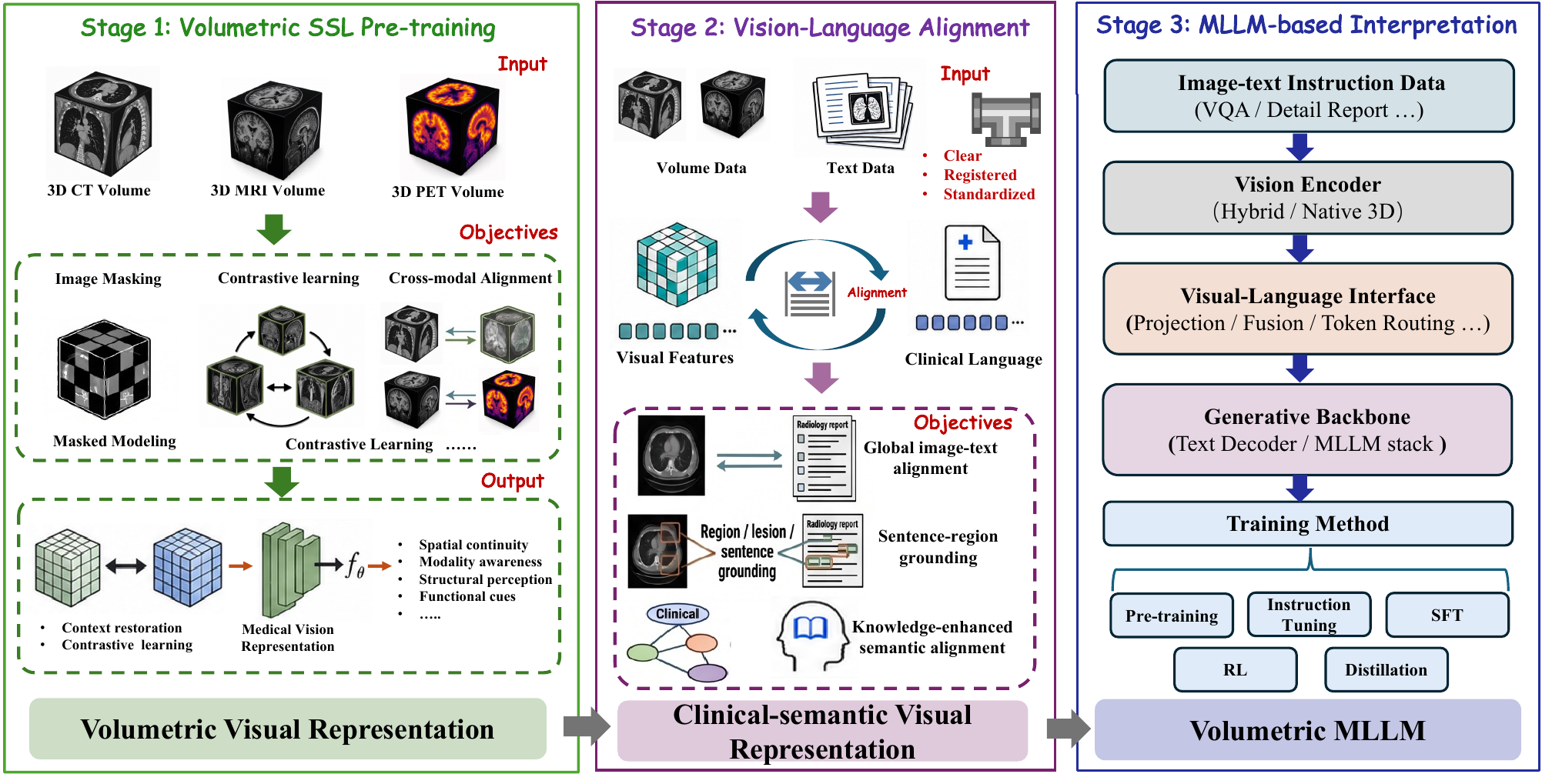}
    \caption{\textbf{Model-level foundation and MLLM components for volumetric radiology.} The model landscape is organized into three linked functional stages. Volumetric self-supervised representation learning acquires anatomy-aware priors from unlabeled scans through masked, restorative, deformation-based, and contrastive objectives. Vision–language alignment makes these representations addressable through clinical language at global, lesion-aware, region-grounded, and knowledge-enhanced levels. MLLM-based interpretation couples vision encoders, vision–language interfaces, generative backbones, and training curricula.}
    \label{fig:MLLM_modules}
\end{figure*}

We therefore organize the model landscape as a progression from volumetric self-supervised representation learning to vision-language alignment and then to MLLMs. This progression is functional rather than purely architectural. Volumetric self-supervised learning (SSL) explains how transferable anatomical and spatial priors are learned from scans; vision–language pre-training explains how these priors become addressable by clinical language; and MLLMs explain how visual representations are converted into language-model context for generation, dialogue, and instruction following. The challenge of compressing dense visual information becomes increasingly explicit across this progression. Two-dimensional slice encoders, native 3D encoders, and hybrid 2D/3D encoders appear in SSL, vision–language pre-training, and MLLMs; their consequences become most explicit in MLLMs because the encoded volume must be tokenized, compressed, and exposed to an autoregressive decoder. Figure~\ref{fig:MLLM_modules} summarizes this three-stage foundation-model layer and the shared constraint of preserving clinically decisive 3D information for grounded reporting.

\subsection{Volumetric Self-Supervised Representation Learning}
\label{sec:VSSRL}

Volumetric self-supervised representation learning provides the perceptual substrate of volumetric radiology foundation models. Its scope is narrower than the full representation-learning literature: supervised, semi-supervised, and weakly supervised training remain important for task-specific radiology systems, but they usually depend on predefined labels, organs, lesions, or clinical endpoints. In contrast, SSL is particularly suitable for the foundation-model layer because unlabeled 3D scans are far more scalable than dense voxel-level annotations or carefully paired reports, while the volumes themselves contain rich intrinsic structure, including slice continuity, anatomical topology, modality-specific appearance, and spatial redundancy. By converting these structures into pretext supervision, SSL can learn transferable anatomical and spatial priors before the model is specialized for language alignment, instruction following, or downstream clinical tasks. Accordingly, this subsection focuses on volumetric SSL objectives, including masked modeling, context restoration, deformation prediction, and contrastive learning, and on how these objectives support reusable 3D visual backbones for later vision–language pre-training and MLLM integration.

\subsubsection{Self-Supervised Objectives}

SSL aims to learn transferable anatomical representations from unlabeled radiology volumes by optimizing pretext tasks that exploit the inherent structural redundancy of volumetric data. Formally, given an unlabeled dataset $\mathcal{D} = \{\mathbf{X}_i\}_{i=1}^N$, SSL optimizes an encoder $f_\theta$ by minimizing a pretext loss:
\begin{equation}
\theta^* = \arg\min_\theta \mathbb{E}_{\mathbf{X} \sim \mathcal{D}} [\mathcal{L}_{\text{pretext}}(\mathbf{X}, \theta)].
\end{equation}
In the context of 3D medical imaging, three broad objective families are especially influential:

\noindent\textbf{Masked Image Modeling (MIM).}
Extending the success of masked autoencoders (MAE~\cite{he2022masked}) to 3D, this approach minimizes a reconstruction loss:
\begin{equation}
\mathcal{L}_{\text{recon}} = \Vert g_\phi(f_\theta(\tilde{\mathbf{X}})) - \mathbf{X} \Vert_2^2,
\end{equation}
where $\tilde{\mathbf{X}}$ is a masked version of the volume. By forcing the model to recover missing voxels, 3D MIM encourages long-range spatial modeling across the volume.

\noindent\textbf{Contrastive Learning.}
This paradigm minimizes an InfoNCE~\cite{he2020momentum} loss:
\begin{equation}
\mathcal{L}_{\text{nce}} = -\log \frac{\exp(\text{sim}(z_i, z_i^+)/\tau)}{\sum_{j} \exp(\text{sim}(z_i, z_j)/\tau)},
\end{equation}
to maximize the similarity between differently augmented views of the same instance. In general, contrastive learning encourages the model to learn discriminative, high-level semantic features by distinguishing between diverse anatomical structures in the latent space. In radiology, however, the design of positives and negatives must be handled carefully because different crops may share identical tissues, and clinically relevant abnormalities may be spatially sparse.

Together, these SSL frameworks establish the visual perception required for subsequent multimodal integration, allowing a 3D encoder to acquire anatomical and spatial priors before linguistic supervision.

\subsubsection{Representative Approaches}

For clarity, we group the representative approaches according to the dominant pre-training signal emphasized in each method. In the surveyed literature, these signals commonly arise from restoring disrupted anatomy, reconstructing masked or deformed structures, enforcing geometric or contrastive invariance, or scaling SSL across dimensions, modalities, and datasets.

\noindent\textbf{Heuristic Context Restoration.}
Early frameworks constructed self-supervision through handcrafted image transformations. For instance, Models Genesis \cite{zhou2021models} systematically designed transformation strategies such as non-linear appearance mapping, local pixel shuffling, and out-painting, forcing the network to restore the original anatomical structure. Recent extensions, such as DAE \cite{valanarasu2024disruptive}, combine local channel masking with low-level feature perturbations (e.g., adding noise and downsampling) to increase reconstruction difficulty and encourage recovery of granular anatomical details.

\noindent\textbf{Anatomically-Guided and Hierarchical Masked Image Modeling.}
The MIM paradigm has been widely adapted for 3D representation learning, with initial transformer-based approaches (e.g., 3D MAE \cite{chen2023masked} and Swin UNETR \cite{tang2022self}) proving its strong transferability. However, naive random masking is inefficient when large portions of 3D scans contain air or background. To address this, HU-based foreground masking \cite{lee2025hu} leverages Hounsfield Unit density distributions to specifically mask and reconstruct diagnostically meaningful tissue regions. To handle the high dimensionality and varied structural scales of 3D data, hierarchical designs such as HybridMIM \cite{xing2024hybrid} and MiM \cite{zhuang2025mim} reconstruct features at both coarse regional levels and fine pixel levels simultaneously. Moreover, Masked Deformation Modeling (MDM \cite{lyu2024masked}) incorporates spatial transformations by predicting dense deformation fields alongside voxel intensities, thereby introducing registration-aligned structural priors into the MIM process.

\noindent\textbf{Geometry-Aware and Invariant Contrastive Learning.}
Standard contrastive methods often construct false negative pairs in medical volumes because different cropped patches may share identical semantic tissues or backgrounds. To resolve this, researchers inject geometric and topological priors into the contrastive objective. The VoCo framework \cite{wu2025large} extracts base crops and predicts the contextual position of random crops by contrasting their similarities, implicitly encoding organ layout relations into the representations. Similarly, Geometric visual similarity learning \cite{he2023geometric} enforces local and global semantic matching guided by topological invariance. Furthermore, to ensure invariant representations across varied acquisition protocols, sequence-invariant contrastive paradigms align augmented contrasts derived from quantitative MRI \cite{chalcroft2025unified}, while human-prior aligned methods (e.g., GzPT \cite{wang2025improving}) leverage eye-tracking sequences to bias contrastive objectives toward clinically relevant regions.

\noindent\textbf{Cross-Dimensional and Foundation-Scale Self-Supervised Learning.}
To reduce the gap between abundant 2D slices and scarcer full-volume supervision, cross-dimensional frameworks such as UniMiSS \cite{xie2022unimiss} and CDSSL-P3D \cite{gao2024cross} jointly exploit planar and volumetric data. Similarly, to maintain high-level semantics alongside pixel-level precision, multi-task frameworks such as PCRLv2 \cite{zhou2023unified} simultaneously optimize contrastive comparison and multi-scale image restoration. Recent work has also expanded SSL toward foundation-scale and multimodal pre-training. Models such as M$^{3}$AE \cite{liu2023m3ae} and BrainMVP \cite{rui2025multi} address naturally grouped multi-sequence MRI inputs, supporting cross-sequence representations even when some sequences are missing. MedCoSS \cite{ye2024continual} further broadens this direction by formulating multi-modal medical SSL as a continual pre-training problem across clinical reports, X-rays, CT, MRI, and pathological images, using rehearsal-based learning to mitigate modality conflicts and catastrophic forgetting. Meanwhile, large-scale CT/MRI foundation models (e.g., BrainSTORM \cite{wang2025towards}, FM-HCT \cite{zhu20253d}, SPECTRE \cite{claessens2025scaling}, and 3DINO \cite{xu2025generalizable}) use broader pre-training corpora and larger compute budgets to improve transfer across anatomical regions, scanners, and tasks. Recent CT-specific SSL work further emphasizes this scaling direction: CoralBay \cite{gatopoulos2026coralbay} extends DINO-style self-distillation to 3D CT with a hierarchical Swin transformer and multi-scale feature distillation, showing how foundation-scale SSL can preserve both global semantics and local anatomical structure.

In summary, volumetric SSL mitigates the annotation bottleneck by converting intrinsic 3D structure into transferable supervision. By encoding anatomical continuity, topological invariants, modality-specific appearance, and spatial relationships, these pre-training objectives provide reusable visual backbones for later alignment and MLLM integration.

\subsection{Vision–Language Alignment}
\label{sec:VLA}
While volumetric SSL establishes robust anatomical perception, it remains inherently language-agnostic. To make radiological representations addressable through clinical semantics and diagnostic language, vision–language pre-training, largely inspired by CLIP\cite{radford2021learning}, aligns visual features with paired clinical text. This stage should be understood as a semantic bridge rather than simply another generic training strategy: it converts visual features into a language-addressable representation space for retrieval, open-vocabulary recognition, report grounding, and downstream MLLM integration.

\subsubsection{Alignment Objectives}

The most influential formulation of vision--language alignment is the CLIP-style contrastive objective~\cite{radford2021learning}, which can be viewed as a cross-modal extension of the InfoNCE principle~\cite{he2020momentum}. In visual contrastive learning, two augmented views of the same image or volume form a positive pair, while other samples serve as negatives. Vision--language contrastive learning instead aligns an image or volume with its associated report and separates mismatched pairs. In radiology, this enables the visual encoder to learn clinically meaningful representations organized around anatomy, findings, diseases, and diagnostic impressions.

Given a dataset $\mathcal{D}=\{(\mathbf{X}_i,\mathbf{T}_i)\}_{i=1}^{N}$, where $\mathbf{X}_i$ is the $i$-th image or volume, $\mathbf{T}_i$ is its associated report, and $N$ is the number of pairs, vision--language pre-training jointly optimizes a vision encoder $f_\theta$ and a text encoder $h_\omega$. For a mini-batch of $B$ pairs, the encoders produce $\ell_2$-normalized embeddings:
\begin{equation}
z_i^v
=
\frac{f_\theta(\mathbf{X}_i)}
{\|f_\theta(\mathbf{X}_i)\|_2},
\qquad
z_i^t
=
\frac{h_\omega(\mathbf{T}_i)}
{\|h_\omega(\mathbf{T}_i)\|_2}.
\end{equation}
The embeddings $z_i^v$ and $z_i^t$ represent the visual study and report, respectively, while $\theta$ and $\omega$ denote the encoder parameters.

The image-to-text contrastive loss is
\begin{equation}
\mathcal{L}_{v \rightarrow t}
=
-\frac{1}{B}\sum_{i=1}^{B}
\log
\frac{\exp\left(\operatorname{sim}(z_i^v,z_i^t)/\tau\right)}
{\sum_{j=1}^{B}\exp\left(\operatorname{sim}(z_i^v,z_j^t)/\tau\right)}.
\end{equation}
In this expression, $\operatorname{sim}(\cdot,\cdot)$ denotes cosine similarity, $\tau$ is the temperature parameter, and $j$ indexes the candidate reports in the mini-batch.

Symmetrically, the text-to-image loss is
\begin{equation}
\mathcal{L}_{t \rightarrow v}
=
-\frac{1}{B}\sum_{i=1}^{B}
\log
\frac{\exp\left(\operatorname{sim}(z_i^t,z_i^v)/\tau\right)}
{\sum_{j=1}^{B}\exp\left(\operatorname{sim}(z_i^t,z_j^v)/\tau\right)}.
\end{equation}
This direction treats each report as a query and retrieves its paired visual study from the $B$ candidates.

The final CLIP-style objective averages the two directional losses:
\begin{equation}
\mathcal{L}_{\mathrm{vlp}}
=
\frac{1}{2}
\left(
\mathcal{L}_{v \rightarrow t}
+
\mathcal{L}_{t \rightarrow v}
\right),
\end{equation}
where $\mathcal{L}_{\mathrm{vlp}}$ denotes the overall bidirectional vision--language alignment objective.

Early medical vision–language pre-training methods largely followed this recipe: they built biomedical image-text pairs from publications, radiology images, or paired clinical captions, and optimized the symmetric contrastive loss so that images and reports became mutually retrievable. The resulting models inherit the practical advantages of CLIP \cite{radford2021learning}: once image and text share a representation space, disease names or report phrases can be used as textual prompts for zero-shot recognition, retrieval, and downstream adaptation. In medical imaging, however, paired data scarcity, report complexity, and the dense spatial structure of 3D scans make this objective only a starting point. Research in this domain has progressed from 2D foundations to more structured and volumetric alignment strategies, as summarized below.

\subsubsection{Representative Approaches}

To structure this literature, we organize vision–language pre-training approaches according to the level at which image--text correspondence is modeled. Large-scale global alignment methods establish study-level compatibility between volumes and reports; fine-grained and lesion-aware methods introduce local or region-level constraints to preserve spatially sparse findings; and knowledge-enhanced or hierarchical methods further structure the text side so that clinical concepts, negation, uncertainty, synonyms, and report hierarchy are handled more explicitly. These categories are not mutually exclusive, but they reflect increasingly explicit modeling of both spatial correspondence and clinical semantics.

\noindent\textbf{From 2D Foundations to 3D Global Alignment.}
The fundamental challenge of medical vision–language pre-training is the domain gap between generic visual concepts and domain-specific clinical semantics, compounded by data scarcity. Early progress focused predominantly on 2D imaging, where large-scale image-text resources are more accessible. Foundation models such as BioMedCLIP \cite{zhang2023biomedclip}, PMC-CLIP \cite{lin2023pmc}, and PubMedCLIP \cite{eslami2023pubmedclip} achieved robust 2D alignment by leveraging millions of image-caption pairs extracted from biomedical literature. Concurrently, frameworks such as MedCLIP \cite{wang2022medclip} reduced dependence on paired data by decoupling images and texts, while UniMed-CLIP \cite{khattak2024unimed} unified 2D medical-image pre-training.

Translating this progress to volumetric radiology introduces the challenge of pairing dense, high-dimensional volumetric data with corresponding clinical reports. Representative 3D methods first scaled global volume-report alignment. Models such as CT-CLIP \cite{hamamci2026generalist}, Percival \cite{beeche2025pan}, and T3D \cite{liu2025t3d} assemble large-scale 3D volume-report corpora (ranging from 25k to 400k pairs) to align global 3D embeddings with corresponding findings. Addressing the scarcity of expert-written reports, MPS-CT \cite{li2025more} extracts low-cost ``silver-standard'' labels via LLMs to augment pre-training, while MR-CLIP \cite{avci2025metadata} bypasses textual reports entirely by aligning brain MRIs with structured Digital Imaging and Communications in Medicine (DICOM) \footnote{DICOM is
the international standard for transmitting, storing, retrieving, processing,
and displaying medical imaging information. The current DICOM standard is
managed by the Medical Imaging \& Technology Alliance, a division of the
National Electrical Manufacturers Association (NEMA).}  metadata. Disease-centered vision–language pre-training is also emerging as a complementary route. RenalCLIP \cite{tao2026renalclip} narrows CT-language alignment to precision oncology in kidney cancer, while disease-centric 3D CT pre-training with hybrid visual encoding \cite{shi2026diseasecentric} uses disease-level supervision to reduce the semantic coarseness of study-level report matching.

\noindent\textbf{Fine-Grained and Lesion-Aware Alignment.}
A critical limitation of global alignment is information compression. Pooling an entire high-resolution 3D volume into a single embedding can dilute spatially sparse but clinically important findings, particularly small lesions. To mitigate this limitation, recent methods decompose visual and textual inputs to enforce anatomically grounded, fine-grained correspondences.

This decomposition also helps address the ``false negative collision'' problem prevalent in contrastive batches, where normal tissues or synonymous diseases may be incorrectly penalized; fVLM \cite{shui2025large} addresses this issue with anatomy-level CT--report alignment and dual false-negative reduction during contrastive pre-training. For stricter spatial localization, RadFinder \cite{ging2026learning} leverages text-mined slice indices as weak axial-depth anchors, while RadZero3D \cite{park2025radzero3d} and SigVLP \cite{wang2026sigvlp} adapt 3D chunking architectures with direct patch-to-text or chunk-level cross-attention. Recent methods make this decomposition more explicit. ASAP \cite{wang2026asap} introduces anatomy-aware semantically adaptive pre-training for volumetric scans, GLINT \cite{park2026glint} uses sparse language-image gates to activate query-relevant image patches, and Jolia/ConQuer \cite{khlaut2026jolia} augments global CT-report contrastive learning with concept-specific queries. Lesion-level grounding methods such as GLeVE \cite{jiang2026gleve} further treat report descriptions as atomic semantic units and verify one-to-one correspondence with 3D lesion proposals. To balance details with holistic context, multi-scale strategies such as MedVista3D \cite{li2025medvista3d} and OpenVocabCT \cite{li2025towards} combine local constraints with global alignment for open-vocabulary segmentation and anomaly detection.

\noindent\textbf{Knowledge-Enhanced and Hierarchical Semantic Alignment.}
Another limitation of standard CLIP is its treatment of clinical text as a flat sequence. Clinical narratives are noisy, structured, and less spatially redundant than volumetric pixels. To reduce misalignment and false positive/negative noise propagated by standard InfoNCE, researchers have added explicit clinical-semantics modules to vision–language pre-training.

To address the rigid, linear processing of standard CLIP, recent frameworks emphasize modeling the inherent structural hierarchies of clinical data. Because standard contrastive learning ignores nested information, methods such as UniBrain \cite{lei2025unibrain} and VELVET-Med \cite{zhang2025velvet} explicitly decouple contrastive losses across multiple semantic levels, ranging from words and sentences to modality-specific sub-reports and global conclusions, to preserve both granularity and holistic context. Extending this to operational workflows, HLIP \cite{zhao2025towards} implements a slice-to-scan-to-study hierarchical attention mechanism, enabling robust alignment directly on uncurated clinical archives. 

Simultaneously, raw text matching is poorly suited to negations, uncertainties, and synonymous terminologies (e.g., ``consolidation'' vs. ``infiltrate''). A prominent strategy is therefore to inject external medical knowledge into the alignment objective. Building on 2D predecessors such as ConceptCLIP \cite{nie2025conceptclip} and CLIP-Lung \cite{lei2023clip}, 3D frameworks such as SCALE-VLP \cite{mahdizadeh2025scale} replace rigid binary matching with a soft-weighted contrastive objective that combines volumetric spatial-coherence weights with report-level medical-knowledge embeddings derived from a frozen medical language model. This replaces the rigid binary matching of standard InfoNCE with a clinically structured similarity target. Furthermore, to bridge the persistent modality gap caused by disparate information densities between sparse text and redundant volumes, models employ explicit intermediaries. BrgSA \cite{lai2025bridged} introduces an LLM-summarized cross-modal knowledge bank as a shared semantic anchor, while ViSD-Boost \cite{cao2025boosting} addresses the gap by modeling normal-appearance distributions via a vector-quantized variational autoencoder (VQ-VAE), deliberately amplifying pathological deviations before projecting them into the cross-modal space.

In summary, vision–language pre-training bridges the semantic gap between volumetric perception and clinical language. The field is moving from global volume-report matching toward alignment mechanisms that preserve disease concepts, anatomical regions, lesions, and report hierarchy. These text-aligned feature spaces provide the semantic substrate for later MLLMs to support language-mediated interpretation.

\subsection{Multimodal Large Language Models}
\label{sec:mllm}

Building on SSL backbones and vision–language pre-training, volumetric radiology MLLMs make volumetric representations available for interactive interpretation. The goal is no longer limited to retrieval or classification in an embedding space; it extends to report generation, multi-turn question answering, abnormality localization, uncertainty communication, and instruction following. We therefore treat this layer as an integration problem. The central design question is how a CT, MRI, or PET study is transformed from a dense volumetric representation into visual tokens or cross-attended states that a generative backbone can condition on. This question has four linked components: the vision encoder defines the dimensionality and content of the visual tokens; the vision–language interface projects, resamples, fuses, or routes those tokens; the generative backbone supplies language generation, instruction following, and context handling; and the training strategy determines which parts of the pipeline are pretrained, frozen, aligned, or task-optimized.

To orient the discussion, Table~\ref{tab:mllm_radiology_architectures} summarizes representative medical MLLMs by input dimensionality, visual backbone, generative backbone or base model, parameter scale, training recipe, and whether an explicit vision-language projector is trained. The table is limited to MLLM systems. It uses input dimensionality to compare how images or volumes are presented to the language model, not to provide a complete taxonomy of SSL or vision–language pre-training encoders. The comparison highlights two broad trends. First, explicit interface learning becomes increasingly important once full volumetric information must be compressed into an LLM-compatible visual token stream. Second, native 3D systems remain more modality-specific than many 2D medical MLLMs because CT, MRI, and PET studies impose heavier demands on token budget, spatial grounding, and instruction data.

\begin{table*}[!htb]
\centering
\footnotesize
\setlength{\tabcolsep}{2.2pt}
\renewcommand{\arraystretch}{1.06}
\caption{Representative medical MLLMs relevant to volumetric radiology, grouped by visual architecture. ``Visual input'' reports the representation presented to the vision pathway after stated preprocessing; a slice sequence is an ordered stack of 2D slices rather than a native 3D tensor. PT: pre-training; SFT: supervised fine-tuning; RL: reinforcement learning; Distill: knowledge distillation. In ``Proj.'', \cmark{}, \xmark{}, and \pmark{} denote an explicit, absent, or unspecified visual--language interface, respectively; ``--'' denotes unavailable or not applicable.}
\label{tab:mllm_radiology_architectures}
\begin{adjustbox}{max width=\textwidth}
\begin{tabularx}{\textwidth}{@{}L{2.2cm} C{1.85cm} >{\raggedright\arraybackslash}X >{\raggedright\arraybackslash}X C{1.4cm} C{1.3cm} C{0.7cm} L{1.8cm}@{}}
\toprule
\textbf{Model} & \textbf{Visual input} & \textbf{Vision encoder} & \textbf{Gen. backbone} & \textbf{Params} & \textbf{Training} & \textbf{Proj.} & \textbf{Imaging scope} \\
\midrule
\multicolumn{8}{@{}l}{\textit{2D-native medical MLLMs}} \\
\methodcell{BiomedGPT}{zhang2024generalist} & 2D image & VQGAN & BERT & \makecell[c]{33M/93M\\182M} & PT, SFT & \xmark & General medical \\
\methodcell{MedVInT}{zhang2023pmc} & 2D image & ResNet-50 & PMC-LLaMA / PMC-LLaMA-ENC & 7B & PT, SFT & \cmark & General medical \\
\methodcell{LLaVA-Med}{li2023llava} & 2D image & CLIP ViT-L/14 & LLaVA & 7B/13B & PT, SFT & \cmark & General medical \\
\methodcell{Med-Flamingo}{moor2023med} & 2D image & CLIP ViT-L/14 & OpenFlamingo & 8.3B & PT, SFT & \cmark & General medical \\
\methodcell{Med-PaLM M}{tu2024towards} & 2D image & ViT-4B/22B & PaLM-E & \makecell[c]{12B/84B\\562B} & SFT & \pmark & General medical \\
\methodcell{Qilin-Med-VL}{liu2023qilin} & 2D image & CLIP ViT-L/14-336 & Chinese-LLaMA2-13B-Chat & 13B & PT, SFT & \cmark & General medical \\
\methodcell{HuatuoGPT-Vision}{chen2024towards} & 2D image & CLIP-Large-336 & LLaVA-v1.5 recipe / Yi-1.5-34B & 8B/34B & PT, SFT & \cmark & General medical \\
\methodcell{HealthGPT}{lin2025healthgpt} & 2D image & CLIP-L/14 & Phi-3-mini / Phi-4 & 3.8B/14B & PT, SFT & \cmark & General medical \\
\methodcell{Lingshu}{xu2025lingshu} & 2D image & Qwen2.5-VL ViT & Qwen2.5-VL-7B/32B-Instruct & 7B/32B & PT, SFT, RL & \cmark & General medical \\
\methodcell{QoQ-Med}{dai2025qoq} & 2D image & Qwen2.5-VL ViT + ECG-JEPA & Qwen2.5-VL & 7B/32B & SFT, RL & \cmark & General medical + ECG \\
\methodcell{MedGemma}{sellergren2025medgemma} & 2D image & MedSigLIP & Gemma 3 & 4B/27B & PT, RL, Distill & \cmark & General medical \\
\methodcell{OctoMed}{ossowski2025octomed} & 2D image & Qwen2.5-VL ViT & Qwen2.5-VL-7B-Instruct & 7B & SFT & \cmark & General medical \\
\midrule
\multicolumn{8}{@{}l}{\textit{Hybrid 2D/3D architectures}} \\
\methodcell{RadFM}{wu2025towards} & \makecell[c]{2D image /\\3D volume} & 3D ViT + Perceiver & MedLLaMA-13B & 14B & PT, SFT & \cmark & General radiology \\
\methodcell{Med-2E3}{shi2025med} & 3D volume & M3D-CLIP + SigLIP & Phi-3-mini & 3.8B & PT, SFT & \cmark & CT \\
\methodcell{Hulu-Med}{jiang2025hulu} & \makecell[c]{2D image /\\slice sequence} & SigLIP-NaViT & Qwen2.5/\allowbreak Qwen3 Instruct & \makecell[c]{4B/7B/8B\\14B/32B} & PT, SFT & \cmark & General medical \\
\methodcell{MedGemma-1.5}{sellergren2026medgemma} & \makecell[c]{2D image /\\slice sequence} & MedSigLIP & MedGemma / Gemma 3 & 4B & PT, SFT, RL, Distill & \cmark & General medical \\
\methodcell{Photon}{fang2026photon} & \makecell[c]{2D image /\\3D volume} & 3D ViT + Qwen2.5-VL ViT & Qwen2.5-VL & 3B/7B & PT, SFT & \cmark & CT \\
\methodcell{OmniCT}{lin2026omnict} & \makecell[c]{2D image /\\3D volume} & SigLIP + SCE/OSE modules & Qwen2.5 & 3B/7B & PT, SFT & \cmark & CT \\
\methodcell{CT-Instruct}{lei2026versatile} & 3D volume & Hybrid ResNet-ViT & Qwen2-VL-7B-Instruct & 7B & PT, SFT & \cmark & CT \\
\methodcell{UniReason-Med}{chen2026unireason} & \makecell[c]{2D image /\\slice sequence} & frozen Qwen2.5-VL vision tower & Qwen2.5-VL-7B-Instruct & 7B & SFT, RL & \cmark & General medical + CT \\
\methodcell{MedM-VL}{shi2025medm} & \makecell[c]{2D image /\\3D volume} & SigLIP / M3D-CLIP & Qwen2.5-3B-Instruct & 3B & PT, SFT & \cmark & General medical + CT \\
\methodcell{RadSight}{liu2026radsight} & \makecell[c]{2D image /\\3D volume} & SigLIP-NaViT + Radar (PlainConvUNet) & Qwen3-VL & 4B/8B & PT, SFT & \cmark & General medical + CT \\
\methodcell{ClinFusion}{yuan2026clinfusion} & \makecell[c]{2D image /\\3D volume} & Qwen ViT + DINOv2 + ConvNeXt + PE-3D & Qwen3-VL & 8B/32B & PT, SFT & \cmark & General medical \\
\methodcell{MedReCo-VLM}{zhang2026vision} & \makecell[c]{2D/3D\\image pair} & modality-aware MoE ViT & Qwen2.5-7B-Instruct & 7B & PT, SFT & \cmark & General radiology \\
\bottomrule
\end{tabularx}
\end{adjustbox}
\end{table*}

\begin{table*}[!htb]
\ContinuedFloat
\centering
\footnotesize
\setlength{\tabcolsep}{2.2pt}
\renewcommand{\arraystretch}{1.12}
\caption{Representative medical MLLMs relevant to volumetric radiology (continued).}
\begin{adjustbox}{max width=\textwidth}
\begin{tabularx}{\textwidth}{@{}L{2.2cm} C{1.85cm} >{\raggedright\arraybackslash}X >{\raggedright\arraybackslash}X C{1.4cm} C{1.3cm} C{0.7cm} L{1.8cm}@{}}
\toprule
\textbf{Model} & \textbf{Visual input} & \textbf{Vision encoder} & \textbf{Gen. backbone} & \textbf{Params} & \textbf{Training} & \textbf{Proj.} & \textbf{Imaging scope} \\
\midrule
\multicolumn{8}{@{}l}{\textit{Native 3D radiology MLLMs}} \\
\methodcell{CT2Rep}{hamamci2024ct2rep} & 3D volume & 3D ViT & -- & -- & SFT & \xmark & CT \\
\methodcell{Dia-LLaMA}{chen2025dia} & 3D volume & 3D ViT + Perceiver & LLaMA2-7B & 7B & SFT & \cmark & CT \\
\methodcell{M$^3$D-LaMed}{bai2024m3d} & 3D volume & 3D ViT + Perceiver & LLaMA-2-7B & 6.7B & PT, SFT & \cmark & General radiology \\
\methodcell{Merlin}{blankemeier2026merlin} & 3D volume & I3D ResNet152 & RadLlama-7B & 7B & PT, SFT & \cmark & CT \\
\methodcell{BrainGPT}{li2024towards} & 3D volume & frozen CLIP ViT-L/14 + Perceiver Resampler & Otter & 7B & SFT & \cmark & Brain CT \\
\methodcell{3D-CT-GPT}{chen20243d} & 3D volume & CT-ViT from CT-CLIP & Vicuna-7B & 7B & PT, SFT & \cmark & CT \\
\methodcell{E3D-GPT}{lai2024e3d} & 3D volume & 3D MAE ViT-base + 3D convolutional projector & Vicuna-7B & 7B & PT, SFT & \cmark & CT \\
\methodcell{Reg2RG}{chen2025large} & 3D volume & RadFM 3D ViT + Perceiver + 3-layer ViT3D mask encoder & LLaMA2-7B & 7B & SFT & \cmark & CT \\
\methodcell{MS-VLM}{lee2024read} & 3D volume & DINO ViT-B/16 + Z-former & Vicuna-7B-v1.5 & 7B & PT, SFT & \cmark & CT \\
\methodcell{MEPNet}{zhang2025mepnet} & 3D volume & ResNet101 & LLaMA3-8B & 8B & SFT & \cmark & Brain CT \\
\methodcell{Med3DVLM}{xin2025med3dvlm} & 3D volume & DCFormer-S + SigLIP & Qwen2.5-7B-Instruct & 7.6B & PT, SFT & \cmark & General radiology \\
\methodcell{HSENet}{shi2025hsenet} & 3D volume & 3D-ViT + 2E3-ViT & Phi-4-4B-Instruct & 4B & PT, SFT & \cmark & CT \\
\methodcell{MedRegion-CT}{kyung2025medregion} & 3D volume & RAD-DINO (ViT-B) + MAIRA-SEG mask extractor & LLaMA3-8B & 8B & PT, SFT & \cmark & CT \\
\methodcell{mpLLM}{vepa2025multimodal} & 3D volume & M3D-CLIP ViT & Phi-3-Mini-4K-Instruct & 3.8B & SFT & \cmark & Brain MRI \\
\methodcell{PETAR}{maqbool2025petar} & 3D volume & M3D-CLIP ViT with PET/CT/mask projectors & M3D / Phi-3-4B & 4B & PT, SFT & \cmark & PET/CT \\
\methodcell{PETRG-3D}{jiao2025vision} & 3D volume & RadFM 3D ViT + Perceiver & Qwen3-8B & 8B & SFT & \cmark & PET/CT \\
\methodcell{Brain3D}{barone2026brain3d} & 3D volume & 3D-ViT & MedGemma-1.5-4B-IT & 4B & PT, SFT & \cmark & Brain MRI \\
\methodcell{Med3D-R1}{lai2026med3d} & 3D volume & 3D CLIP ViT & Qwen2.5-3B & 3B & PT, SFT, RL & \cmark & CT \\
\methodcell{RAD3D-Prefix}{sharma2026rad3dprefix} & 3D volume & CT-CLIP / CT-ViT + anomaly logits & LLaMA-3.2-1B & 1B & SFT & \cmark & CT \\
\methodcell{MRI2Rep}{li2026mri2rep} & 3D volume & 3D CNN + visual Transformer & AR Transformer decoder & -- & PT, SFT & \xmark & Liver MRI \\
\bottomrule
\end{tabularx}
\end{adjustbox}
\end{table*}
\FloatBarrier

\subsubsection{Vision Encoders}

Visual encoder dimensionality is a cross-cutting design axis in volumetric radiology AI. The same 2D, native 3D, and hybrid 2D/3D choices appear in SSL and vision–language pre-training, but their role changes in an MLLM. They determine the visual token sequence or visual states that the language model can condition on. The vision encoder is therefore the point where the representation-learning literature in Sections \ref{sec:VSSRL} and \ref{sec:VLA} enters the MLLM framework most directly. An LLM can only reason over the visual tokens it receives: if the encoder has not learned volumetric continuity, through-plane relations may be lost; if the encoder has not been aligned with clinical language, the interface must learn the semantic bridge from scarce instruction data. The models in Table~\ref{tab:mllm_radiology_architectures} can therefore be read as different answers to the same question: should the visual stream inherit a mature 2D biomedical vision–language pre-training prior, a self-supervised 3D anatomical prior, a 3D vision-language prior, or a hybrid combination of these sources?

\noindent\textbf{2D biomedical and generalist encoders.}
Early medical MLLMs mainly adapted the general large vision–language model (LVLM) recipe to biomedical images: a 2D visual encoder, a lightweight connector, and an instruction-tuned language model. BiomedGPT \cite{zhang2024generalist} uses a VQGAN-style visual tokenizer and unified image-text token modeling, while MedVInT \cite{zhang2023pmc}, LLaVA-Med \cite{li2023llava}, and Med-Flamingo \cite{moor2023med} establish the more common pattern of connecting 2D visual features to a medical or general multimodal decoder. These systems are useful baselines because they show how biomedical vision-language priors can support medical VQA, dialogue, and report-style generation with limited architectural change.

For volumetric radiology, however, the 2D-native family is mainly relevant as a source of transferable semantic priors rather than as a sufficient solution. Recent systems based on stronger Qwen-series or MedGemma-style multimodal backbones \cite{bai2025qwen3,sellergren2025medgemma} improve high-resolution image understanding and instruction following, but they still primarily operate on images, selected views, or long slice sequences. Once a CT or MRI study is reduced to a few planar observations, through-plane continuity, lesion extent, and volumetric morphology can be weakened before the vision-language interfaces or LLM receives the visual tokens.

\noindent\textbf{Native 3D radiology encoders.}
Native 3D models accept the volume itself as the primary visual object. This is the most direct trajectory for volumetric radiology tasks in which findings depend on slice-to-slice continuity. CT2Rep \cite{hamamci2024ct2rep} is an early representative of this direction: it uses a 3D ViT \cite{dosovitskiy2020image} encoder for CT report generation, avoiding the information loss caused by key-slice selection. Although CT2Rep is not a conversational LLM system, it establishes the important premise that report generation can be conditioned on full-volume representations rather than 2D projections alone.

Several later systems explicitly connect volumetric self-supervised or vision-language priors to generative MLLMs. E3D-GPT \cite{lai2024e3d} first builds a self-supervised 3D foundation model and then uses 3D spatial aggregation before instruction tuning, making it a clear example of SSL-to-MLLM transfer. Dia-LLaMA \cite{chen2025dia} uses a pre-trained ViT3D and adds disease-aware attention with a prototype memory bank, so its encoder is not merely anatomical but abnormality sensitive. M$^3$D-LaMed \cite{bai2024m3d} constructs large-scale 3D image-text and instruction corpora, couples a 3D vision encoder with spatial token pooling, and extends the visual interface to positioning and segmentation. 3D-CT-GPT \cite{chen20243d} uses CT-ViT features derived from CT-CLIP-style pre-training, showing how the global volume-report alignment discussed in Section 3.2 can become the visual front-end of a generative model.

A related group expands the 3D encoder toward broader CT foundation capability. Merlin \cite{blankemeier2026merlin} uses an I3D ResNet-style CT encoder and maps its features to a radiology language model, emphasizing scalability and multi-task competence on CT. CT-CHAT \cite{hamamci2026generalist} starts from CT-CLIP-style volumetric features and uses attention pooling to build a chat-oriented CT assistant. Med3DVLM \cite{xin2025med3dvlm} combines DCFormer-S with SigLIP-derived semantic signals, so the encoder carries both efficient 3D spatial features and medically aligned visual semantics. HSENet \cite{shi2025hsenet} emphasizes local and global spatial information by combining a 3D-ViT branch with a 2D-enhanced branch, which is especially relevant when small findings must be preserved without losing study-level context.

Domain-specialized 3D systems show that the desired encoder depends strongly on modality and anatomy. BrainGPT \cite{li2024towards}, MEPNet \cite{zhang2025mepnet}, and Brain3D \cite{barone2026brain3d} focus on brain CT or MRI, where subtle structural changes and lesion distributions require a narrower but more specialized visual prior. mpLLM \cite{vepa2025multimodal} uses M3D-CLIP features for multiparametric brain MRI, so the encoder must handle T1, T1Gd, T2, and FLAIR as complementary sources of volumetric information rather than interchangeable channels. MRI2Rep \cite{li2026mri2rep} extends structured report generation to 3D liver MRI, reinforcing that volumetric MLLMs are beginning to specialize beyond chest CT. PETAR \cite{maqbool2025petar} and PETRG-3D \cite{jiao2025vision} extend 3D MLLMs to PET/CT, where the visual encoder has to preserve both metabolic uptake and anatomical localization. These models indicate that a generic 3D encoder is unlikely to be sufficient across all clinical settings; the encoder must reflect what the downstream clinical question treats as image-grounded information.

A further trend inside native 3D encoders is the shift from study-level features toward localized representation and spatial grounding. Reg2RG \cite{chen2025large} augments volumetric texture encoding with mask-derived geometric information so generated reports can be tied to regions. MS-VLM \cite{lee2024read} uses slice-level DINO features and a Z-former to retain cross-slice structure before language generation. MedRegion-CT \cite{kyung2025medregion} brings pseudo-mask and region tokens into the visual stream to reduce region-level hallucination, while MedVL-SAM2 \cite{xing2026medvl} links 3D visual encoding with segmentation-oriented prompting. Human-body-prior 3D MLLMs \cite{zeng2026enhancing} and discriminative-guided spatial grounding \cite{wang2026enhancing} pursue the same goal from different angles: the encoder should not collapse the scan into a single global embedding before clinically important locations have been protected. At the efficiency frontier, U-VLM \cite{shi2026u} injects hierarchical visual information into multiple decoder layers, and BTB3D \cite{hamamci2025better} redesigns volumetric tokenization itself so fewer but more informative 3D tokens reach the language model.

\noindent\textbf{Hybrid 2D/3D encoders.}
Hybrid encoders occupy a pragmatic middle ground between mature 2D vision–language pre-training systems and native 3D volumetric backbones. One strategy treats the volume as an ordered slice or video sequence. Med-Gemini \cite{yang2024advancing} reuses Gemini's video pathway by interpreting CT depth as a temporal dimension, which lets the model benefit from a powerful pre-existing multimodal stack. Hulu-Med \cite{jiang2025hulu} and Fleming-VL \cite{shu2025fleming} similarly exploit strong 2D visual encoders, but add token-reduction and positional strategies so long medical image sequences remain manageable. MedGemma 1.5 \cite{sellergren2026medgemma} extends MedGemma to 3D by representing volumes as long axial slice sequences, showing that a medically aligned 2D encoder can be stretched toward 3D with careful context handling and training.

Efficiency-oriented hybrid methods make the same slice-sequence assumption but focus on selecting the right visual tokens. MedPruner \cite{liu2026medpruner} removes redundant slice and patch tokens hierarchically, which is important because adjacent CT slices often contain highly repetitive anatomy. Photon \cite{fang2026photon} goes further by learning instruction-conditioned token scheduling, so the retained visual tokens can vary with the clinical question rather than following a fixed sampling rule. The adaptation study of Yu et al. \cite{yu2026adapting} reinforces this point from an empirical angle: adapting 2D encoders to 3D is attractive under limited data, but it works best when the model has explicit mechanisms for preserving volumetric consistency. UniReason-Med \cite{chen2026unireason} makes the transfer problem more explicit by aligning 2D images and slice-serialized 3D volumes through a shared grounded reasoning interface with region-token injection.

A second hybrid strategy keeps an explicit 3D branch together with slice-level information. RadFM \cite{wu2025towards} unifies 2D and 3D scans through visually conditioned generative pre-training, using a 3D-compatible visual stream so both planar and volumetric inputs can be fed to the same MLLM. Med-2E3 \cite{shi2025med} combines M3D-CLIP features with SigLIP slice features and uses text-guided inter-slice scoring, allowing the instruction to decide which 2D visual tokens should complement the 3D representation. CTInstruct \cite{lei2026versatile} uses a hybrid ResNet-ViT encoder to support diagnosis, segmentation, report generation, and multiple-choice reasoning with one CT-oriented backbone. MedM-VL \cite{shi2025medm} systematically compares 2D and 3D CT encoder choices for medical LVLMs, while OmniCT \cite{lin2026omnict} introduces unified slice-volume encoding with spatial consistency and organ-level semantic enhancement. RadSight \cite{liu2026radsight} uses dedicated SigLIP-NaViT and Radar/PlainConvUNet pathways so 2D images and 3D CT volumes retain their native spatial structure before entering a shared Qwen3-VL interface. ClinFusion \cite{yuan2026clinfusion} instead enriches Qwen ViT tokens with DINOv2, ConvNeXt, and PE-3D features through cascaded spatial-aware locality fusion, including a 2D-anchored depth-aware operator for volumetric inputs. MedReCo-VLM \cite{zhang2026vision} combines per-slice and through-depth attention in a modality-aware MoE ViT and applies the shared encoder to paired studies, supporting entity-conditioned comparison across radiography, CT, MRI, and ultrasound. Across these hybrid designs, the encoder balances 2D semantic maturity with 3D anatomical completeness.

Across these families, encoder paradigms differ not only in dimensionality but also in the priors they inherit, the information they preserve, and the bottlenecks they introduce. Table~\ref{tab:vision_encoder_synthesis} condenses these trade-offs at the family level. Unlike Table~\ref{tab:mllm_radiology_architectures}, which inventories model architectures, this comparison links each encoder paradigm to its main limitation and resulting design implication.

\begin{table*}[!b]
\centering
\footnotesize
\setlength{\tabcolsep}{2.5pt}
\renewcommand{\arraystretch}{1.14}
\caption{Comparison of visual encoder paradigms for volumetric radiology MLLMs, ordered by increasing native volumetric integration. The rows summarize recurring architectural trade-offs reflected in representative systems rather than a universal performance ranking.}
\label{tab:vision_encoder_synthesis}
\begin{tabularx}{\textwidth}{@{}L{0.14\textwidth} L{0.18\textwidth} L{0.18\textwidth} L{0.19\textwidth} >{\raggedright\arraybackslash}X@{}}
\toprule
\textbf{Encoder paradigm} & \textbf{Dominant prior} & \textbf{Visual evidence emphasized} & \textbf{Characteristic bottleneck} & \textbf{Key design implication} \\
\midrule
\textbf{2D-native encoders}
& Biomedical image--text alignment and multimodal instruction following
& In-plane appearance and medically aligned semantics
& No native modeling of through-plane continuity
& Natural fit for 2D or selected-view tasks; volumetric use requires explicit cross-slice aggregation \\
\midrule
\textbf{Slice-sequence encoders}
& 2D image--language representations extended across ordered slices
& Ordered slice content with broader volumetric coverage
& Long token sequences; cross-slice structure depends on positional modeling and token reduction
& Preserve slice order and select task-relevant tokens before aggressive compression \\
\midrule
\textbf{Hybrid 2D/3D encoders}
& 2D semantic features combined with 3D spatial features
& Fine in-plane detail and global volumetric context
& Explicit fusion and alignment across heterogeneous feature streams
& Use task-conditioned fusion when both fine detail and 3D context are required \\
\midrule
\textbf{Native 3D encoders}
& Volumetric structure learned through 3D SSL or image--text alignment
& Through-plane continuity, lesion extent, and global anatomy
& High-dimensional inputs, limited 3D data, and costly token processing
& Prioritize when spatial continuity or lesion extent is central; compress without discarding spatial structure \\
\bottomrule
\end{tabularx}
\end{table*}

Taken together, these comparisons indicate that no encoder paradigm is uniformly optimal across all tasks. Nevertheless, increasing native volumetric integration raises the ceiling for preserving through-plane and full-volume evidence, and should be prioritized when such evidence is required by the intended clinical claim (Table~\ref{tab:vision_encoder_synthesis}). Three principles emerge. First, tasks that depend on anatomy, spatial continuity, or subtle lesion morphology are likely to benefit from encoders that learn volumetric priors, for example through volumetric SSL or 3D image--text pre-training. Second, tasks involving open-vocabulary recognition, retrieval, report generation, or spatially grounded language additionally benefit from medically aligned semantic priors established through vision--language pre-training. Third, regardless of encoder family, compression should be task-adaptive and should not discard local evidence before it can be integrated with organ-level context and global study semantics. The vision encoder therefore does more than initialize the MLLM pipeline: it sets an upper bound on the visual evidence that subsequent interfaces can make available to the language model.

\subsubsection{Vision-Language Interfaces: Projection, Resampling, and Fusion}
After the vision encoder produces image or volume tokens, the next question is how these tokens are made available to the generative backbone. We use vision–language interface as the umbrella term for modules that project, resample, fuse, prune, route, or inject visual states into the generative context.  The term adapter remains useful for linear or multilayer perceptron (MLP) projectors, but it does not fully capture the broader functions of current 3D visual--language interfaces. In volumetric radiology, this interface is the bottleneck between dense volumetric information and the LLM token stream. Its role is broader than dimensional matching: it decides how visual tokens are selected, compressed, ordered, grounded, and inserted into language space. The models in Table~\ref{tab:mllm_radiology_architectures} therefore expose a design spectrum. At one end, the connector is minimal. BiomedGPT \cite{zhang2024generalist} avoids a conventional continuous projector by converting images into discrete VQGAN visual tokens and modeling them together with text tokens. LLaVA-style medical systems such as LLaVA-Med \cite{li2023llava}, Qilin-Med-VL \cite{liu2023qilin}, and R2GenGPT \cite{wang2023r2gengpt} mainly use linear or shallow MLP projectors to map 2D visual features into the LLM embedding space. MedVInT \cite{zhang2023pmc} explicitly compares MLP and transformer-decoder projection variants, while Med-Flamingo \cite{moor2023med} follows the Flamingo design with a perceiver resampler and gated cross-attention layers. These 2D designs are best viewed as interface baselines for volumetric radiology: they solve vision-language dimensional alignment, but not the more demanding problem of selecting and preserving clinically relevant 3D information.

For volumetric radiology, this simple projector pattern is often insufficient because the interface must reduce thousands of volumetric tokens without erasing small or spatially sparse findings. One direct solution is query-based compression. RadFM \cite{wu2025towards} pads 2D inputs into a 3D-compatible representation, uses a shared 3D ViT, and applies a perceiver with learned latent queries so variable-size 2D/3D scans become a fixed visual prefix. Dia-LLaMA \cite{chen2025dia} similarly projects ViT3D patch features through a perceiver before LLM insertion, but further attaches disease-aware attention and a prototype memory bank so the visual prefix is accompanied by abnormality-sensitive diagnostic prompts. M$^3$D-LaMed \cite{bai2024m3d} makes the compression spatially explicit: 3D ViT outputs are reconstructed into a volume-like grid, pooled in 3D space, and then passed through linear or MLP layers, reducing the visual input from dense volumetric patches to LLM-sized tokens while retaining coarse anatomical layout. BrainGPT \cite{li2024towards} adapts the Otter/Flamingo paradigm to brain CT with a trainable Perceiver Resampler and cross-gated attention layers, and CT-CHAT \cite{hamamci2026generalist} uses attention pooling with learned latent queries plus an MLP multimodal projector to connect CT-CLIP features to the LLM. These systems share the same interface logic: learned queries serve as a controlled interface between arbitrarily long volumetric information and the fixed context budget of an autoregressive decoder.

A second family keeps the connector lightweight but makes the spatial path before or inside the connector more structured. CT2Rep \cite{hamamci2024ct2rep} is not an LLM-prefix model and therefore has no explicit vision-language projector; its transformer decoder consumes 3D encoded states directly, and CT2RepLong adds cross-attention over previous volumes and reports for longitudinal fusion. Merlin \cite{blankemeier2026merlin} takes a more conventional MLLM route, mapping I3D ResNet features to RadLlama with a single linear adapter and low-rank adaptation (LoRA) tuning. 3D-CT-GPT \cite{chen20243d} similarly uses CT-ViT features, 3D average pooling, and a trainable linear projection, with later experiments comparing simple linear projection and MLP alternatives. RAD3D-Prefix \cite{sharma2026rad3dprefix} revisits the same adaptation problem by studying how diagnostic priors and scaling choices affect LLM adaptation for 3D CT report generation. E3D-GPT \cite{lai2024e3d} argues that a 3D adapter should not behave like a generic Q-Former: its 3D convolutional perceiver merges and projects features while preserving local spatial neighborhoods. HSENet \cite{shi2025hsenet} develops this principle further with twin spatial packers for global and local 3D features; its Voxel2Point cross-attention aggregates high-resolution voxel features into centroid tokens, then uses two-layer MLPs to reach the LLM latent dimension. Brain3D \cite{barone2026brain3d} follows a compact prefix strategy for brain MRI, compressing inflated 3D transformer tokens to a small fixed set, projecting them through a two-layer MLP, and controlling visual conditioning with a learnable scalar gate.

Hybrid 2D/3D models turn the interface into a fusion module rather than a single projection. Med-Gemini \cite{yang2024advancing} reuses Gemini's video pathway, relying on the model's native long-context multimodal interface rather than a separately specified visual--language projector. Med-2E3 \cite{shi2025med} uses separate 3D and 2D branches with MLP-based connectors, then computes text-guided inter-slice scores so the 2D features most relevant to the instruction enhance the 3D feature sequence. MedM-VL \cite{shi2025medm} studies the connector choice itself: for 3D inputs encoded slice by slice, it compares cross-attention compression to fixed-length tokens against average pooling, and shows that even a simple linear connector can remain competitive when the encoder and training data are well chosen. CTInstruct \cite{lei2026versatile} uses a lightweight linear bridger from hybrid ResNet-ViT features to the text latent space, but deliberately preserves a dynamic token count determined by volumetric patch decomposition instead of forcing every scan into a fixed perceiver prefix. OmniCT \cite{lin2026omnict} introduces a more explicit slice-volume interface: volumetric slice composition and tri-axial positional encodings build unified slice/volume tokens, while its mixture-of-experts (MoE) Hybrid Projection contains slice-specific, volume-specific, and shared projection matrices. This lets projection behavior learned from 2D slices transfer to 3D volumes while still allowing modality-dependent transformations. RadSight \cite{liu2026radsight} and MedReCo-VLM \cite{zhang2026vision} retain lightweight two-layer MLP projectors, but apply them to modality-specific or paired entity-aware features, respectively. ClinFusion \cite{yuan2026clinfusion} makes fusion itself the interface: its CaSL operator incrementally injects local specialist features into the language-aligned Qwen ViT stream and extends the same mechanism across volumetric depth.

A third line makes the interface region- or mask-aware. Reg2RG \cite{chen2025large} reuses RadFM's ViT3D and Perceiver adapter for texture features, but pairs it with a lightweight ViT3D mask encoder so local texture tokens, geometric mask tokens, and global context can be aligned to region-level report descriptions. MedRegion-CT \cite{kyung2025medregion} combines R$^2$ token pooling with a mask-driven visual extractor: pseudo-masks generate mask tokens, spatial tokens, and patient-specific attributes that are injected with global CT tokens to reduce region-level hallucination. MEPNet \cite{zhang2025mepnet} uses a visual adaptor to map scan features to the LLM space, then adds knowledge-driven joint attention to mine entity-specific visual embeddings and learning-status prompts, making the interface partly semantic and entity balanced. PETAR \cite{maqbool2025petar} extends this principle to PET/CT by jointly encoding PET, CT, and lesion masks with modality-specific projectors and mask-conditioned PET embeddings; focal crops and mask-aware visual tokens force the language model to describe the highlighted lesion rather than the entire scan without lesion conditioning. PETRG-3D \cite{jiao2025vision} instead uses dual PET and CT streams initialized from RadFM-style 3D encoders, retrains Perceiver samplers for each modality, projects the compressed tokens linearly into the LLM space, and combines them with hospital-style templates. MedVL-SAM2 \cite{xing2026medvl} illustrates the segmentation-oriented version of the same idea, using an MLP-Mixer projection layer to compress 3D visual tokens and a generated \texttt{[SEG]} token to drive prompt-based 3D segmentation.

Recent interfaces are also becoming schedulers and routers. Med3DVLM \cite{xin2025med3dvlm} replaces a plain MLP with a multi-scale MLP-Mixer projector that mixes token and channel dimensions across low- and high-level DCFormer features, improving the transfer of both fine spatial and semantic information. mpLLM \cite{vepa2025multimodal} designs a prompt-conditioned hierarchical MoE projector for multiparametric brain MRI, with routers over modality-level and token-level projection experts so the connector can choose different transformations for T1, T1Gd, T2, and FLAIR according to the question. Med3D-R1 \cite{lai2026med3d} adds adaptive weighted pooling and a residual anchor mapping module, blending projected image tokens with a fixed text-space anchor through token-specific gates before reinforcement learning encourages clinically consistent reasoning. Efficiency-oriented models reinterpret the interface as token budgeting. Hulu-Med \cite{jiang2025hulu} uses a two-layer MLP projector but adds medical-aware token reduction for 3D and video inputs. Fleming-VL \cite{shu2025fleming} combines pixel unshuffle, a two-layer MLP projector, and variable visual position encoding so long multi-image sequences occupy less positional space. MedPruner \cite{liu2026medpruner} prunes redundant slice and patch tokens hierarchically, Photon \cite{fang2026photon} learns instruction-conditioned token scheduling with surrogate gradient propagation so retained tokens vary by question, and BTB3D \cite{hamamci2025better} addresses the same bottleneck at tokenization time by producing compact volumetric tokens before LLM injection. U-VLM \cite{shi2026u} provides another route by injecting hierarchical visual information into multiple decoder layers rather than relying on one front-loaded prefix. Across these designs, the interface becomes an active control point: it determines not only computational cost, but also whether the LLM receives global study context, local lesion information, modality-specific cues, and spatial provenance needed for a grounded radiological conclusion.

\subsubsection{Generative Backbones: Text Decoders and Multimodal Foundation Stacks}

After visual states have been encoded and routed through the vision-language interfaces, the next design choice is the generation-side model that receives them. We use the term generative backbone for this receiving component. It may be a standalone text decoder, a report-generation decoder, or an inherited VLM/MLLM stack rather than a text-only LLM. This distinction matters because pretrained multimodal stacks already carry assumptions about visual token format, context handling, and cross-modal conditioning, whereas text-decoder-centric systems depend more heavily on the 3D encoder and interface for visual grounding. We therefore organize this subsection by the type of generative backbone inherited by each system.

\noindent\textbf{Report-generation and unified token-modeling decoders.}
Some systems use dedicated report-generation decoders or unified token-modeling frameworks rather than a conversational, instruction-tuned LLM as the generative backbone. BiomedGPT \cite{zhang2024generalist} uses a BERT-like generative modeling framework \cite{devlin2019bert}, making it closer to unified biomedical token modeling than to instruction-following dialogue. CT2Rep \cite{hamamci2024ct2rep} also follows a report-generation route: its transformer decoder generates CT reports directly from 3D encoded states rather than from a conversational LLM prefix. These systems remain important because they show that full-volume radiology generation can be formulated before the adoption of current instruction-tuned MLLM stacks. MedVInT \cite{zhang2023pmc} is a transitional case, moving from task-specific biomedical decoders toward LLaMA-series medical language backbones \cite{wu2024pmc}.

\noindent\textbf{Text-decoder-centric MLLMs.}
A large group of medical MLLMs attaches visual states to a text-oriented instruction model. The LLaMA-series \cite{touvron2023llama,touvron2023llama2,grattafiori2024llama} and related Vicuna-style derivatives became common targets for early medical and volumetric radiology systems, including Dia-LLaMA \cite{chen2025dia}, M$^3$D-LaMed \cite{bai2024m3d}, 3D-CT-GPT \cite{chen20243d}, E3D-GPT \cite{lai2024e3d}, Reg2RG \cite{chen2025large}, MS-VLM \cite{lee2024read}, MEPNet \cite{zhang2025mepnet}, and MedRegion-CT \cite{kyung2025medregion}. Medical variants of this lineage, such as MedLLaMA in RadFM \cite{wu2025towards} and RadLlama in Merlin \cite{blankemeier2026merlin}, provide domain-tuned language priors and report style, but visual factuality still depends on the encoder-interface pipeline.

More recent text-decoder choices reflect the need for stronger instruction following, lower deployment cost, or domain-specific adaptation. Qwen-series text decoders \cite{bai2025qwen3} are used in several hybrid or native 3D systems, including MedM-VL \cite{shi2025medm}, Med3DVLM \cite{xin2025med3dvlm}, Hulu-Med \cite{jiang2025hulu}, OmniCT \cite{lin2026omnict}, Med3D-R1 \cite{lai2026med3d}, PETRG-3D \cite{jiao2025vision}, RadSight \cite{liu2026radsight}, ClinFusion \cite{yuan2026clinfusion}, and MedReCo-VLM \cite{zhang2026vision}. Phi-series models \cite{abdin2024phi} support compact systems such as HealthGPT \cite{lin2025healthgpt}, Med-2E3 \cite{shi2025med}, HSENet \cite{shi2025hsenet}, mpLLM \cite{vepa2025multimodal}, and PETAR \cite{maqbool2025petar}. In these text-decoder-centric designs, the generative backbone supplies fluency, instruction following, and medical terminology handling. It does not by itself solve volumetric grounding.

\noindent\textbf{Pretrained VLM/MLLM foundation stacks.}
Another group inherits a pretrained multimodal stack rather than a standalone text decoder. LLaVA-style systems such as LLaVA-Med \cite{li2023llava}, Qilin-Med-VL \cite{liu2023qilin}, and HuatuoGPT-Vision \cite{chen2024towards} adapt an existing visual-instruction architecture to medical images. Med-Flamingo \cite{moor2023med} inherits OpenFlamingo-style gated cross-attention \cite{awadalla2023openflamingo}, while BrainGPT \cite{li2024towards} adapts the Otter/Flamingo paradigm \cite{li2025otter} to brain CT report generation. Med-PaLM M \cite{tu2024towards} represents the PaLM-E route \cite{anil2023palm}, where the base model already contains broad multimodal instruction-tuning priors.

Qwen-VL and MedGemma-style systems illustrate the same issue in newer foundation stacks. Lingshu \cite{xu2025lingshu}, QoQ-Med \cite{dai2025qoq}, OctoMed \cite{ossowski2025octomed}, Photon \cite{fang2026photon}, and CTInstruct \cite{lei2026versatile} build on Qwen-VL or Qwen2.5-VL-style multimodal backbones. MedGemma \cite{sellergren2025medgemma}, MedGemma 1.5 \cite{sellergren2026medgemma}, and Brain3D \cite{barone2026brain3d} similarly inherit medically tuned multimodal or vision-language foundation stacks based on Gemma-family models \cite{team2024gemma}. These systems should not be described as using only an LLM backbone. They inherit multimodal context interfaces developed mainly for 2D images or video-like inputs, and volumetric radiology adaptation must make volumetric tokens compatible with those inherited assumptions.

Across these choices, parameter scale is insufficient as a standalone indicator of clinical usefulness because the three backbone paradigms inherit different capabilities and depend on different routes for visual conditioning. A more informative comparison asks what the backbone contributes, which volumetric information must be supplied by the encoder-interface pipeline, and how the resulting generation should be evaluated. Table~\ref{tab:generative_backbone_synthesis} summarizes these dependencies.

\begin{table*}[!b]
\centering
\footnotesize
\setlength{\tabcolsep}{3pt}
\renewcommand{\arraystretch}{1.14}
\caption{Comparison of generative backbone paradigms for volumetric radiology MLLMs. The rows summarize the capability inherited by each paradigm, its dependence on the encoder-interface pipeline, and the corresponding evaluation priority.}
\label{tab:generative_backbone_synthesis}
\begin{tabularx}{\textwidth}{@{}L{0.20\textwidth} L{0.22\textwidth} L{0.28\textwidth} >{\raggedright\arraybackslash}X@{}}
\toprule
\textbf{Generative backbone paradigm} & \textbf{Capability inherited} & \textbf{Volumetric dependency or limitation} & \textbf{Key evaluation implication} \\
\midrule
\textbf{Report-generation or unified token decoders}
& Task-specific structured report generation or joint visual--text token modeling
& Task-specific output scope; grounding depends on the encoded visual states delivered directly to the decoder
& Assess report completeness, factuality, and spatial fidelity for the intended task \\
\midrule
\textbf{Text-decoder-centric MLLMs}
& Language fluency, instruction following, and medical terminology
& Volumetric grounding remains external to the decoder and depends on encoder-interface delivery
& Assess grounded factuality and hallucination jointly with encoder-interface quality rather than parameter scale alone \\
\midrule
\textbf{Pretrained VLM/MLLM foundation stacks}
& Multimodal conditioning, visual-token interfaces, and broader context handling
& Inherited token-format and positional assumptions may be optimized for 2D images or video
& Verify 3D token compatibility, spatial provenance, and grounding after volumetric adaptation \\
\bottomrule
\end{tabularx}
\end{table*}

Taken together, generative backbones should be assessed as part of the complete encoder--interface--generation pipeline rather than in isolation (Table~\ref{tab:generative_backbone_synthesis}). Report-generation and unified token decoders should be judged by task-specific completeness and spatial fidelity; text decoders by whether fluent output remains grounded in the delivered 3D evidence; and pretrained multimodal stacks by whether their inherited interfaces remain compatible with volumetric tokens. Across all three paradigms, clinical usefulness depends on the alignment of spatial grounding, visual-token delivery, and the training objective. The clinical value of a generation-side model therefore depends on its ability to faithfully express the clinically relevant 3D evidence retained by the encoder-interface pipeline.

\subsubsection{Training Strategies}

Training strategies are discussed here according to the specific learning problem addressed by each stage: acquiring volumetric priors, constructing 3D supervision from reports and task labels, enforcing spatial grounding, managing the visual-token budget, and adapting or optimizing the model under limited medical data. This organization reflects a central constraint of volumetric radiology MLLMs: training must preserve clinically relevant volumetric information while converting it into a compact and clinically grounded language interface.

\noindent\textbf{From generic alignment to volumetric curricula.}
The training strategy of a volumetric radiology MLLM should not be understood as a simple extension of the 2D recipe. In 2D medical MLLMs, training often consists of aligning a mature image encoder to an LLM and then performing instruction tuning. In volumetric radiology, the central difficulty is different: the model must learn to preserve a complete volumetric study, compress it into a limited token budget, and still generate clinically grounded language from weak report-level supervision. As a result, the practical training pipeline becomes a curriculum over preservation of clinically relevant 3D information. The visual stream is first initialized with volumetric priors through SSL or 3D vision–language pre-training, the interface then learns how to expose visual tokens to the LLM, and instruction tuning finally teaches the model which parts of the scan should be verbalized, localized, or ignored for a given clinical question. ClinFusion \cite{yuan2026clinfusion} makes this curriculum explicit through five stages that progress from shallow 2D multi-encoder alignment to deep visual--language alignment, 2D instruction tuning, rapid 3D alignment, and joint 2D/3D instruction tuning.

\noindent\textbf{Volumetric initialization before instruction tuning.}
Most native 3D systems therefore begin with a stronger visual initialization than their 2D counterparts. The reason is not only data efficiency, but also stability: directly coupling raw CT, MRI, or PET volumes to an autoregressive decoder would force the model to learn anatomy, spatial continuity, modality appearance, token compression, and clinical language simultaneously. E3D-GPT \cite{lai2024e3d} makes this dependence explicit by first learning a self-supervised 3D foundation model before instruction tuning. M$^3$D-LaMed \cite{bai2024m3d} and 3D-CT-GPT \cite{chen20243d} rely on 3D vision-language or CT-CLIP-style pre-training so that the visual features already carry report-level semantics. Merlin \cite{blankemeier2026merlin}, CT-CHAT \cite{hamamci2026generalist}, and Med3DVLM \cite{xin2025med3dvlm} reflect the same pattern: the model is not trained to discover radiological perception from SFT alone; it imports a volumetric or semantically aligned encoder and then trains the vision-language interfaces around it. MedReCo-VLM \cite{zhang2026vision} provides a comparison-oriented variant: its modality-aware encoder is first trained with image--report alignment and entity-conditioned contrastive ranking, then frozen while a projector and Qwen2.5 decoder learn comparative generation from paired studies. This makes volumetric MLLM training a continuation of the representation-learning and vision--language alignment stages described in Sections \ref{sec:VSSRL} and \ref{sec:VLA} rather than an isolated final step.

\noindent\textbf{Volumetric instruction data construction.}
The second difference is the form of supervision. A 2D medical VQA example can often be treated as an image-question-answer triple, but a volumetric radiology case is usually paired with a study-level report that does not explicitly label every slice, organ, lesion, or negative finding. Training data must therefore be converted into instructions that preserve the structure of the scan. M$^3$D-LaMed \cite{bai2024m3d} scales 3D instruction-response data across multiple tasks, while CTInstruct \cite{lei2026versatile} organizes CT supervision around diagnosis, segmentation, report generation, and multiple-choice reasoning. CT2Rep \cite{hamamci2024ct2rep}, BrainGPT \cite{li2024towards}, MEPNet \cite{zhang2025mepnet}, Brain3D \cite{barone2026brain3d}, and PETRG-3D \cite{jiao2025vision} show a more report-centric version of this strategy, where the model learns to translate full-volume information into structured radiological language. Accordingly, instruction construction in 3D is not merely prompt formatting; it determines whether the supervision teaches global impressions only, or also laterality, anatomical location, lesion extent, modality-specific cues, and uncertainty.

\noindent\textbf{Grounding-aware multi-task training.}
Because study-level reports can be too coarse, many volumetric MLLMs introduce auxiliary supervision that forces generated language to remain connected to spatial grounding targets. This creates a training pattern that is less common in general 2D MLLMs: report generation is combined with classification, localization, segmentation, region description, or mask-conditioned reasoning. Reg2RG \cite{chen2025large} uses region and mask information so report text can be tied to local image-grounded information. MedRegion-CT \cite{kyung2025medregion} uses pseudo-mask-derived region tokens to reduce region-level hallucination. PETAR \cite{maqbool2025petar} trains with PET, CT, and lesion-mask information so the response focuses on the highlighted metabolic abnormality rather than the whole scan generically. U-VLM \cite{shi2026u} progressively supervises segmentation, classification, and report generation, while MedVL-SAM2 \cite{xing2026medvl} connects language reasoning to prompt-driven 3D segmentation. UniReason-Med \cite{chen2026unireason} further frames grounded reasoning as a shared interface across 2D images and slice-serialized 3D volumes. RadSight \cite{liu2026radsight} organizes this supervision into a four-stage evidence hierarchy, visual--language alignment, fine-grained attribute and spatial perception, clinical diagnosis, and report generation, so diagnostic interpretation is introduced only after lesion attributes and 2D/3D grounding have been learned explicitly. CLarGen \cite{mayelasserre2026templatecollapse} identifies template collapse as a failure mode of 3D CT report generation and separates clinical detection from language synthesis. These methods suggest that, for volumetric radiology, SFT should include grounding supervision rather than only response imitation: the model must learn not just the surface form of a plausible report, but where each statement comes from in the volume.

\noindent\textbf{Token-budget-aware training.}
A third distinction is that training must account for the cost of visual tokens. A 3D scan contains far more tokens than an LLM can consume, so token selection is part of the training task rather than a pure inference optimization. Hybrid slice-sequence systems such as Hulu-Med \cite{jiang2025hulu}, Fleming-VL \cite{shu2025fleming}, and MedGemma 1.5 \cite{sellergren2026medgemma} train on long image sequences or video-like representations to make 2D backbones usable for volumetric studies. Efficiency-oriented models make this more explicit: MedPruner \cite{liu2026medpruner} learns hierarchical slice and patch pruning, Photon \cite{fang2026photon} learns instruction-conditioned visual token scheduling, and BTB3D \cite{hamamci2025better} redesigns volumetric tokenization so compact tokens can be learned before LLM injection. Med-2E3 \cite{shi2025med} and mpLLM \cite{vepa2025multimodal} further show that token routing can be question- or modality-dependent. Thus, the 3D training objective asks a more constrained question than 2D alignment: which subset of visual tokens should survive for this instruction, and how should the model be penalized when the discarded information is critical for answering the clinical question?

\noindent\textbf{Parameter-efficient and reward-driven post-training.}
Finally, 3D medical instruction data remains limited and expensive for unrestricted end-to-end tuning in most settings. Many systems therefore freeze the vision encoder or LLM, train only the vision-language interfaces or lightweight adaptation modules, and add LoRA or other parameter-efficient components to control overfitting and preserve the language model's clinical fluency. This is especially important for domain-specialized settings such as brain MRI, multiparametric MRI, and PET, where the model must learn modality-specific information without degrading general instruction-following ability. Reward-driven post-training is still emerging, but it is conceptually important for 3D because some targets are more verifiable than open-ended dialogue. Med3D-R1 \cite{lai2026med3d} extends SFT with RL-style optimization for volumetric abnormality diagnosis, suggesting that rewards can be defined around diagnostic correctness, spatial grounding, response format, and clinical consistency. Recent reward designs make this visual constraint more explicit. TIF-GRPO \cite{lin2026tifgrpo} regulates anatomy-aware rewards for volumetric CT analysis, while E-MRL \cite{li2026emrl} trains diagnosis-localization-verification trajectories with cross-view consistency rewards. The value of RL-style post-training in volumetric radiology is therefore not simply to produce longer reasoning chains, but to penalize fluent statements that are unsupported by the volume and reward answers that remain anatomically and diagnostically grounded.

Overall, the distinguishing feature of volumetric MLLM training is not the mere presence of pre-training, SFT, or RL, but the way these stages are reorganized around volumetric information preservation and spatial grounding. A model must first acquire
3D anatomical and semantic priors, learn a compressed visual--language interface, and develop the ability to answer with an appropriate level of spatial detail before being optimized for correctness and grounding under clinical constraints. This makes training strategy inseparable from the encoder and interface choices discussed above: in volumetric radiology, what the
model learns is tightly constrained by the slices, regions, modalities, and masks that the training pipeline teaches it to preserve.

In summary, the model landscape for volumetric radiology is moving along two coupled axes. Representation learning is evolving from generic reconstruction and contrastive learning toward anatomy-aware, fine-grained, and knowledge-enhanced alignment. MLLM construction is evolving from simple 2D encoder--LLM coupling toward native or hybrid 3D tokenization, richer vision-language interfaces, stronger generative backbones, and grounding-aware post-training. ClinFusion's retrieval and specialist-tool extension~\cite{yuan2026clinfusion} illustrates the resulting bridge to agentic systems: Once volumetric information can be represented, compressed into visual tokens, spatially grounded, and discussed in language, the next system-level challenge is to orchestrate these capabilities into reliable multi-step clinical workflows.


\section{Agentic Systems in Volumetric Radiology}
\label{sec:agent}

The volumetric radiology MLLMs reviewed in Section~\ref{sec:mllm} enable language-mediated interpretation of volumetric imaging, but many remain single-pass input–output models: they encode a study, expose a compressed set of visual tokens to a generative backbone, and produce an output from fixed context. Agentic systems extend this design by embedding an LLM, MLLM, or VLM within a structured workflow that supports iterative evidence acquisition, planning, tool use, and context management. This relation is better understood as co-evolution rather than replacement: stronger volumetric representations expand the evidence available to agentic systems, whereas agentic workflows impose additional requirements for representation fidelity, provenance, and traceability. Section ~\ref{sec:foundation} focused on how volumetric representations are learned, aligned, and preserved; this section examines how they are dynamically accessed, supplemented, and coordinated across multiple stages of analysis.

Architecturally, these systems pursue multimodal reasoning through two complementary paths. Some augment MLLM perception directly with planning, memory, and iterative refinement. Others route perceptual tasks using the language model as an indirect multimodal controller that acquires volumetric information without an internal visual encoder. Both paths extend the multimodal reasoning paradigm of Section~\ref{sec:foundation} to workflow-level systems, and both share the same requirement: clinically relevant information must remain spatially grounded and traceable to their source, whether it originates from an internal encoder or an external tool call.



These characteristics motivate a functional taxonomy centered on the operations required for volumetric radiology analysis. Because clinically relevant evidence may be distributed across slices, series, phases, anatomical regions, measurements, and prior examinations, an agent must solve four linked problems: deciding what evidence to inspect or acquire; translating clinical intent into spatial or quantitative operations; retaining and updating evidence across analysis steps and time points; and coordinating intermediate outputs with tools, other agents, and human review. These problems correspond, respectively, to reasoning and planning, tool-augmented perception and grounded action, memory and dynamic context management, and workflow interaction and multi-agent collaboration.

\begin{figure*}[t]
    \centering
    \includegraphics[width=0.9\textwidth]{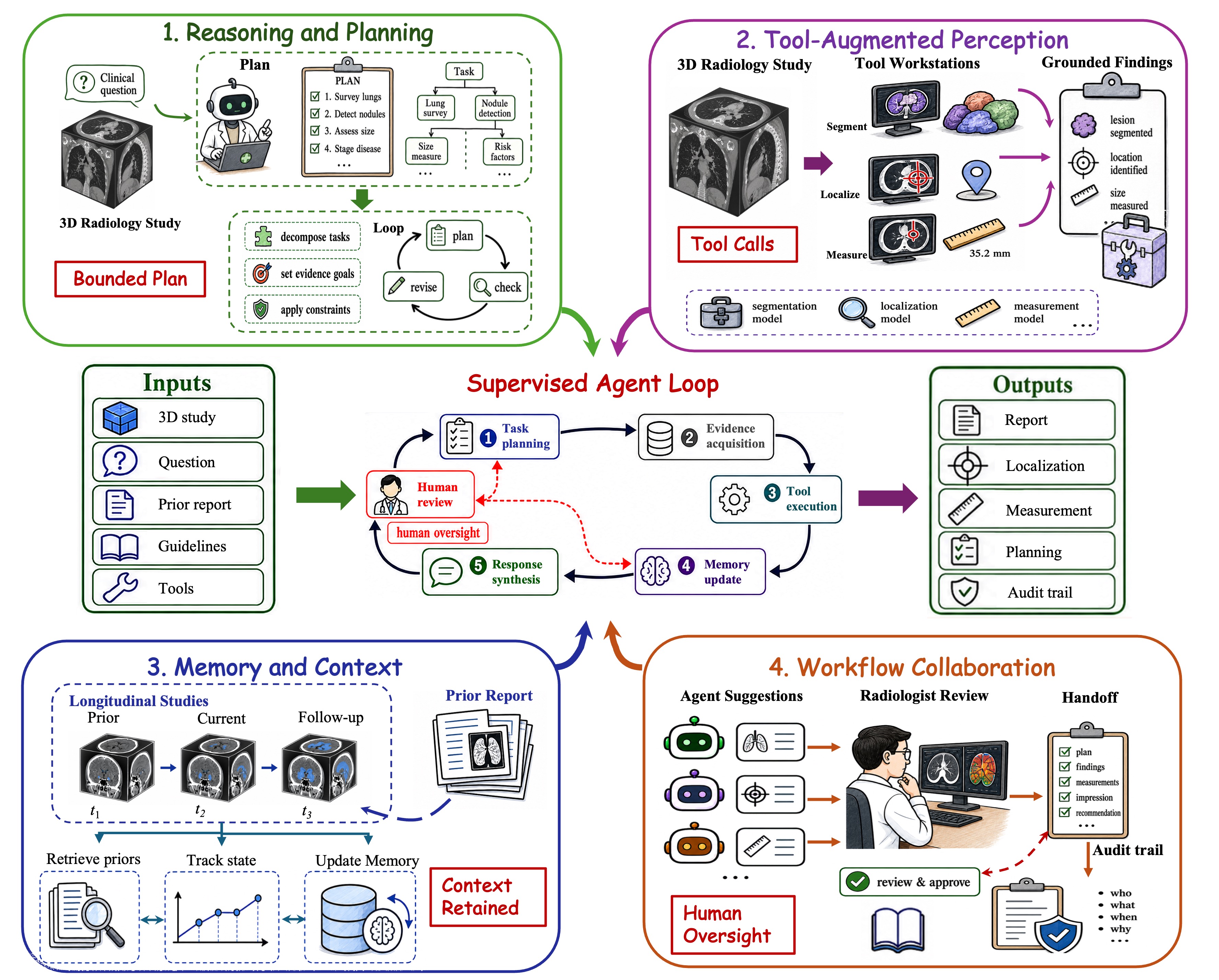}
    \caption{\textbf{Agentic workflow and modules for volumetric radiology analysis.} The four capability families correspond to the main operations required for volumetric analysis: selecting and sequencing evidence, converting clinical intent into spatial or quantitative actions, retaining context across analysis steps and examinations, and coordinating intermediate outputs within supervised workflows. Solid arrows denote workflow progression, dashed arrows denote feedback, and red tags indicate supervision or operational constraints.}
    \label{fig:agentic_workflow_modules}
\end{figure*}

We use these four capability families as a functional taxonomy and examine each through a common analytical lens: its role in volumetric radiology analysis, the design patterns used to implement it, and the scope and strength of its validation. Figure~\ref{fig:agentic_workflow_modules} illustrates how these capabilities form an iterative analysis loop, while Table~\ref{tab:agentic_systems_3d_radiology} maps representative systems to their dominant functional roles. The following subsections examine their design patterns and validation boundaries in greater detail.

\begin{table*}[!t]
\centering
\scriptsize
\setlength{\tabcolsep}{1.2pt}
\renewcommand{\arraystretch}{1.08}
\caption{Representative agentic systems for volumetric radiology analysis, organized by the primary capability emphasized by each system. MCP: Model Context Protocol}
\label{tab:agentic_systems_3d_radiology}
\begin{adjustbox}{max width=\textwidth}
\begin{tabularx}{\textwidth}{@{}
>{\raggedright\arraybackslash}p{0.24\textwidth}
>{\raggedright\arraybackslash}p{0.16\textwidth}
>{\raggedright\arraybackslash}p{0.17\textwidth}
>{\raggedright\arraybackslash}p{0.14\textwidth}
>{\raggedright\arraybackslash}p{0.12\textwidth}
>{\raggedright\arraybackslash}X
@{}}
\toprule
\textbf{Representative systems} & \textbf{Main capability} & \textbf{3D-specific design} & \textbf{Input and output} & \textbf{Evaluation focus} & \textbf{Main limitations} \\
\midrule
\multicolumn{6}{@{}l}{\textit{Reasoning and planning}} \\
CT-Agent~\cite{mao2025ct}, 3DMedAgent~\cite{wang20263dmedagent}, RadAgent~\cite{roschewitz2026radagent}, MedScribe~\cite{orlando2026medscribe}, MedOpenClaw~\cite{shen2026medopenclaw}, MedVistaGym~\cite{lu2026medvistagym}, PD-CTAgent~\cite{dong2026policy}
& Decompose volumetric analysis into intermediate reasoning, targeted inspection or evidence acquisition, tool use, and synthesis steps
& Global-to-local CT analysis, phase-aware evidence sufficiency, iterative phase escalation, full-study navigation, checklist-guided interpretation, and tool-integrated reasoning
& Full radiology studies or partial multi-phase evidence, with outputs including answers, phase requests, localized findings, reports, and reasoning or tool-use trajectories
& VQA accuracy, report quality, diagnostic and phase-escalation accuracy, reasoning trace, tool-use trajectory, expert review
& Reasoning traces and acquisition policies are mostly evaluated indirectly or retrospectively and are not yet reliable indicators of faithful clinical control \\
\midrule
\multicolumn{6}{@{}l}{\textit{Tool-augmented perception and grounded action}} \\
MedToolica~\cite{hosseini2026medtoolica}, Neuro-Radiological Agent~\cite{erdur2026agenticlargelanguagemodels}, BAAI Cardiac Agent~\cite{qu2026baai}, CTPA-Agent~\cite{zhong2025vision}, GAZE~\cite{alim2026gaze}, AgentMRI~\cite{sajua2025agentmri}, NeuroAgent~\cite{zhong2026neuroagent}, NEXUS~\cite{han2026nexus}, VoxelPrompt~\cite{hoopes2024voxelprompt}, MedSAM-Agent~\cite{liu2026medsam}, MedSegAgent~\cite{huang2026medsegagent}
& Invoke specialized perception, reconstruction, measurement, retrieval, viewer, or pipeline-orchestration tools
& Quantitative abdominal CT reasoning, 3D masks, volumetry, cardiac indices, CTPA abnormality labels, MRI correction, viewer operations, multimodal preprocessing pipelines, multi-agent workflow orchestration, and promptable segmentation
& Volume or imaging study to mask, measurement, corrected image, abnormality label, answer, or report
& Quantitative reasoning, segmentation, detection, AUROC, reconstruction quality, grounding quality, tool-call validity
& Final performance depends on tool reliability, tool selection, spatial parameterization, and failure detection \\
\midrule
\multicolumn{6}{@{}l}{\textit{Memory and dynamic context management}} \\
3DMedAgent~\cite{wang20263dmedagent}, Neuro-Oracle~\cite{aiersilan2026neuro}, Agent-MIRA~\cite{vahistha2026agent}, TheraAgent~\cite{chen2026theraagent}, BT-RADS Agent~\cite{jabal2026agentic}, LungNoduleAgent~\cite{yang2026lungnoduleagent}
& Retain intermediate observations, retrieved cases, prior studies, longitudinal lesion state, or treatment context
& Similar-case PET retrieval, pre-to-postoperative MRI trajectories, longitudinal tumor response, lesion follow-up, and shared multi-agent memory
& Current and prior imaging with reports or clinical text to diagnosis, prognosis, response category, or report
& Prognostic accuracy, response agreement, score agreement, retrieval relevance, memory ablation
& Memory quality, source attribution, temporal comparability, and lesion correspondence remain insufficiently validated \\
\midrule
\multicolumn{6}{@{}l}{\textit{Workflow interaction and multi-agent collaboration}} \\
Radiologist Copilot~\cite{yu2025radiologist}, MARCH~\cite{lin2026march}, CT-Flow~\cite{gu2026ct}, SpineAgent~\cite{xiao2026spineagent}, DosimeTron~\cite{tzanis2026dosimetron}, GPT-Plan~\cite{wang2025feasibility}, DOLA~\cite{nusrat2025autonomous}, MARTP~\cite{wang2026martp}, SAGE~\cite{nusrat2026sage}, Scan-do Attitude~\cite{kang2025scan}, PET/CT Agent~\cite{choi2026petct}, NEXUS~\cite{han2026nexus}
& Coordinate reporting, protocol management, dosimetry, treatment planning, neuroimaging research execution, or end-to-end imaging workflows
& Role-specialized agents, MCP-based orchestration, multi-sequence MRI reporting, DICOM selection, PET/CT quantification, dose optimization, composable primitives, and constraint checking
& Clinical workflow inputs to draft report, protocol action, dose estimate, treatment plan, or decision-support output
& Workflow completion, expert preference, dose or plan quality, processing time, correction burden
& Prospective validation, workload measurement, failure recovery, and human handoff are still incompletely studied \\
\bottomrule
\end{tabularx}
\end{adjustbox}
\end{table*}

\subsection{Reasoning and Planning}
\label{sec:agent_reasoning_planning}

This subsection examines reasoning and planning as the control layer for navigating distributed volumetric evidence. It considers how agents select the anatomy, slices, series, or phases to inspect; determine whether additional evidence or a tool call is required; and synthesize local observations into a study-level conclusion. The discussion compares anatomy-aware decomposition, procedure-guided interpretation, policy-constrained evidence acquisition, and planner--executor control, and evaluates how the resulting plans and tool-use trajectories are validated.

\noindent\textbf{Capability definition and radiological role.}
Reasoning and planning refer to the ability of an agent to decompose a radiological request into intermediate operations before producing an answer, report, or workflow action. In volumetric radiology, the central difficulty is that the relevant information may be distributed across slices, series, anatomical regions, measurements, and prior observations. Planning therefore acts as a control layer between volumetric information and language generation. It determines which anatomy should be inspected, which slices or views should be selected, whether a tool call is required, and how local observations should be synthesized into a study-level conclusion.

\noindent\textbf{Representative design patterns.}
Four design patterns help organize the current literature. Anatomy-aware decomposition narrows a study to clinically relevant regions or slices before synthesis; CT-Agent~\cite{mao2025ct} and 3DMedAgent~\cite{wang20263dmedagent} illustrate this route through global-to-local CT analysis, informative-slice selection, and aggregation of intermediate observations. Procedure-guided interpretation makes the reading process explicit. RadAgent~\cite{roschewitz2026radagent} uses stepwise chest CT understanding, while MedScribe~\cite{orlando2026medscribe} starts from candidate hypotheses and acquires localized observations through pathology-specific tools before report synthesis. Policy-constrained evidence acquisition treats planning as a decision about whether the current evidence is sufficient. PD-CTAgent~\cite{dong2026policy} abstracts each available CT phase into a structured evidence package and combines guideline retrieval, uncertainty signals, and rule-prioritized VLM control to request another phase only when the current state is judged insufficient. Planner-executor control appears in systems that link imaging findings to management, guideline, or interactive-viewer actions, including INFORM-CT~\cite{Tankel2026InformCT}, ACR-oriented reasoning agents~\cite{tziakouri2025reinforcement}, MedOpenClaw~\cite{shen2026medopenclaw}, and MedVistaGym~\cite{lu2026medvistagym}. These patterns differ less in their nominal reasoning label than in the granularity of the exposed plan: region selection, clinical checklist, evidence-acquisition policy, management action, or interactive tool trajectory.

\noindent\textbf{Reported results and practical constraints.}
Reported results support planning mainly as a practical control mechanism. Current systems are usually evaluated through final VQA accuracy, report quality, expert review, task completion, or logged tool trajectories. These endpoints show whether a planned workflow can improve observable performance or make the process more inspectable. They do not establish that the final statement is causally supported by the intermediate steps. A plausible trace may contain unnecessary actions, weakly grounded observations, or retrospectively convincing language. The current conclusion should therefore remain narrow: planning helps organize targeted inspection and tool use, but trace fidelity and clinical reasoning validity remain incompletely tested.

\subsection{Tool-Augmented Perception and Grounded Action}
\label{sec:agent_tool_grounding}

This subsection examines tool-augmented perception as the mechanism that converts language-mediated intent into explicit spatial or quantitative operations on a volumetric study. It distinguishes specialist perception tools, viewer or retrieval tools, and language-to-spatial-action interfaces, and then considers tool governance and failure detection.

\noindent\textbf{Capability definition and radiological role.}
Tool-augmented perception refers to the use of external modules for operations that language generation cannot perform reliably by itself. In volumetric radiology, these operations include segmentation, registration, reconstruction correction, measurement, viewer manipulation, retrieval, DICOM handling, standardized uptake value (SUV) conversion, and dose simulation. The role of tool use is to turn a textual or clinical request into explicit spatial grounding targets or quantitative outputs, such as a mask, measurement, corrected image, selected series, or dose map. A statement about a pulmonary embolus, enhancing lesion, ventricular volume, or absorbed dose is clinically interpretable only when it is tied to the correct image region, coordinate frame, series, or measurement pipeline.

\noindent\textbf{Representative design patterns.}
Tool-augmented perception supports validation when the tool output becomes part of a traceable path from image data to final claim. A first pattern is specialist tool coordination, in which an agent delegates spatial or quantitative operations to domain-specific modules. MedToolica~\cite{hosseini2026medtoolica} applies this principle to quantitative 3D abdominal CT reasoning, coordinating expert tools for organ-centric measurement, pathology assessment, spatial comparison, and clinical interpretation. PD-CTAgent~\cite{dong2026policy} uses organ-aware segmentation and disease-specific tool routing to construct evidence packages that include lesion measurements, quality-control flags, and structure-aware uncertainty before phase-sufficiency reasoning. The Neuro-Radiological Agent~\cite{erdur2026agenticlargelanguagemodels} combines skull stripping, registration, tumor segmentation, volumetry, and response assessment for brain MRI, and the BAAI Cardiac Agent~\cite{qu2026baai} integrates cardiac MRI segmentation, functional quantification, tissue characterization, diagnosis, and reporting. In nuclear medicine, DosimeTron~\cite{tzanis2026dosimetron} automates DICOM metadata extraction, PET/CT preprocessing, Monte Carlo simulation, organ segmentation, and dosimetric reporting, whereas the PET/CT Agent~\cite{choi2026petct} coordinates raw DICOM series selection, registration and resampling, SUV conversion, segmentation, maximum-intensity projection or fusion-image generation, vision-enabled understanding, and structured draft reporting. A second pattern gives the agent access to viewer, reconstruction, retrieval, or preprocessing tools. AgentMRI~\cite{sajua2025agentmri}, GAZE~\cite{alim2026gaze}, and NeuroAgent~\cite{zhong2026neuroagent} illustrate this broader information-access role. A third pattern treats language as an interface for spatial action, as in VoxelPrompt~\cite{hoopes2024voxelprompt}, MedSAM-Agent~\cite{liu2026medsam}, MedSegAgent~\cite{huang2026medsegagent}, MedSAM3~\cite{liu2025medsam3}, and IBISAgent~\cite{jiang2026ibisagent}. The key distinction among these patterns is the form of checkable output: numeric measurements and masks are more directly verifiable than viewer manipulations or retrieved contextual material.

\noindent\textbf{Reported results and practical constraints.}
Support for tool-augmented perception is most direct when the tool produces a directly measurable output. Masks, volumetry, cardiac indices, reconstruction corrections, abnormality labels, and dose estimates can be assessed with task-specific metrics. The central unresolved problem is governance of the tool chain. An agent may choose the wrong tool, pass incorrect parameters, accept a flawed segmentation, use the wrong DICOM series, or combine incompatible outputs. ToolSelect~\cite{saha2026picking} and tool-expertise-aware agents~\cite{huai2026tool} highlight this issue in broader medical-agent settings. In volumetric radiology, such errors are amplified by spatial dependence: an inaccurate registration can change longitudinal response assessment, and an incorrect mask can alter dose or volume estimates. Tool validity therefore has to be evaluated both at the module level and at the workflow level, where tool selection, parameterization, failure detection, and final claim attribution are all visible.

\subsection{Memory and Dynamic Context Management}
\label{sec:agent_memory_context}

This subsection examines memory and dynamic context management as the mechanism for integrating evidence across slices, series, analysis steps, retrieved cases, and longitudinal examinations. It distinguishes working memory, longitudinal patient memory, and retrieval-augmented memory, and then considers provenance, comparability, and lesion correspondence.

\noindent\textbf{Capability definition and radiological role.}
Memory and context management refer to the ability of an agent to retain, retrieve, and update information that is not contained in a single model input. In volumetric radiology, this capability is required because a study can include many slices and series, and a patient may also have prior scans, prior reports, treatment history, and longitudinal lesion measurements. Memory supports intra-study aggregation, prior-study comparison, similar-case retrieval, and response assessment over time. It also addresses a practical limitation of direct MLLM inference: not all visual tokens, reports, measurements, retrieved cases, and tool outputs can remain in the active context simultaneously.

\noindent\textbf{Representative design patterns.}
Current memory designs differ by the time scale and source of the stored information. Working memory operates within a single analysis episode. CT-Agent~\cite{mao2025ct} and 3DMedAgent~\cite{wang20263dmedagent} store selected regions, intermediate observations, retrieved context, or tool outputs while analyzing a study. PD-CTAgent~\cite{dong2026policy} records a compact phase-by-phase state containing the active CT phase, structured evidence package, policy decision, and matched guideline trace, enabling newly requested phases to be reprocessed rather than merely appended to the language context. Longitudinal patient memory is more clinically distinctive. Neuro-Oracle~\cite{aiersilan2026neuro} models pre-to-postoperative MRI trajectories for seizure prognosis, TheraAgent~\cite{chen2026theraagent} uses self-evolving memory for PET theranostic response prediction, BT-RADS Agent~\cite{jabal2026agentic} supports brain-tumor follow-up scoring, and LungNoduleAgent~\cite{yang2026lungnoduleagent} uses shared memory and multi-agent discussion for lung nodule diagnosis. Retrieval-augmented memory expands the context beyond the current case, as in Agent-MIRA~\cite{vahistha2026agent}, GAZE~\cite{alim2026gaze}, and ACR guideline agents~\cite{pambudi2025bridging}. These patterns are not interchangeable: working memory supports local aggregation, longitudinal memory supports change assessment, and retrieval-augmented memory supports comparison with external cases or guidance.

\noindent\textbf{Reported results and practical constraints.}
Memory-related claims are usually supported by task-level outcomes such as prognostic accuracy, response agreement, score agreement, retrieval relevance, ablation of memory components, or longitudinal consistency. These signals indicate whether stored context helps the target task, but they rarely isolate the memory module from the perception model, retrieval system, or decision layer. Stored or retrieved information may also be inaccurate, outdated, or non-comparable. A prior report may contain an error, a retrieved PET case may differ in disease stage, and lesion correspondence may fail after surgery or treatment. For retrieval systems, nearest-neighbor similarity in feature space does not necessarily imply clinical comparability. For longitudinal systems, valid temporal reasoning depends on registration quality, lesion matching, and treatment timing. Memory therefore strengthens agentic radiology only when source attribution, update rules, and comparability are visible to the reviewer.

\subsection{Workflow Interaction and Multi-Agent Collaboration}
\label{sec:agent_workflow_interaction}

This subsection examines workflow interaction and multi-agent collaboration as the coordination layer linking models, tools, intermediate artifacts, and human review across volumetric radiology procedures. It covers reporting, planning and operational workflows, reproducible multi-agent execution, supervised assistance, handoff, and workload validation.

\noindent\textbf{Capability definition and radiological role.}
Workflow interaction and multi-agent collaboration refer to systems that place LLMs, MLLMs, or VLMs within clinical, technical, or research procedures. Instead of producing a single answer, the system coordinates roles, tools, intermediate artifacts, and human review. This capability is relevant to volumetric radiology because reporting, dosimetry, treatment planning, protocol management, spine MRI understanding, and neuroimaging analysis all depend on multi-step procedures with handoffs between people, software, and quantitative outputs. In this setting, the agent functions as an organizer of the workflow rather than only an image interpreter.

\noindent\textbf{Representative design patterns.}
Workflow-oriented agents differ from task agents because the unit of operation is a clinical or technical process rather than a single prediction. Reporting workflow assistance is the first pattern. Radiologist Copilot~\cite{yu2025radiologist} coordinates reporting tools with quality control, MARCH~\cite{lin2026march} models a resident-fellow-attending hierarchy for CT report generation, and CT-Flow~\cite{gu2026ct} uses model context protocol servers to make interpretation state and tool invocation explicit. SpineAgent~\cite{xiao2026spineagent} extends this reporting logic to multi-sequence spine MRI by combining a multi-sequence foundation model with specialized agents for diagnosis, pathological-region localization, similar-case retrieval, and a medical report agent. A second pattern is planning and operational workflow assistance. PD-CTAgent~\cite{dong2026policy} conditions evidence-sufficiency decisions and report emphasis on an active institutional rule set; its scope is to decide whether additional phase evidence is needed from partially observed studies, rather than to prescribe acquisition autonomously. DosimeTron~\cite{tzanis2026dosimetron}, GPT-Plan~\cite{wang2025feasibility}, DOLA~\cite{nusrat2025autonomous}, MARTP~\cite{wang2026martp}, and SAGE~\cite{nusrat2026sage} show how agentic control can assist PET/CT dosimetry or radiotherapy planning, where dose estimates, constraints, and role-specific checks matter more than open-ended dialogue. A third pattern focuses on reproducible execution, including NeuroClaw~\cite{shen2026neuroclaw}, artifact-based medical image processing agents~\cite{zuo2026artifact}, and NEXUS~\cite{han2026nexus}. These examples show that agentic workflows can support upstream acquisition or reconstruction tasks, downstream interpretation and reporting, and research-oriented processing pipelines.

\noindent\textbf{Reported results and practical constraints.}
Workflow agents are supported by outcomes matched to their intended use: report quality, expert preference, RADPEER-style review, dose agreement, processing time, plan quality, constraint satisfaction, or reproducible execution logs. SpineAgent illustrates this standard by combining automated report-generation metrics with review by five radiologists, but even this framing remains explicitly assistive and in-scope. Across the broader literature, relatively few studies measure whether the agent selected the correct tool, used correct parameters, detected failures, preserved intermediate observations, or reduced clinician workload under realistic constraints. The relevant validation standard also changes by workflow: reporting assistants require claim accuracy and review efficiency, dosimetry agents require quantitative dose agreement and segmentation reliability, planning agents require constraint satisfaction and deliverability, and research agents require reproducibility. Current systems should therefore be interpreted as bounded and supervised assistance systems. Their near-term value is to organize complex volumetric radiology workflows and make intermediate steps more traceable, while clinical authority remains with human experts.


Together, these four capability families form an iterative volumetric-analysis loop: reasoning and planning select and sequence the evidence to inspect; tool-augmented perception produces grounded spatial or quantitative outputs; memory integrates evidence across analysis steps and examinations; and workflow interaction coordinates these outputs with clinical procedures and human review. Agentic volumetric radiology systems thereby extend or complement volumetric radiology MLLMs at the workflow level. 


\section{Clinical Applications and Evaluation in Volumetric Radiology}
\label{sec:evaluation}

Clinical evaluation begins with the intended claim. The capabilities reviewed in Sections~\ref{sec:foundation} and~\ref{sec:agent} do not by themselves establish clinical validity: the appropriate evaluation depends on what a system is intended to predict, communicate, or support in practice. Accordingly, validation should be matched to both the type of clinical output and the strength of the claim being made.

We therefore organize clinical applications and evaluation according to four claim-driven families: diagnostic interpretation and reporting, prognosis and treatment response, image-derived planning and clinical decision support, and cross-task evaluation and validation. These families differ in the clinical question being addressed and, consequently, in the validation required to support those claims. Within each family, we examine the intended clinical role, the corresponding evaluation strategy, and the principal limitations of current validation practice.

\begin{figure*}[t]
    \centering
    \includegraphics[width=0.95\textwidth]{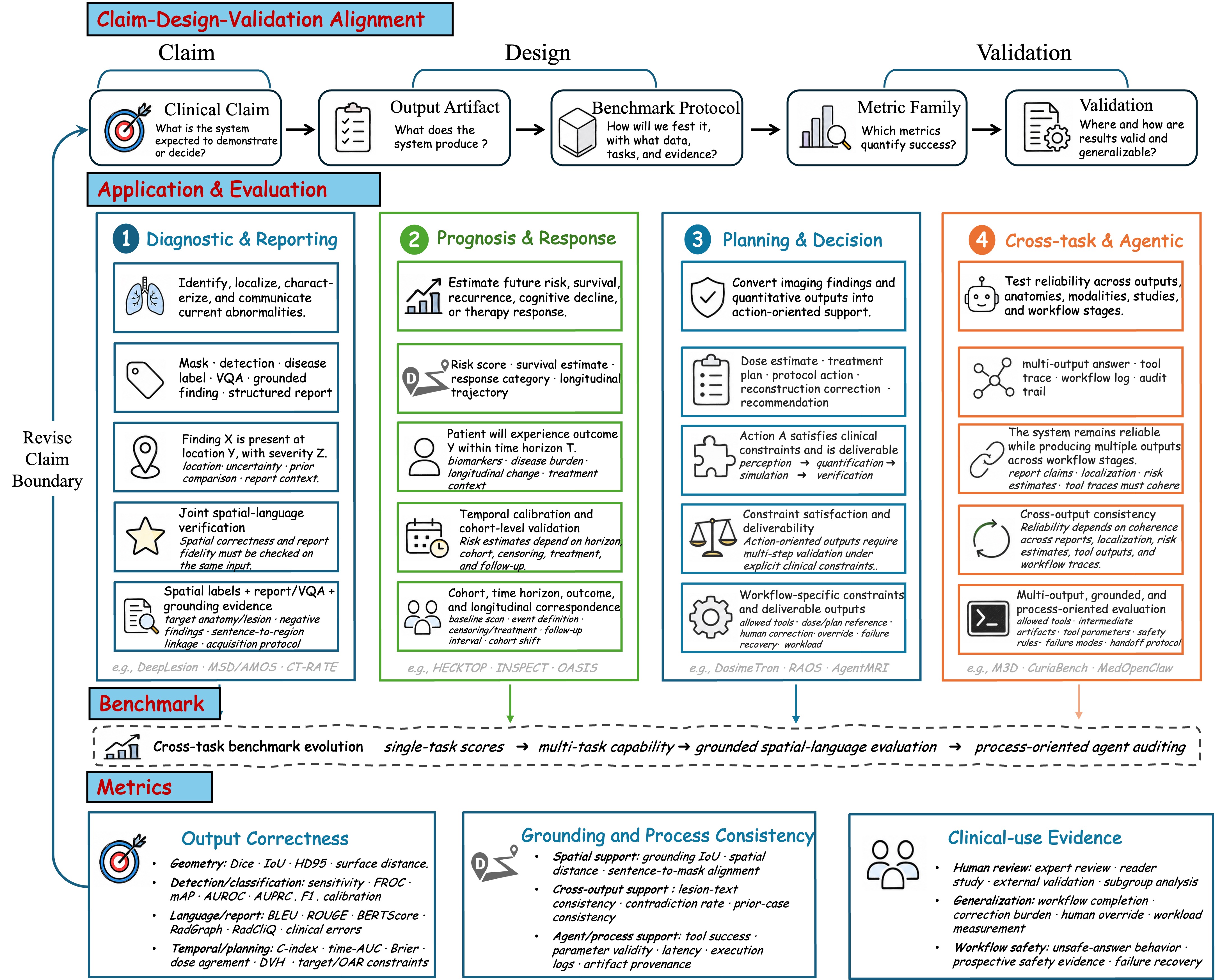}
    \caption{\textbf{Clinical applications and evaluation for volumetric radiology.} The figure maps four application families to their claim types, validation demands, benchmark protocols, and metric evidence, emphasizing that evaluation standards should follow the clinical claim and intended output.}
    \label{fig:evaluation_benchmark_framework}
\end{figure*}

The strength of a clinical claim should also be commensurate with the level of validation. Internal benchmarks primarily establish technical feasibility, external or multi-institutional studies assess robustness, reader studies and workflow simulations examine clinical usability, and prospective evaluation is required to support deployment-oriented claims. This progression operationalizes the validation component of the Claim–Design–Validation framework introduced in Section~\ref{sec:intro}. Figure~\ref{fig:evaluation_benchmark_framework} summarizes the relationship among clinical claims, evaluation strategies, and validation levels, while Tables~\ref{tab:clinical_applications_evaluation_3d_radiology} and~\ref{tab:benchmark_resources_3d_radiology} summarize representative systems and benchmark resources.

\begin{table*}[!t]
\centering
\footnotesize
\setlength{\tabcolsep}{3pt}
\renewcommand{\arraystretch}{1.15}

\caption{Clinical application families and validation requirements in volumetric radiology AI. Representative systems and evaluation resources are summarized by clinical claim, task scope, commonly reported evidence, and
principal validation gap. Dataset-level details are provided separately in Table~\ref{tab:benchmark_resources_3d_radiology}. AUC, area under the curve; AUPRC, area under the precision--recall curve; AUROC, area under the receiver operating characteristic curve; C-index, concordance index; DVH, dose--volume histogram; FROC, free-response receiver operating characteristic; HD95, 95th percentile Hausdorff distance; IoU, intersection over union; PE, pulmonary embolism.}
\label{tab:clinical_applications_evaluation_3d_radiology}

\begin{tabularx}{\textwidth}{@{}
>{\raggedright\arraybackslash}p{0.25\textwidth}
>{\raggedright\arraybackslash}p{0.16\textwidth}
>{\raggedright\arraybackslash}p{0.17\textwidth}
>{\raggedright\arraybackslash}p{0.20\textwidth}
>{\raggedright\arraybackslash}X
@{}}
\toprule
\textbf{Systems/resources}
&
\textbf{Clinical claim}
&
\textbf{Representative tasks/outputs}
&
\textbf{Common evaluation evidence}
&
\textbf{Principal validation gap}
\\

\midrule
\multicolumn{5}{@{}l}{\textit{Diagnostic interpretation and reporting}} \\

CT-RATE~\cite{hamamci2026generalist},
CT2Rep~\cite{hamamci2024ct2rep},
M$^3$D~\cite{bai2024m3d},
Merlin~\cite{blankemeier2026merlin},
MedVista3D~\cite{li2025medvista3d},
CT-Agent~\cite{mao2025ct}, and
MedScribe~\cite{orlando2026medscribe}
&
Identify, localize, characterize, and communicate clinically relevant findings
&
Segmentation, detection, classification, VQA, grounded findings, and report
generation
&
Dice, IoU, HD95, and FROC; AUROC and calibration; BLEU/ROUGE, RadGraph,
RadCliQ, and grounding IoU
&
Clinical correctness and traceability to volumetric evidence remain
incompletely assessed
\\

\midrule
\multicolumn{5}{@{}l}{\textit{Prognosis and treatment response}} \\

CLIP-Lung~\cite{lei2023clip},
LoV3D~\cite{Jiang2026LoV3D},
CTPA-Agent~\cite{zhong2025vision},
Neuro-Oracle~\cite{aiersilan2026neuro},
TheraAgent~\cite{chen2026theraagent}, and
BT-RADS Agent~\cite{jabal2026agentic}
&
Predict future risk or clinically meaningful change over time
&
Malignancy and survival prediction, longitudinal response assessment, and
cognitive outcomes
&
AUROC, AUPRC, C-index, time-dependent AUC, Brier score, calibration, and
response agreement
&
Temporal validity, including calibration and lesion correspondence, remains
insufficiently established
\\

\midrule
\multicolumn{5}{@{}l}{\textit{Image-derived planning and clinical decision support}} \\

DosimeTron~\cite{tzanis2026dosimetron},
GPT-Plan~\cite{wang2025feasibility},
SAGE~\cite{nusrat2026sage},
AgentMRI~\cite{sajua2025agentmri},
Scan-do Attitude~\cite{kang2025scan}, and
the PET/CT Agent~\cite{choi2026petct}
&
Translate imaging evidence into technically valid planning or management
outputs
&
Dosimetry, radiotherapy planning, reconstruction and protocol control, and
incidental finding management
&
Dose agreement, dose--volume constraints, reconstruction quality, workflow
completion, and expert correction
&
End-to-end error propagation and recovery remain poorly characterized
\\

\midrule
\multicolumn{5}{@{}l}{\textit{Cross-task evaluation and validation}} \\

M$^3$D~\cite{bai2024m3d},
CuriaBench~\cite{saporta2026curia},
CT-SpatialVQA~\cite{monon2026ctspatialvqa},
Med-StepBench~\cite{nguyen2026medstepbench},
CORTEX~\cite{malik2026cortex},
ABRA~\cite{maksudov2026abra},
MedCTA~\cite{ashraf2026medcta}, and
RadSaFE-200~\cite{wind2026safety}
&
Assess coherence across outputs and workflow stages
&
Multi-task VQA and reporting, spatial reasoning, grounding, tool-use
evaluation, and safety assessment
&
Task-specific scores, grounding IoU, spatial error, QAScore, trace fidelity,
workflow completion, and unsafe-answer rate
&
Integrated case-level validation and detection of propagated errors remain
limited
\\

\bottomrule
\end{tabularx}

\vspace{3pt}
\begin{minipage}{\textwidth}
\scriptsize
\textit{Abbreviations:}
AUC, area under the curve;
AUPRC, area under the precision--recall curve;
AUROC, area under the receiver operating characteristic curve;
BLEU, Bilingual Evaluation Understudy;
C-index, concordance index;
FROC, free-response receiver operating characteristic;
HD95, 95th percentile Hausdorff distance;
IoU, intersection over union;
ROUGE, Recall-Oriented Understudy for Gisting Evaluation;
VQA, visual question answering.
\end{minipage}

\end{table*}

\begin{table*}[!t]
\centering
\scriptsize
\setlength{\tabcolsep}{1.6pt}
\renewcommand{\arraystretch}{1.06}
\caption{Representative benchmark datasets and evaluation resources for volumetric radiology models. Dataset names follow the corresponding papers or project names, and resources are grouped by the application and evaluation tasks emphasized in this section. CTPA, computed tomography pulmonary angiography; CXR, chest radiography; OAR, organ at risk; PACS, picture archiving and communication system; PSMA, prostate-specific membrane antigen; RT, radiotherapy}
\label{tab:benchmark_resources_3d_radiology}
\begin{adjustbox}{max width=\textwidth}
\begin{tabularx}{\textwidth}{@{}
>{\raggedright\arraybackslash}p{0.20\textwidth}
>{\raggedright\arraybackslash}p{0.28\textwidth}
>{\raggedright\arraybackslash}p{0.18\textwidth}
>{\raggedright\arraybackslash}p{0.14\textwidth}
>{\raggedright\arraybackslash}X
@{}}
\toprule
\textbf{Dataset Name} & \textbf{Task} & \textbf{Modality} & \textbf{Sample Size} & \textbf{Highlight} \\
\midrule
\multicolumn{5}{@{}l}{\textit{Diagnostic interpretation and reporting}} \\
MedMNIST v2~\cite{Yang2023MedMNIST}
& 2D and 3D biomedical image classification
& Multiple 2D and 3D modalities
& 708K 2D images, 10K 3D images
& Lightweight, standardized, diverse labels \\
DeepLesion~\cite{Yan2018DeepLesion}
& Multi-class lesion detection and categorization
& CT
& 33,688 annotated lesion images
& PACS-mined, heterogeneous lesions \\
AMOS~\cite{Ji2022Amos}
& Multi-organ abdominal segmentation
& Abdominal CT and MRI
& 500 CT, 100 MRI scans
& Multi-center, multi-modality, 15 organs \\
MSD~\cite{Antonelli2022MSD}
& Segmentation across multiple anatomical tasks
& 3D and 4D CT and MRI
& 2,633 3D volumes
& 10 tasks, generalizability benchmark \\
AbdomenAtlas~\cite{Li2025AbdomenAtlas}
& Abdominal multi-organ segmentation and transfer learning
& Contrast-enhanced CT
& 20,460 CT volumes
& 112 hospitals, large-scale annotations \\
TotalSegmentator MRI~\cite{Akinci2025TotalSegmentatorMRI}
& Whole-body multi-structure segmentation
& Whole-body MRI
& 616 MRI examinations
& Sequence-independent, 50 structures \\
AMOS-MM~\cite{Ji2022Amos}
& Abdominal report generation and VQA
& Multi-phase abdominal CT
& 2,300 CT scans
& Image-text pairs, abdominal multimodal tasks \\
CT2Rep~\cite{hamamci2024ct2rep}
& 3D CT report generation
& Chest CT and reports
& 25,701 CT volumes
& 3D report generation, automated reporting \\
CT-RATE~\cite{hamamci2026generalist}
& CT classification and report-grounded modeling
& CT and text
& 25,692 scans
& Public paired 3D images and reports \\
CT-3DRRG~\cite{Liu2025Argus}
& 3D CT report generation, retrieval, and grounding
& Multimodal 3D CT
& --
& 3DRRG benchmark, visual token compression \\
RadGenome-Chest CT~\cite{Zhang2025RadGenomeChestCT}
& Grounded report generation, VQA, organ segmentation
& Chest CT and aligned text
& 25,692 CTs, 1.3M VQA pairs
& Region-guided, 197 organ masks \\
AutoRG-Brain~\cite{Lei2024AutoRGBrain}
& Brain MRI report generation and region grounding
& Brain MRI and reports
& 3,408 imaging-report pairs
& Grounded brain MRI reporting \\
3D-BrainCT~\cite{li2024towards}
& Brain CT report generation and MLLM evaluation
& Brain CT and text
& 18,885 text-scan pairs
& FORTE metric, brain CT reporting \\
DeepTumorVQA~\cite{Chen2025DeepTumorVQA}
& Tumor-centric VQA and clinical reasoning
& Abdominal CT
& 9,262 CT volumes
& Small tumor detection, 4 QA categories \\
NOVA~\cite{Bercea2025Nova}
& Anomaly localization and step-wise reasoning
& Brain MRI and text
& 900 scans
& Rare pathologies, stress-test setting \\
ReXGroundingCT~\cite{Baharoon2025ReXGroundingCT}
& Free-text finding grounding and lesion localization
& Chest CT
& 3,142 CT scans
& Sentence-level grounding, 3D masks \\
ViPET-ReportGen~\cite{Nguyen2025ViPETReportGen}
& PET/CT report generation and auxiliary classification
& PET/CT and text
& 2,757 PET/CT volumes
& Whole-body functional imaging, Vietnamese reports \\
SGMRI-VQA~\cite{moukheiber2026sgmri}
& Detection, localization, classification, counting, diagnosis
& Volumetric MRI
& 41,307 QA pairs
& Multi-frame reasoning, box supervision \\
SpatialMed~\cite{Trinh2026SpatialMed}
& 3D spatial reasoning and volume estimation
& CT and text
& 10K QA pairs
& Spatial reasoning, numerical estimation \\
CT-SpatialVQA~\cite{monon2026ctspatialvqa}
& Semantic-spatial VQA
& CT and text
& 9,077 QA pairs from 1,601 CT volumes and reports
& Anatomical localization, laterality, structural comparison, 3D relational reasoning \\
CORTEX~\cite{malik2026cortex}
& Structured reasoning, VQA, and report generation
& Chest CT and text
& 76,177 validated reasoning traces
& Four-stage diagnostic traces, stage-wise rubric scoring, expert review \\
PET-CLIP Captioner~\cite{yu2025location}
& Whole-body PET/CT lesion captioning
& Whole-body PET/CT
& 1,867 subjects
& Private cohort, location-guided captioning \\
Med-StepBench~\cite{nguyen2026medstepbench}
& Step-wise hallucination detection
& Oncological PET/CT
& $>$12,000 images and $>$1,000,000 image-statement pairs
& Four-stage diagnostic reasoning, clinician-verified hallucination evaluation \\
Oncology VQA~\cite{liu2026oncologyvqa}
& Report-derived oncology VQA
& 3D oncology CT and MRI
& 2,511 cases and 42,997 questions
& Private-cohort benchmark, RADS-style and report-derived questions, blind ablation \\
ReportQA~\cite{shi2026reportqa}
& QA-based report evaluation
& CXR, brain CT, chest CT, abdominal CT
& 6,857 reports and approximately 660,000 QA pairs
& QAScore, knowledge-tree report evaluation, multi-modality report utility \\
\midrule
\multicolumn{5}{@{}l}{\textit{Prognosis and treatment response}} \\
INSPECT~\cite{Huang2023Inspect}
& Pulmonary embolism diagnosis and prognosis
& CTPA, CXR, clinical data
& 19,402 patients
& Longitudinal records, multiple outcomes \\
HECKTOR~\cite{Andrearczyk2023Hecktor}
& Tumor segmentation and outcome prediction
& Head-and-neck PET/CT
& 224 cases
& Survival prediction, PET/CT tumors \\
OASIS-3~\cite{LaMontagne2019Oasis3}
& Longitudinal neuroimaging and cognitive outcomes
& Brain MRI, PET, clinical data
& 1,098 participants
& 15-year tracking, Alzheimer's cohort \\
BraTS series~\cite{Menze2015Brats}
& Brain tumor segmentation and outcome-oriented tracks
& Multi-parametric brain MRI
& Challenge series
& Expert annotations, multi-track evaluation \\
3D-RAD~\cite{Gai2025ThreeDRad}
& Multi-temporal VQA and longitudinal diagnosis
& 3D CT
& 136,195 expert-aligned samples
& Static and temporal diagnosis tasks \\
Gastric-X~\cite{Lu2026GastricX}
& Gastric cancer detection, staging, response assessment
& Multi-phase CT, endoscopy
& 1.7K cases
& Multi-phase CT, multimodal cancer benchmark \\
\bottomrule
\end{tabularx}
\end{adjustbox}
\end{table*}

\newcommand{\benchmarkresourcescontinuation}{%
\begin{table*}[!t]\ContinuedFloat
\centering
\scriptsize
\setlength{\tabcolsep}{1.6pt}
\renewcommand{\arraystretch}{1.06}
\caption{Representative benchmark datasets and evaluation resources for volumetric radiology models. Continued.}
\begin{adjustbox}{max width=\textwidth}
\begin{tabularx}{\textwidth}{@{}
>{\raggedright\arraybackslash}p{0.20\textwidth}
>{\raggedright\arraybackslash}p{0.28\textwidth}
>{\raggedright\arraybackslash}p{0.18\textwidth}
>{\raggedright\arraybackslash}p{0.14\textwidth}
>{\raggedright\arraybackslash}X
@{}}
\toprule
\textbf{Dataset Name} & \textbf{Task} & \textbf{Modality} & \textbf{Sample Size} & \textbf{Highlight} \\
\midrule
\multicolumn{5}{@{}l}{\textit{Image-derived planning and clinical decision support}} \\
CT-FlowBench~\cite{gu2026ct}
& Quantitative, spatial, and diagnostic tool-use reasoning
& CT
& --
& Agentic workflow, tool invocation \\
RAOS~\cite{Luo2024RAOS}
& Robust abdominal organ segmentation for radiotherapy scenarios
& Abdominal CT
& 413 CT scans
& Challenging cases, organ hallucination \\
HaN-Seg~\cite{Podobnik2024HaNSeg}
& Head-and-neck organ-at-risk segmentation for RT planning
& CT and T1-weighted MRI
& 42 training, 14 test cases
& MultimodalOAR segmentation \\
AAPM-RT-MAC~\cite{Cardenas2020AAPMRTMAC}
& Normal tissue auto-segmentation for MR-guided RT planning
& T2-weighted head-and-neck MRI
& 55 patients
& Expert contours, RT grand challenge \\
DosimeTron cohort~\cite{tzanis2026dosimetron}
& Patient-specific Monte Carlo internal dosimetry
& PSMA PET/CT
& 597 studies, 378 patients
& System-level cohort, dosimetry validation \\
AgentMRI evaluation set~\cite{sajua2025agentmri}
& MRI artifact detection and reconstruction model selection
& Brain MRI
& 150 subjects, 1,978 slices
& Reconstruction control, degradation-specific tasks \\
Scan-do Attitude protocol set~\cite{kang2025scan}
& CT protocol configuration and protocol-file generation
& CT protocol metadata
& --
& System-level set, protocol management \\
\midrule
\multicolumn{5}{@{}l}{\textit{Cross-task evaluation and validation}} \\
M3D~\cite{bai2024m3d}
& Retrieval, report generation, VQA, positioning, segmentation
& CT, MRI, multimodaldata
& 120K image-text pairs, 662K instructions
& Generalist volumetric MLLM benchmark \\
MedVL-CT69K~\cite{shui2025large}
& Zero-shot abnormality detection and report generation
& CT
& 272,124 CT scans
& Fine-grained CT VLP, anatomy-level labels \\
Triad~\cite{Wang2025Triad}
& Segmentation, classification, registration
& 3D MRI
& 131,170 MRI volumes
& Large-scale MRI pretraining, 25 downstream datasets \\
Merlin-Abdominal-CT~\cite{blankemeier2026merlin}
& Segmentation, lesion captioning, disease prediction
& CT and text
& 25,494 CT-report pairs
& Pan-anatomy CT, long-context reasoning \\
CuriaBench~\cite{saporta2026curia}
& Segmentation, regression, survival prediction, lesion detection
& MultimodalCT and MRI
& -
& Foundation-model transfer, volumetric evaluation \\
ABRA~\cite{maksudov2026abra}
& Radiology-agent workstation interaction
& Chest CT and breast MRI
& 655 tasks
& OHIF/Orthanc environment, DICOM navigation, planning-execution-outcome scoring \\
MedCTA~\cite{ashraf2026medcta}
& Clinical tool-agent evaluation
& Radiology images, pathology slides, reports
& 107 clinical tasks
& Clinician-verified tool trajectories, tool selection, argument validity, rollout reliability \\
Perception-Bench~\cite{liu2026radsight}
& Attribute judgment, spatial grounding and understanding, diagnosis, anomaly detection, and report generation
& 2D radiology images and 3D CT
& 1.13M samples
& Six-level evidence hierarchy linking low-level perception to clinical interpretation \\
MedReCo-DB~\cite{zhang2026vision}
& Entity-conditioned retrieval and comparative VQA across reference cases and longitudinal studies
& CXR, CT, MRI, and ultrasound
& $>$690K images, $>$160K patients
& Eight institutions, seven modalities, entity-aware cross-image comparison \\
MedIF-Bench~\cite{yuan2026clinfusion}
& Medical instruction-following and structured-output compliance
& Mixed text and 2D/3D medical imaging tasks
& 900 samples
& Clinically relevant formatting constraints with rule-based compliance scoring \\
\bottomrule
\end{tabularx}
\end{adjustbox}
\end{table*}
}

\subsection{Diagnostic Interpretation and Reporting}
\label{sec:eval_diagnostic_reporting}
\label{sec:eval_diagnostic}

\noindent\textbf{Task definition and clinical role.}
Diagnostic interpretation and reporting assess whether a system can identify, localize, characterize, and communicate clinically relevant findings from a volumetric radiology study. Outputs may include spatial predictions (e.g., masks and detections), diagnostic predictions (e.g., labels or scores), and language-mediated interpretations (e.g., VQA responses, grounded findings, and structured reports). This application family supports routine interpretation, triage, lesion measurement, follow-up documentation, and communication with downstream care teams. Clinical usefulness therefore requires both evidence preservation and appropriate communication: the system must retain the volumetric information supporting each claim, and the output must convey relevant attributes such as anatomical location, laterality, extent or severity, uncertainty, negative findings, and comparison with prior studies when applicable.

\noindent\textbf{Representative methods and validation demands.}
Early evaluation in this family focused primarily on segmentation and detection, for which reference masks, bounding boxes, and lesion annotations made spatial correctness directly measurable. Representative datasets include LIDC-IDRI~\cite{Armato2011LidcIdri}, LUNA16~\cite{Setio2017Luna16}, BraTS~\cite{Menze2015Brats},
MSD~\cite{Antonelli2022MSD}, AMOS~\cite{Ji2022Amos}, AbdomenAtlas~\cite{Li2025AbdomenAtlas}, and TotalSegmentator~\cite{Wasserthal2023TotalSegmentator}. These tasks support geometric and lesion-level evaluation using metrics such as Dice, intersection over union (IoU), the 95th percentile Hausdorff distance (HD95), free-response receiver operating characteristic (FROC) analysis, and lesion-level recall.

As 3D foundation models and MLLMs extended the output space toward diagnostic prediction, VQA, and report generation, evaluation also had to address semantic correctness, report completeness, and correspondence between language and image. Representative resources and systems include CT-RATE~\cite{hamamci2026generalist}, CT2Rep~\cite{hamamci2024ct2rep}, M$^3$D~\cite{bai2024m3d}, Merlin~\cite{blankemeier2026merlin}, MedVista3D~\cite{li2025medvista3d}, 3D-BrainCT~\cite{li2024towards}, and Brain3D~\cite{barone2026brain3d}. Agentic radiology systems, including CT-Agent~\cite{mao2025ct}, PD-CTAgent~\cite{dong2026policy}, RadAgent~\cite{roschewitz2026radagent}, MedScribe~\cite{orlando2026medscribe}, Radiologist Copilot~\cite{yu2025radiologist}, MARCH~\cite{lin2026march}, and the PET/CT Agent~\cite{choi2026petct}, further broaden the evaluated process to include full-study navigation, targeted inspection, phase-sufficiency decisions, tool-mediated evidence acquisition, and report verification.

The required validation evidence depends on the output type. Spatial outputs require geometric and lesion-level assessment, whereas diagnostic labels require discrimination, calibration, and error analysis. Language outputs are commonly evaluated with BLEU, ROUGE, METEOR, CIDEr, BERTScore, RadGraph, RadCliQ, clinical error measures, and question-answering metrics. These metrics capture complementary dimensions of report quality, but they do not by themselves establish whether generated statements are supported by the underlying volume. High textual similarity can coexist with clinically incorrect or ungrounded claims, while spatially accurate predictions can still be communicated inaccurately. Diagnostic evaluation should therefore determine whether the resulting interpretation is clinically correct and traceable to the relevant volumetric evidence. Expert review remains necessary when these properties cannot be reliably established by automated metrics.

\noindent\textbf{Current resources and remaining gaps.}
Current resources increasingly evaluate whether language outputs remain grounded in volumetric evidence. RadGenome-Chest CT~\cite{Zhang2025RadGenomeChestCT}, AutoRG-Brain~\cite{Lei2024AutoRGBrain}, and ReXGroundingCT~\cite{Baharoon2025ReXGroundingCT} link free-text findings to anatomical regions or lesion masks. CT-SpatialVQA ~\cite{monon2026ctspatialvqa} evaluates semantic and spatial reasoning over 3D anatomy, whereas ReportQA~\cite{shi2026reportqa} assesses report utility using knowledge-tree-derived question and answer pairs and QAScore. Perception-Bench~\cite{liu2026radsight} traces errors from lesion attributes and 2D/3D spatial grounding through diagnosis and report generation. MedReCo-DB~\cite{zhang2026vision} adds entity-conditioned reference retrieval and prior--current comparative reasoning across seven modalities, and ClinFusion~\cite{yuan2026clinfusion} complements these capability tests with MedIF-Bench for structured instruction following and an RoI-grounded factuality protocol for report evaluation. Oncology VQA~\cite{liu2026oncologyvqa} extends this evaluation to private 3D oncology cohorts with contamination-aware testing.

Collectively, these resources broaden evaluation beyond text similarity, but each addresses only part of the evidence required for a diagnostic claim. Calibration of probabilistic predictions, reader or workflow studies, and external validation across institutions and acquisition settings remain limited.

\subsection{Prognosis and Treatment Response}
\label{sec:eval_prognosis_response}
\label{sec:eval_prognosis}

\noindent\textbf{Task definition and clinical role.} Prognosis and treatment response assess whether imaging can support the prediction of future outcomes or clinically meaningful change over time. Targets include risk and time-to-event outcomes, such as malignancy, recurrence, and survival, as well as longitudinal outcomes, such as treatment response, cognitive decline, and structured follow-up scores. These tasks support risk stratification, follow-up planning, response assessment, clinical-trial enrichment, and identification of patients who require closer longitudinal review. Unlike diagnostic interpretation, the target is often not directly observable in a single study. Reliable inference must therefore relate imaging biomarkers and disease burden to temporal change, treatment exposure, and relevant clinical context.

\noindent\textbf{Representative methods and validation demands.} Radiomics first linked handcrafted imaging features to clinical outcomes through statistical models~\cite{lambin2012radiomics,gillies2016radiomics}. Supervised 3D deep learning later learned prognostic features directly from imaging. Representative cohorts and benchmarks include HECKTOR~\cite{Andrearczyk2023Hecktor}, INSPECT~\cite{Huang2023Inspect}, OASIS-3~\cite{LaMontagne2019Oasis3}, BraTS~\cite{Menze2015Brats}, and Gastric-X~\cite{Lu2026GastricX}, covering survival, progression, cognitive outcomes, and treatment response.

Recent approaches broaden the evidence available to prediction. CLIP-Lung~\cite{lei2023clip} and AutoRad-Lung~\cite{Khademi2025AutoRadLung} apply vision--language modeling to nodule malignancy assessment, whereas LoV3D~\cite{Jiang2026LoV3D} models change across longitudinal brain MRI. Systems such as CTPA-Agent~\cite{zhong2025vision}, Neuro-Oracle~\cite{aiersilan2026neuro}, Agent-MIRA~\cite{vahistha2026agent},
TheraAgent~\cite{chen2026theraagent}, and BT-RADS Agent~\cite{jabal2026agentic} further incorporate longitudinal or retrieved context into outcome prediction and structured response assessment.

Validation in this family must establish more than discrimination. AUROC and the C-index indicate whether a model ranks patients by risk, but they do not establish accurate absolute risk or reliable change assessment at a clinically relevant time point. The endpoint, prediction horizon, cohort construction, and treatment exposure must therefore be specified, with censoring and competing events handled appropriately. Without these elements, apparent performance may reflect endpoint ambiguity or cohort composition rather than clinically actionable prognostic information.

\noindent\textbf{Current resources and remaining gaps.} Current resources span binary-risk prediction, time-to-event modeling, and longitudinal response assessment. Binary outcomes are commonly evaluated with AUROC, AUPRC, sensitivity, specificity, and F1 score, whereas time-to-event models use the C-index, time-dependent AUC, and Brier score. Treatment-response studies additionally require agreement with task-appropriate clinical criteria, such as the RECIST~\cite{eisenhauer2009new}, PERCIST~\cite{wahl2009recist}, and RANO~\cite{wen2010updated}. Interpretation of these measures depends on clearly defined endpoints, cohorts, and prediction horizons.

The principal gap is incomplete temporal validation. Current studies often report discrimination or response agreement without establishing calibration at the intended horizon or reliable lesion correspondence across serial examinations. External validation also remains limited, leaving transportability beyond the development cohort uncertain. For retrieval- or memory-augmented systems, historical or external evidence must additionally
be temporally aligned and traceable to its source.

\benchmarkresourcescontinuation
\FloatBarrier

\subsection{Image-Derived Planning and Clinical Decision Support}
\label{sec:eval_planning_support}
\label{sec:eval_workflow}

\noindent\textbf{Task definition and clinical role.} Image-derived planning and clinical decision support assess whether imaging evidence can be translated into quantitative or operational outputs that inform subsequent clinical or technical decisions. These outputs include dose estimates, candidate treatment plans, protocol adjustments, and management recommendations. Their role is to provide technically grounded outputs for review by relevant clinical and technical specialists. Because such outputs may influence image acquisition, follow-up, or treatment delivery, evaluation must establish both technical validity and suitability for the intended workflow.


\noindent\textbf{Representative methods and validation demands.}
Evaluation in this family is governed by technical validity, constraint satisfaction, and workflow compatibility rather than linguistic fluency. Earlier workflows relied on rules, manual contouring, optimization engines, and institutional protocols. Deep learning automated individual steps such as organ-at-risk segmentation, dose prediction, synthetic CT, and reconstruction correction, with RAOS~\cite{Luo2024RAOS}, HaN-Seg~\cite{Podobnik2024HaNSeg}, and AAPM-RT-MAC~\cite{Cardenas2020AAPMRTMAC} providing benchmarks for anatomical delineation under planning constraints. Recent work shifts from component-level automation toward workflow orchestration. DosimeTron~\cite{tzanis2026dosimetron} automates PET/CT internal dosimetry, GPT-Plan~\cite{wang2025feasibility}, DOLA~\cite{nusrat2025autonomous}, MARTP~\cite{wang2026martp}, and SAGE~\cite{nusrat2026sage} coordinate radiotherapy or stereotactic radiosurgery planning, AgentMRI~\cite{sajua2025agentmri} selects correction models for MRI degradation, Scan-do Attitude~\cite{kang2025scan} manages CT protocols, Agent4MR~\cite{zaiss2026agentic} explores MR sequence development, INFORM-CT~\cite{Tankel2026InformCT} manages incidental findings, and the PET/CT Agent~\cite{choi2026petct} spans raw DICOM processing to structured staging support.

Validation must extend beyond component accuracy. A correct segmentation or dose estimate is insufficient if downstream constraints are not met or the output cannot be integrated into the intended workflow. DosimeTron~\cite{tzanis2026dosimetron}, for example, reports multi-scanner dosimetric validation across 597 studies from 378 patients, whereas several recent planning agents have been evaluated in small retrospective
cohorts~\cite{nusrat2025autonomous,nusrat2026sage}. Variation in study scale and design limits cross-system comparison and constrains the strength of workflow-level claims.

\noindent\textbf{Current resources and remaining gaps.} Current resources span component-level benchmarks, system-level cohorts, and workflow-level assessments~\cite{Luo2024RAOS,tzanis2026dosimetron,gu2026ct}. These settings evaluate different parts of the imaging and planning pipeline, but they are rarely linked within a common protocol. Consequently, strong performance on an isolated component does not establish reliable end-to-end operation.

The principal gap is the limited evaluation of error propagation and recovery. Most studies do not establish whether upstream failures are detected before they affect downstream decisions, or how much expert correction is required. External and prospective workflow studies are therefore needed before claims of clinical utility or deployment readiness can be justified.


\subsection{Cross-Task Evaluation and Validation}
\label{sec:eval_cross_task_validation}
\label{sec:eval_synthesis}


\noindent\textbf{Task definition and clinical role.}
Cross-task evaluation extends task-based evaluation from individual clinical outputs to systems that perform multiple tasks across anatomies, modalities, and workflow stages. Each constituent task should retain a clearly specified clinical claim, target population, imaging process, reference standard, and task-appropriate figure of merit~\cite{jha2021objective,jha2022nuclear}. Broad task coverage alone should not be interpreted as evidence of generalist competence if performance on clinically important tasks is degraded. General-Level operationalizes task-level synergy as a generalist surpassing a state-of-the-art specialist on a specific task~\cite{fei2025path}. In medicine, this principle suggests that generalist models should be compared with strong specialist or standard-of-care baselines for each clinically relevant task, using prespecified superiority or non-inferiority criteria appropriate to the intended use and clinical risk~\cite{saboury2023artificial}. Cross-task validation additionally requires case-level coherence across outputs: a report claim should be compatible with localization, a risk estimate should not contradict the described disease burden, and a tool-using workflow should expose rather than conceal upstream processing failures.

\noindent\textbf{Representative methods and validation demands.}
Evaluation in this family is often organized around benchmarks rather than individual models. M3D~\cite{bai2024m3d}, MedVL-CT69K~\cite{shui2025large}, Triad~\cite{Wang2025Triad}, MedVista3D~\cite{li2025medvista3d}, Merlin~\cite{blankemeier2026merlin}, and CuriaBench~\cite{saporta2026curia} probe broad volumetric task performance across retrieval, reporting, segmentation, registration, downstream prediction, and foundation-model transfer. Grounding- and reasoning-focused benchmarks then examine whether model outputs remain consistent with the underlying volumetric evidence. CT-SpatialVQA~\cite{monon2026ctspatialvqa} targets semantic-spatial CT reasoning; Med-StepBench~\cite{nguyen2026medstepbench} evaluates step-wise hallucination in oncological PET/CT; CORTEX~\cite{malik2026cortex} adds structured four-stage reasoning traces for chest CT; ReportQA~\cite{shi2026reportqa} evaluates report utility through QA; and Oncology VQA~\cite{liu2026oncologyvqa} tests private 3D oncology cohorts with contamination-aware blind ablations.

Agent-era evaluation adds a process dimension. MedOpenClaw~\cite{shen2026medopenclaw}, MedVistaGym~\cite{lu2026medvistagym}, ABRA~\cite{maksudov2026abra}, MedCTA~\cite{ashraf2026medcta}, RadA-BenchPlat~\cite{zheng2024well}, and RadSaFE-200~\cite{wind2026safety} evaluate full-study interaction, viewer or tool use, agent-core behavior, trajectory fidelity, and unsafe-answer behavior.

These resources broaden evaluation beyond isolated tasks, but aggregate performance remains insufficient when outputs are assessed independently. Cross-task validation should examine case-level consistency and, for agentic systems, the validity of intermediate actions and acquired evidence. Final-answer accuracy alone cannot validate a workflow built on incorrect or unsupported intermediate steps.


\noindent\textbf{Current resources and remaining gaps.}
Current benchmarks combine task-specific scores with measures of grounding, trace fidelity, workflow completion, and safety. These signals are usually reported separately, and few protocols determine whether an upstream error is propagated, detected, or corrected before it affects a downstream claim. The principal gap is therefore integrated validation across outputs and workflow stages. Contamination control, subgroup analysis, and external evaluation also remain limited, constraining claims of generality and clinical reliability. These limitations motivate the broader research agenda in Section~\ref{sec:discussion}.

The four application families operationalize the validation component of the Claim--Design--Validation alignment framework. Diagnostic claims require joint validation of spatial and language outputs~(\ref{sec:eval_diagnostic}). Prognostic claims require temporally defined, cohort-level evidence~(\ref{sec:eval_prognosis}). Planning claims require end-to-end technical and workflow validation~(\ref{sec:eval_planning_support}). Cross-task claims add the requirement that outputs and intermediate actions remain coherent~(\ref{sec:eval_cross_task_validation}). Across all four families, validation evidence often lags behind claim strength: many studies establish technical feasibility, fewer demonstrate external robustness, and only a limited number of studies evaluate workflow benefit or clinical utility.

\FloatBarrier


\section{Discussion and Future Directions}
\label{sec:discussion}

The preceding sections demonstrate that progress in this field has been driven not merely by model scaling, but by a fundamental sequence of representational and system-level shifts. We summarize this trajectory as a continuous progression: from 2D adaptation to native 3D foundation models, and ultimately toward agentic systems. This pathway aligns with the broader evolution of multimodal AI from lower-dimensional perception toward higher-dimensional, action-oriented capabilities~\cite{hu2026simulating, xiao2025comprehensive, bluethgen2025agenticradiology}.

This evolution began with a fundamental shift in how volumetric features are extracted. Early frameworks (e.g., LLaVA-Med~\cite{li2023llava}, Med-Flamingo~\cite{moor2023med}) often adapted mature 2D priors, processing volumetric data as independent slices or reformatted views. While practically accessible, this slice-level adaptation risks losing through-plane continuity and long-range anatomical context. To bridge this gap, intermediate hybrid architectures (e.g., RadFM~\cite{wu2025towards}, Photon~\cite{fang2026photon}, and OmniCT~\cite{lin2026omnict}) have explored how far mature 2D priors can be reused while maintaining a pathway for volumetric context. Ultimately, however, the field is transitioning toward native 3D foundational representations (e.g., CT2Rep~\cite{hamamci2024ct2rep}, Merlin~\cite{blankemeier2026merlin}, and Med3D-R1~\cite{lai2026med3d}) to fully preserve spatial integrity. Yet, this full-scan extraction introduces a new structural bottleneck: the sheer volume of 3D data vastly exceeds the context limits of current LLMs, necessitating advanced compression and token-routing strategies.

\begin{figure*}[!ht]
\centering
\includegraphics[width=0.98\textwidth]{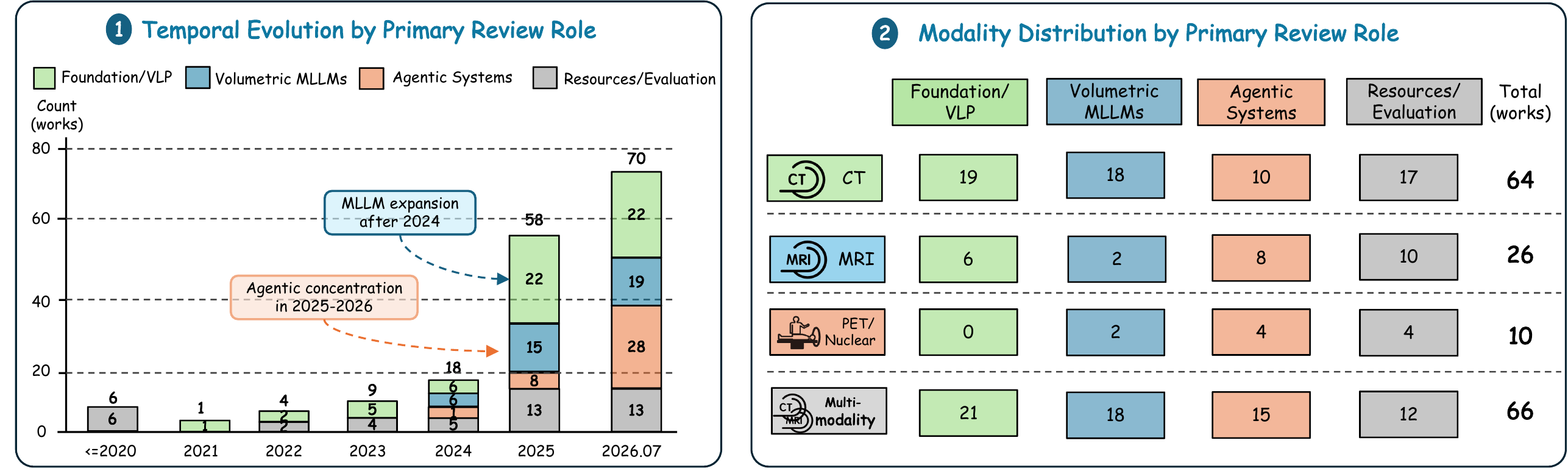}
\caption{\textbf{Core-corpus landscape of volumetric radiology AI.} The figure summarizes the deduplicated core corpus of volumetric-radiology publications and resources included in the quantitative landscape analysis. Each work is assigned once to one of four primary review roles: foundation models/vision–language pre-training, volumetric radiology MLLMs, agentic systems, or clinical resources/evaluation. The right panel cross-tabulates the same assignments by dominant modality. General AI, review articles, and 2D-only background references are excluded.}
\label{fig:full_corpus_landscape_3d_radiology}
\end{figure*}

To manage these context constraints and support the procedural complexity of volumetric interpretation, the field is increasingly extending toward agentic workflows. Rather than relying on a single forward pass, recent architectures (e.g., CT-Agent~\cite{mao2025ct}, RadAgent~\cite{roschewitz2026radagent}, MedScribe~\cite{orlando2026medscribe}) have decomposed complex radiology tasks into iterative, multi-step processes involving targeted volumetric inspection, dynamic tool invocation, and long-term memory updates. The core-corpus landscape in Figure~\ref{fig:full_corpus_landscape_3d_radiology} maps this transition along two complementary axes: temporal evolution and modality coverage. The temporal panel shows that early work established volumetric representations and evaluation resources, 3D and hybrid MLLMs expanded rapidly after 2024, and advanced agentic workflow systems became concentrated in 2025--2026. The modality panel shows that the literature remains heavily CT-centric, while sequence-aware MRI models, PET/nuclear-medicine systems, and multi-modality or general 3D resources increasingly diversify the field. Read together, these patterns indicate that agentic architectures are a recent workflow-level extension of the 3D foundation-model literature rather than an independent replacement for it. They also show that 3D foundation models are transitioning from standalone solvers into modular components of traceable clinical pipelines. Thus, future validation frameworks should shift toward effective workflow integration, verifiable tool execution bounds, and collaborative human oversight.

\begin{figure*}[t]
\centering
\includegraphics[width=0.98\textwidth]{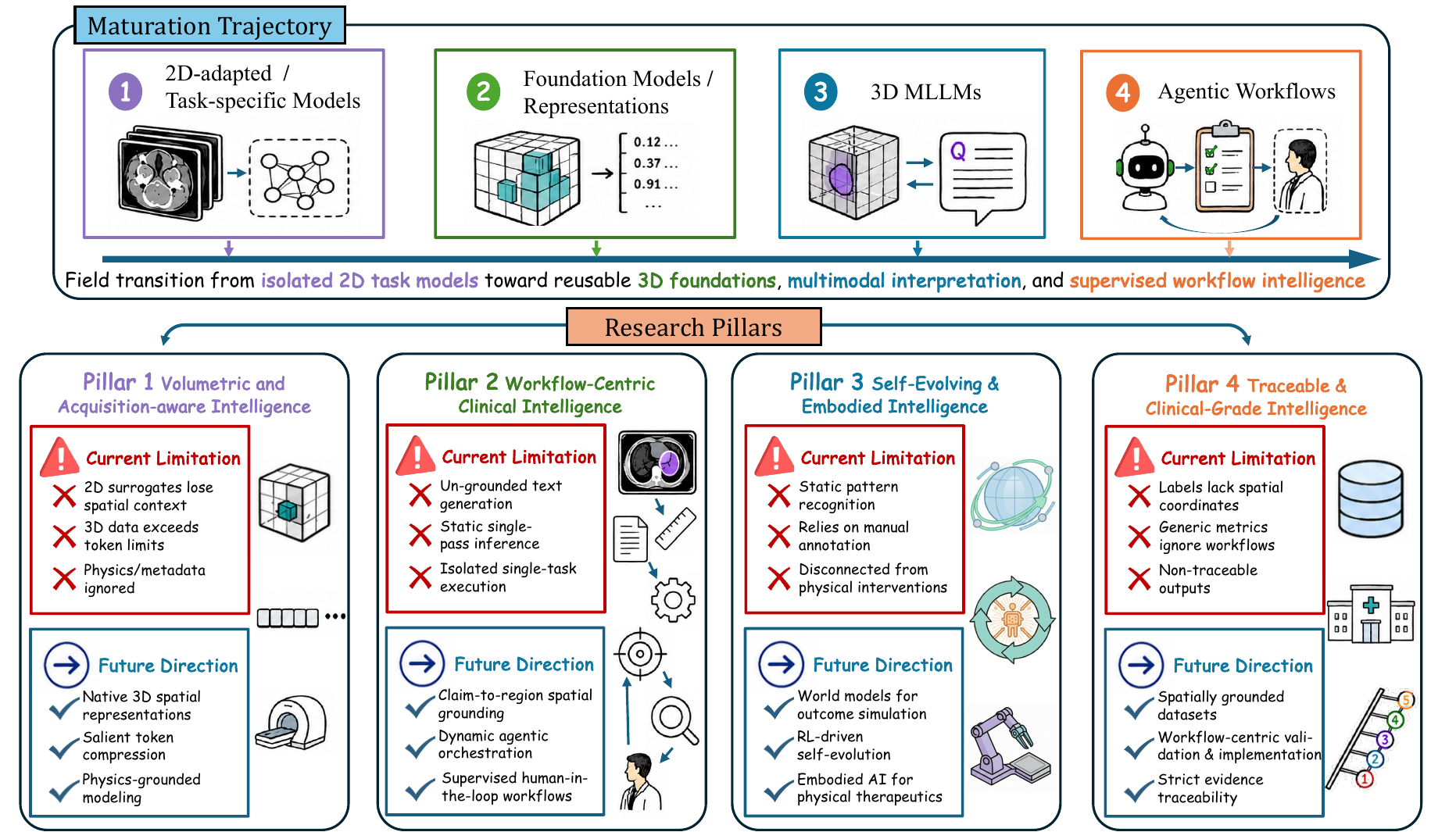}
\caption{
\textbf{A research agenda for volumetric radiology intelligence.}
The top panel illustrates the field's transition from isolated, 2D-adapted task models toward reusable 3D foundations, multimodal interpretation, and supervised agentic workflows. To drive this evolution, the bottom panel outlines a research agenda structured across four core pillars of intelligence: (1) volumetric and acquisition-aware, (2) workflow-centric clinical, (3) self-evolving and embodied, and (4) traceable and clinical-grade, contrasting current limitations with future directions.
}
\label{fig:future_roadmap_challenges}
\end{figure*}

Building upon these observations, Figure~\ref{fig:future_roadmap_challenges} summarizes our proposed research agenda. Rather than merely listing distinct challenges, this roadmap reflects a unified clinical trajectory: to systematically advance volumetric radiology, fundamental volumetric representations should be deeply coupled with scanning physics, while static multimodal interpretation should evolve into collaborative human-AI workflows. Ultimately, bridging the gap between digital diagnostic outputs and physical clinical reality will require agents to be grounded in world models, self-evolving technologies, and embodied physical infrastructure, all supported by rigorous, claim-specific data validation standards.

\subsection{From 2D Surrogates to Volumetric and Acquisition-Aware Intelligence}
\label{sec:future_modeling_principles}

Progress in 3D volumetric models depends on treating volumetric scans as native, physics-grounded structures rather than proxy modalities. The core modeling challenges involve managing informational density through clinically salient compression and unifying visual data with scanning physics metadata. This transition reflects the shifting trade-offs between computational tractability and clinical fidelity.

\textbf{Volumetric modeling should preserve native spatial continuity.} Volumetric radiology contains a combination of local details, global anatomy, and sparse, clinically decisive abnormalities. Therefore, native 3D architectures should account for this complexity rather than just scaling up 2D networks. Self-supervised methods already exploit this structure through volumetric restoration, masked modeling, spatial deformation, and geometry-aware contrastive objectives. Models such as Genesis~\cite{zhou2021models}, Swin UNETR~\cite{tang2022self}, VoCo~\cite{wu2025large}, and SPECTRE~\cite{claessens2025scaling} show that volumetric continuity should be preserved as a fundamental representation principle, rather than being treated as a limited set of 2D surrogates.

\textbf{Compression policies should optimize token efficiency and be guided by clinical salience.} 
A major bottleneck in scaling medical VLMs from 2D to 3D is efficiently encoding massive, highly redundant volumetric data into a compact visual space. 
Because full-resolution 3D information is too large for current MLLM interfaces to process, achieving low-budget tokenization for language alignment is critical. 
Consequently, current systems should adopt specific compression policies for inputs or extracted visual tokens to balance performance and computational cost. 
Approaches like M$^3$D-LaMed~\cite{bai2024m3d} use perceiver-style pooling, while Med3DVLM~\cite{xin2025med3dvlm}, MedPruner~\cite{liu2026medpruner}, and Photon~\cite{fang2026photon} make token efficiency an explicit design target. Future compression policies should be evaluated not only by aggregate task scores but also by their ability to selectively preserve small findings, spatial relations, and task-relevant regions, even when much of the background volume is discarded.

\textbf{Acquisition context should be explicitly represented alongside spatial anatomy.} CT phase, reconstruction kernel, MRI sequence, slice thickness, and scanner protocols are not incidental metadata; they are major sources of domain shift that alter the diagnostic meaning of visual patterns. Early methods point in this direction: MR-CLIP~\cite{avci2025metadata} uses DICOM metadata for semantic alignment, while M$^3$AE~\cite{liu2023m3ae} learns from multi-sequence MRI inputs. Without integrating physics and protocol constraints into the representation space, models may produce plausible language while ignoring the underlying conditions that make a radiological finding valid and safe for clinical use.

\subsection{From Static Perception to Workflow-centric Clinical Intelligence}
\label{sec:future_workflow_intelligence}

Progress in volumetric radiology models will inevitably shift from static, perception-based interpretation toward operational, workflow-centric clinical intelligence. Looking forward, we anticipate a transition organized along three main axes: establishing token-level spatial grounding, expanding into multi-step agentic orchestration, and ultimately designing for bounded, human-AI symbiotic workflows.

\textbf{Interpretability should be grounded in image regions and clinical claims.} To advance beyond current limitations, future interpretability in volumetric radiology should transition from surface-level text generation to rigorous spatial grounding. Currently, a fluent report might lack volumetric support, and correct VQA answers often lack reliable localization. While fine-grained alignment methods, such as CT-GLIP~\cite{lin2024ct}, MG-3D~\cite{ni2024mg}, RadFinder~\cite{ging2026learning}, SCALE-VLP~\cite{mahdizadeh2025scale}, MedRegion-CT~\cite{kyung2025medregion}, and discriminative-guided 3D CT grounding~\cite{wang2026enhancing}, have begun to decompose alignment across anatomy and report sentences, the critical next step is to enforce strict traceability. Future models should guarantee that every generated clinical claim can be systematically traced back to exact anatomical regions, precise measurements, or explicitly grounded visual evidence.

\textbf{Agents should evolve from passive inference to dynamic workflow orchestration.} Realizing true clinical intelligence requires transforming models into operational agents rather than relying on static perception. Current MLLM inference remains a passive exercise, constrained by the assumption that all volumetric context is captured in a single forward pass. To overcome this, architectures should grant models the autonomy to navigate 3D scans, invoke clinical tools, and iteratively refine findings. Recent developments have begun introducing these multi-step capabilities, including CT-Agent~\cite{mao2025ct} and 3DMedAgent~\cite{wang20263dmedagent} for volumetric QA; PD-CTAgent~\cite{dong2026policy} for policy-constrained phase-sufficiency decisions; RadAgent~\cite{roschewitz2026radagent}, MedScribe~\cite{orlando2026medscribe}, and Radiologist Copilot~\cite{yu2025radiologist} for report generation; and CT-Flow~\cite{gu2026ct} and MedOpenClaw~\cite{shen2026medopenclaw} for interactive navigation. However, these methods currently focus on isolated tasks or retrospective evidence-acquisition settings.
Looking ahead, research should expand these isolated, single-task systems to span the entire clinical continuum, from initial diagnosis to treatment planning and longitudinal patient management, ultimately delivering comprehensive care through structured, actionable operations~\cite{ferber2026towards} and interactive conversational support~\cite{saab2026advancing,lievin2026towards}. As evidenced by the LungIMPACT trial~\cite{woznitza2026ai}, optimizing diagnostic steps in silos fails to resolve downstream bottlenecks without parallel expansions in subsequent pathway capacities.
Therefore, future agents should be designed to orchestrate macro-workflows, dynamically harmonizing operations to optimize multi-disciplinary care delivery.

\textbf{Workflow intelligence should be anchored in supervised human-AI collaboration.} As medical agents evolve from static question-answering to dynamic, environment-aware conversational  systems~\cite{saab2026advancing,lievin2026towards,hu2025landscape}, their integration into volumetric radiology should reject unsupervised autonomy in favor of role-bounded collaboration. True intelligence requires systems that process multimodal inputs and yield inspectable results strictly under human jurisdiction. Methods like DosimeTron~\cite{tzanis2026dosimetron}, MARTP~\cite{wang2026martp}, and SAGE~\cite{nusrat2026sage} exemplify this synergy by navigating complex pipelines while deferring to clinicians during high uncertainty. Ultimately, the true measure of a clinical agent is not its capacity for independent execution, but its ability to safely amplify human expertise. By ensuring transparent reasoning and seamless handoffs, future agents will transcend mere automation to become trusted clinical copilots, bridging the gap between computational power and ultimate medical accountability.

\subsection{From Digital Boundaries to Self-Evolving and Medical Embodied Intelligence}
\label{sec:future_frontiers}

Progress in 3D medical AI should transcend static processing to embrace real-world interactions. To bridge digital diagnostics and physical interventions, research should augment clinical agents across three trajectories: integrating world models for dynamic environmental understanding, enabling continuous adaptation via self-evolving reinforcement, and grounding decisions through physical embodiment. Ultimately, these trajectories will transform agents from passive tools into active clinical participants.

\textbf{Environmental understanding should be expanded via clinical world models.} While current 3D MLLMs typically operate as static pattern discriminators, recent breakthroughs such as Brain-WM~\cite{wang2026brain} and CLARITY~\cite{ding2025clarity} challenge this bottleneck by predicting sequential MRI trajectories and simulating patient outcomes. Synthesizing these advancements reveals a paradigm shift: environmental understanding requires expansion via clinical world models. Crucially, 
this modeling can extend beyond the visual domain and be fundamentally anchored in language. Inspired by general-purpose LLMs like Qwen-AgentWorld~\cite{qwen2026agentworld}, which simulate complex interactive environments via next-state text prediction, clinical world models can leverage agentic LLMs to project patient trajectories. 
Under this setting, projecting these patient trajectories essentially constructs a simulated clinical world. Rather than acting as static predictive generators, these models serve as multi-modal sandboxes that execute a continuous ``\textit{perception-dynamics-planning}'' loop~\cite{yang2025medical,liu2026medical}.
By internalizing this anticipatory meta-reasoning, agents transition from passive observers to active planners, mentally simulating post-interventional clinical notes and evaluating treatments prior to real-world execution.


\textbf{Sequential planning should progress from structured loop engineering to self-evolution.} Complex volumetric workflows, such as longitudinal tumor tracking and multimodal image analysis, require robust architectural foundations before cognitive scaling. System development in future radiology intelligence will naturally begin with loop engineering~\cite{ridnik2024code,yao2023react,sumers2024cognitive}. Such control loops decompose complex imaging tasks into executable and verifiable steps, such as anatomical localization, cross-modal registration, and diagnostic validation, while enabling agents to iteratively correct intermediate errors through feedback. Once this structured pipeline reliably generates high-quality diagnostic trajectories, agentic RL can be integrated as the cognitive layer to optimize these volumetric decision paths~\cite{jiangmedvr}. Beyond individual trajectory optimization, autoresearch enables agents to continuously propose, evaluate, and retain improvements to their planning strategies, tools, and workflow configurations~\cite{liu2026towards,shen2026neuroclaw}. Together with agentic RL, this capability supports macro-level self-evolution. Successful clinical workflows can, for example, be distilled into reusable skill documents~\cite{zhang2026seer,sun2026experiencemakesskillfulenabling,ossowski2026codeclinic}. This structured developmental paradigm offers a highly scalable pathway to mitigate scanner-induced domain drift and expand clinical capabilities without continuous human annotation.

\textbf{Therapeutic execution should be grounded in embodied AI.} The ultimate capability scaling of clinical agents involves translating high-dimensional computational reasoning into physical action domains, effectively eliminating the boundary between passive diagnostic radiology and direct physical intervention~\cite{chen2026ai}. In an embodied paradigm, agents will seamlessly link real-time imaging modalities, such as fluoroscopy and cone-beam CT, to autonomous navigation and robotic manipulation. Specific operational scenarios include utilizing multi-agent inference pipelines to analyze real-time angiographic sequences and autonomously guide endovascular catheters through complex vascular topographies~\cite{wang2026can}. Additionally, integrating intraoperative MRI and interventional ultrasound with robotic manipulators facilitates precise needle trajectory tracking for minimally invasive percutaneous procedures~\cite{dong2022shape}. By executing closed-loop physical interventions~\cite{granados2026evolving}, including image-guided targeted tissue biopsy, radiofrequency ablation, and adaptive radiotherapy beam alignment, embodied agents will anchor digital predictions into real-world physical therapeutics, realizing true end-to-end clinical intelligence.

\subsection{From Benchmarks to Traceable and Clinical-Grade Intelligence}
\label{sec:future_data_evaluation_translation}

The transition of volumetric radiology AI into real-world workflows requires moving beyond technical feasibility. Achieving true clinical-grade intelligence relies on a holistic ecosystem comprising spatially grounded data, workflow-centric evaluation, and traceable clinical translation.

\textbf{Data infrastructure should co-evolve to support precise spatial grounding.} The shift from adapted 2D models to native 3D systems is bottlenecked by annotation granularity. Volumetric datasets like MedMNIST v2~\cite{Yang2023MedMNIST} and AbdomenAtlas~\cite{Li2025AbdomenAtlas}, alongside multimodal resources mapping scans to reports (e.g., CT-RATE~\cite{hamamci2026generalist} and M$^3$D~\cite{bai2024m3d}), are foundational but capture only a fraction of clinical diversity. Crucially, weakly supervised image-report pairs risk introducing label noise if they lack exact anatomical coordinates. Future curation should prioritize spatially grounded annotations, exemplified by RadGenome-Chest CT~\cite{Zhang2025RadGenomeChestCT}, AutoRG-Brain~\cite{Lei2024AutoRGBrain}, and SpatialMed~\cite{Trinh2026SpatialMed}. These datasets explicitly tether textual outputs to volumetric evidence, forming the bedrock for reliable AI.

\textbf{Evaluation paradigms should shift from generic metrics to workflow-centric clinical validation.} Traditional isolated metrics (e.g., Dice and AUROC) solely focus on geometric or discriminative performance but fail to guarantee workflow safety or diagnostic triage readiness. Platforms like Merlin~\cite{blankemeier2026merlin}, MedOpenClaw~\cite{shen2026medopenclaw}, and RadSaFE-200~\cite{wind2026safety} are broadening capability assessments. Furthermore, as recent guidelines emphasize~\cite{vandesande2024warrant}, evaluations should strictly align with the intended clinical function. A diagnostic tool requires rigorous calibration; a report generator demands factual consistency; and an agentic orchestrator should be judged on tool validity and human handoff efficacy~\cite{bluethgen2025agenticradiology}. Performance should be measured by fitness for the intended clinical workflow rather than benchmark dominance.

\textbf{Clinical translation should employ a staged validation ladder and strict evidence traceability.} Bridging the gap from targeted benchmarks to actual deployment requires a phased validation approach. This progression should begin with retrospective testing, advance through reader studies and workflow simulations, and culminate in prospective clinical trials and continuous post-deployment monitoring. While existing frameworks like MI-CLAIM~\cite{norgeot2020minimum}, CONSORT-AI~\cite{liu2020consort}, DECIDE-AI~\cite{vasey2022decide}, CLAIM 2024~\cite{tejani2024claim}, STARD-AI~\cite{sounderajah2025stard}, and FUTURE-AI~\cite{lekadir2025future} formalize this progression, they should be specifically adapted for volumetric radiology by incorporating image-grounded spatial traces.
Furthermore, executing this pipeline safely demands granular evidence traceability. General risks such as dataset shift~\cite{finlayson2021clinician}, algorithmic bias~\cite{obermeyer2019dissecting}, data poisoning~\cite{alber2025medical}, and cognitive bias~\cite{mahajan2025cognitive} are heavily compounded by imaging-specific vulnerabilities like scanner protocol variability and reconstruction artifacts. To mitigate these multifaceted threats, models should explicitly link every clinical output to the exact study, series, slice range, or prior examination that supported it. This transparent traceability forms the indispensable basis for debugging, safe human review, and institutional governance.

\textbf{Implementation science should be integrated across the clinical translation pathway.}
Implementation science studies the methods and contextual determinants that shape how evidence-based interventions are adopted, integrated, scaled, and sustained in routine practice. Within the AI translation pathway, technical and clinical validation assess whether a system performs as intended under defined conditions, whereas implementation science examines how that system is incorporated into a specific workflow, organization, and user environment. It therefore complements validation by addressing infrastructure, usability, human--AI interaction, organizational readiness, trust, and other determinants of routine use~\cite{fayaz2026implementation}. For volumetric radiology, hybrid effectiveness--implementation studies, which evaluate clinical effectiveness and implementation outcomes within the same study, should assess adoption, fidelity, scalability, and sustainability alongside task-specific performance and clinical outcomes, with stakeholders engaged from early development through post-deployment monitoring. In precision oncology and theranostics, the augmented-oncologist concept illustrates this perspective: implementation-ready AI and patient-specific digital twins may support clinician-led diagnosis, treatment planning, education, and longitudinal decision-making~\cite{abdollahi2026computational}. Together, these considerations extend the bounded human--AI collaboration described in Section~\ref{sec:future_workflow_intelligence} and frame implementation as an iterative process across the clinical translation pathway.

\subsection{Scope and Limitations}
\label{sec:scope_limitations}
This review focuses specifically on 3D volumetric radiology, including CT, MRI, PET, single-photon emission computed tomography (SPECT), and hybrid acquisitions, and on the technical progression from representation learning to foundation models, MLLMs, agents, evaluation, and applications. It does not attempt to provide a complete survey of all 2D medical imaging models, all non-imaging medical LLMs, or all legal, economic, and reimbursement issues related to clinical deployment. Furthermore, the literature window for this review closes in July 2026. Because the agentic literature is developing rapidly, several systems discussed are available as preprints or early reports. We therefore interpret them as indicators of emerging design patterns rather than as proof of deployment readiness.

\section{Conclusion}

Volumetric radiology AI is progressing from 2D-adapted task models toward native volumetric representations and agentic workflows. This progression comprises a representational advance that preserves clinically decisive volumetric information and aligns it with clinical language, and an operational advance that enables this evidence to be acquired, verified, and coordinated beyond a single forward pass. Selected 2D views remain effective when decisive evidence is confined to a limited set of images, whereas native volumetric modeling becomes essential when interpretation depends on full-volume spatial relationships. Agentic systems extend this foundation by coupling volumetric perception with tools, memory, and iterative workflow orchestration.

Clinical credibility depends on alignment among the intended claim, the evidence preserved by the model, the actions performed by the system, and the validation supporting that claim. The Claim--Design--Validation framework formalizes this alignment and identifies traceability and clearly defined human oversight as cross-cutting requirements. Future progress will depend on stronger native volumetric data resources, more reliable grounding and validation, and tighter integration with clinical tools and longitudinal context. As these components mature, volumetric radiology systems may evolve into traceable, human-supervised systems that support radiologists across diagnostic, planning, and longitudinal workflows.

\clearpage

\bibliography{custom}

@article{kelly2022radiology,
  title={Radiology artificial intelligence: a systematic review and evaluation of methods ({RAISE})},
  author={Kelly, Brendan S and Judge, Conor and Bollard, Stephanie M and Clifford, Simon M and Healy, Gerard M and Aziz, Awsam and Mathur, Prateek and Islam, Shah and Yeom, Kristen W and Lawlor, Aonghus and others},
  journal={European Radiology},
  volume={32},
  number={11},
  pages={7998--8007},
  year={2022},
  publisher={Springer},
}

@article{doshi2024quantitative,
  title={Quantitative Evaluation of Large Language Models to Streamline Radiology Report Impressions: A Multimodal Retrospective Analysis},
  author={Doshi, Rushabh and Amin, Kanhai S and Khosla, Pavan and Bajaj, Simar S and Chheang, Sophie and Forman, Howard P},
  journal={Radiology},
  volume={310},
  number={3},
  pages={e231593},
  year={2024},
  publisher={Radiological Society of North America},
}

@article{salmanpour2026handcrafted,
  title={Handcrafted vs. Deep Radiomics vs. Fusion vs. Deep Learning: A Comprehensive Review of Machine Learning -Based Cancer Outcome Prediction in {PET} and {SPECT} Imaging},
  author={Salmanpour, Mohammad R and Mehrnia, Somayeh Sadat and Jabarzadeh Ghandilu, Sajad and Safahi, Zhino and Falahati, Sonya and Taeb, Shahram and Mousavi, Ghazal and Maghsudi, Mehdi and Shariftabrizi, Ahmad and Hacihaliloglu, Ilker and others},
  journal={Journal of Imaging Informatics in Medicine},
  pages={1--50},
  year={2026},
  publisher={Springer},
}

@article{holm2026explainable,
  title={Explainable {AI} in nuclear medicine},
  author={Holm, Sune and Ferrara, Daria and Pepponi, Miriam and Abenavoli, Elisabetta and Frille, Armin and Duke, Shaul and Gr{\"u}nert, Stefan and Hacker, Marcus and Hennig, Bengt and Hesse, Swen and others},
  journal={European Journal of Nuclear Medicine and Molecular Imaging},
  volume={53},
  number={4},
  pages={2648--2651},
  year={2026},
  publisher={Springer},
}

@article{jha2022nuclear,
  title={Nuclear Medicine and Artificial Intelligence: Best Practices for Evaluation (the {RELAINCE} Guidelines)},
  author={Jha, Abhinav K and Bradshaw, Tyler J and Buvat, Ir{\`e}ne and Hatt, Mathieu and Kc, Prabhat and Liu, Chi and Obuchowski, Nancy F and Saboury, Babak and Slomka, Piotr J and Sunderland, John J and others},
  journal={Journal of Nuclear Medicine},
  volume={63},
  number={9},
  pages={1288--1299},
  year={2022},
  publisher={Society of Nuclear Medicine},
}

@article{saboury2023artificial,
  title={Artificial Intelligence in Nuclear Medicine: Opportunities, Challenges, and Responsibilities Toward a Trustworthy Ecosystem},
  author={Saboury, Babak and Bradshaw, Tyler and Boellaard, Ronald and Buvat, Ir{\`e}ne and Dutta, Joyita and Hatt, Mathieu and Jha, Abhinav K and Li, Quanzheng and Liu, Chi and McMeekin, Helena and others},
  journal={Journal of Nuclear Medicine},
  volume={64},
  number={2},
  pages={188--196},
  year={2023},
  publisher={Society of Nuclear Medicine},
}

@article{jha2021objective,
  title={Objective task-based evaluation of artificial intelligence-based medical imaging methods: framework, strategies, and role of the physician},
  author={Jha, Abhinav K and Myers, Kyle J and Obuchowski, Nancy A and Liu, Ziping and Rahman, Md Ashequr and Saboury, Babak and Rahmim, Arman and Siegel, Barry A},
  journal={PET Clinics},
  volume={16},
  number={4},
  pages={493--511},
  year={2021},
  publisher={Elsevier},
}

@inproceedings{fei2025path,
  title={On Path to Multimodal Generalist: {General-Level} and {General-Bench}},
  author={Fei, Hao and Zhou, Yuan and Li, Juncheng and Li, Xiangtai and Xu, Qingshan and Li, Bobo and Wu, Shengqiong and Wang, Yaoting and Zhou, Junbao and Meng, Jiahao and others},
  booktitle={Proceedings of the 42nd International Conference on Machine Learning},
  year={2025},
  volume={267},
  pages={16423--16542},
}

@article{salmanpour2026clinically,
  title={A clinically anchored radiomics dictionary for explainable {TI-RADS}--based thyroid nodule classification in ultrasound; dictionary version {TU1.0}},
  author={Salmanpour, Mohammad and Taeb, Shahram and Jouzdani, Ali Fathi and Ayazi, Mohammad and Saffarian, Siavash Hosseinpour and Maghsudi, Mehdi and Hacihaliloglu, Ilker and Rahmim, Arman},
  journal={European Journal of Radiology},
  pages={113014},
  year={2026},
  publisher={Elsevier},
  volume={203},
}

@article{salmanpour2026radiological,
  title={Radiological and Biological Dictionary of Radiomics Features: Addressing Understandable {AI} Issues in Personalized Prostate Cancer, Dictionary Version {PM1.0}},
  author={Salmanpour, Mohammad R and Amiri, Sajad and Gharibi, Sara and Shariftabrizi, Ahmad and Xu, Yixi and Weeks, William B and Rahmim, Arman and Hacihaliloglu, Ilker},
  journal={Journal of Imaging Informatics in Medicine},
  volume={39},
  number={3},
  pages={1929--1950},
  year={2026},
  publisher={Springer},
}

@article{jouzdani2026towards,
  title={Towards interpretable {AI} in personalized medicine through a radiological-biological radiomics dictionary linking semantic {Lung-RADS} and imaging radiomics features},
  author={Jouzdani, Ali Fathi and Taeb, Shahram and Maghsudi, Mehdi and Gorji, Arman and Rahmim, Arman and Salmanpour, Mohammad R},
  journal={Journal of Biomedical Informatics},
  pages={105047},
  year={2026},
  publisher={Elsevier},
  volume={179},
}

@article{gorji2026radiological,
  title={Radiological and biological dictionary of radiomics features: addressing understandable {AI} issues in personalized breast cancer; dictionary version {BM1.0}},
  author={Gorji, Arman and Sanati, Nima and Pouria, Amir Hossein and Mehrnia, Somayeh Sadat and Hacihaliloglu, Ilker and Rahmim, Arman and Salmanpour, Mohammad R},
  journal={Physics in Medicine \& Biology},
  volume={71},
  number={2},
  pages={025008},
  year={2026},
  publisher={IOP Publishing},
}

@article{lambin2012radiomics,
  title={Radiomics: Extracting more information from medical images using advanced feature analysis},
  author={Lambin, Philippe and Rios-Velazquez, Emmanuel and Leijenaar, Ralph and Carvalho, Sara and Van Stiphout, Ruud GPM and Granton, Patrick and Zegers, Catharina ML and Gillies, Robert and Boellard, Ronald and Dekker, Andr{\'e} and others},
  journal={European Journal of Cancer},
  volume={48},
  number={4},
  pages={441--446},
  year={2012},
  publisher={Elsevier},
}

@article{gillies2016radiomics,
  title={Radiomics: Images Are More than Pictures, They Are Data},
  author={Gillies, Robert J and Kinahan, Paul E and Hricak, Hedvig},
  journal={Radiology},
  volume={278},
  number={2},
  pages={563--577},
  year={2016},
  publisher={Radiological Society of North America},
}

@inproceedings{cicek20163d,
  title={{3D} {U-Net}: Learning Dense Volumetric Segmentation from Sparse Annotation},
  author={Cicek, Ozgun and Abdulkadir, Ahmed and Lienkamp, Soeren S and Brox, Thomas and Ronneberger, Olaf},
  booktitle={International Conference on Medical Image Computing and Computer-Assisted Intervention},
  pages={424--432},
  year={2016},
  organization={Springer},
}

@inproceedings{milletari2016vnet,
  title={{V-Net}: Fully Convolutional Neural Networks for Volumetric Medical Image Segmentation},
  author={Milletari, Fausto and Navab, Nassir and Ahmadi, Seyed-Ahmad},
  booktitle={2016 Fourth International Conference on 3D Vision},
  pages={565--571},
  year={2016},
  organization={IEEE},
}

@article{kamnitsas2017efficient,
  title={Efficient multi-scale {3D} {CNN} with fully connected {CRF} for accurate brain lesion segmentation},
  author={Kamnitsas, Konstantinos and Ledig, Christian and Newcombe, Virginia FJ and Simpson, Joanna P and Kane, Andrew D and Menon, David K and Rueckert, Daniel and Glocker, Ben},
  journal={Medical Image Analysis},
  volume={36},
  pages={61--78},
  year={2017},
  publisher={Elsevier},
}

@article{isensee2021nnu,
  title={{nnU-Net}: a self-configuring method for deep learning-based biomedical image segmentation},
  author={Isensee, Fabian and Jaeger, Paul F and Kohl, Simon AA and Petersen, Jens and Maier-Hein, Klaus H},
  journal={Nature Methods},
  volume={18},
  number={2},
  pages={203--211},
  year={2021},
  publisher={Nature Publishing Group US New York},
}

@inproceedings{hatamizadeh2022unetr,
  title={{UNETR}: Transformers for {3D} Medical Image Segmentation},
  author={Hatamizadeh, Ali and Tang, Yucheng and Nath, Vishwesh and Yang, Dong and Myronenko, Andriy and Landman, Bennett and Roth, Holger R and Xu, Daguang},
  booktitle={Proceedings of the IEEE/CVF Winter Conference on Applications of Computer Vision},
  pages={574--584},
  year={2022},
}

@inproceedings{vaswani2017attention,
  title={Attention is all you need},
  author={Vaswani, Ashish and Shazeer, Noam and Parmar, Niki and Uszkoreit, Jakob and Jones, Llion and Gomez, Aidan N and Kaiser, Lukasz and Polosukhin, Illia},
  booktitle={Advances in Neural Information Processing Systems},
  volume={30},
  year={2017}
}

@inproceedings{brown2020language,
  title={Language models are few-shot learners},
  author={Brown, Tom and Mann, Benjamin and Ryder, Nick and Subbiah, Melanie and Kaplan, Jared D and Dhariwal, Prafulla and Neelakantan, Arvind and Shyam, Pranav and Sastry, Girish and Askell, Amanda and others},
  booktitle={Advances in Neural Information Processing Systems},
  volume={33},
  pages={1877--1901},
  year={2020}
}

@inproceedings{ouyang2022training,
  title={Training Language Models to Follow Instructions with Human Feedback},
  author={Ouyang, Long and Wu, Jeffrey and Jiang, Xu and Almeida, Diogo and Wainwright, Carroll and Mishkin, Pamela and Zhang, Chong and Agarwal, Sandhini and Slama, Katarina and Ray, Alex and others},
  booktitle={Advances in Neural Information Processing Systems},
  volume={35},
  pages={27730--27744},
  year={2022},
}

@inproceedings{wei2022chain,
  title={Chain-Of-Thought Prompting Elicits Reasoning in Large Language Models},
  author={Wei, Jason and Wang, Xuezhi and Schuurmans, Dale and Bosma, Maarten and Ichter, Brian and Xia, Fei and Chi, Ed and Le, Quoc and Zhou, Denny},
  booktitle={Advances in Neural Information Processing Systems},
  volume={35},
  pages={24824--24837},
  year={2022},
}

@article{ossowski2026codeclinic,
  title={{CodeClinic}: Evaluating Automation of Coding Skills for Clinical Reasoning Agents},
  author={Ossowski, Timothy and Liu, Xinchi and Maqbool, Danyal and Dhanuka, Vaibhav and Zhang, Sheng and Poon, Hoifung and Afshar, Majid and Bradshaw, Tyler and Hu, Junjie},
  journal={arXiv preprint arXiv:2605.09675},
  year={2026}
}

@inproceedings{yao2023react,
  title={{ReAct}: Synergizing reasoning and acting in language models},
  author={Yao, Shunyu and Zhao, Jeffrey and Yu, Dian and Du, Nan and Shafran, Izhak and Narasimhan, Karthik and Cao, Yuan},
  booktitle={International Conference on Learning Representations},
  year={2023}
}

@article{sumers2024cognitive,
  title={Cognitive Architectures for Language Agents},
  author={Sumers, Theodore R and Yao, Shunyu and Narasimhan, Karthik and Griffiths, Thomas L},
  journal={Transactions on Machine Learning Research},
  volume={2024},
  year={2024},
  publisher={Transactions on Machine Learning Research}
}

@article{xiao2025comprehensive,
  title={A comprehensive survey of large language models and multimodal large language models in medicine},
  author={Xiao, Hanguang and Zhou, Feizhong and Liu, Xingyue and Liu, Tianqi and Li, Zhipeng and Liu, Xin and Huang, Xiaoxuan},
  journal={Information Fusion},
  volume={117},
  pages={102888},
  year={2025},
  publisher={Elsevier},
}

@article{xiao2026medical,
  title={Medical multimodal large language models: A survey},
  author={Xiao, Hanguang and Hui, Ningzhi and Xu, Yong and Li, Zhipeng and Peng, Jincheng},
  journal={Information Fusion},
  pages={104386},
  year={2026},
  publisher={Elsevier},
  volume={134},
}

@article{lee2025multimodal,
  title={Multimodal generative {AI} for interpreting {3D} medical images and videos},
  author={Lee, Jung-Oh and Zhou, Hong-Yu and Berzin, Tyler M and Sodickson, Daniel K and Rajpurkar, Pranav},
  journal={npj Digital Medicine},
  volume={8},
  number={1},
  pages={273},
  year={2025},
  publisher={Nature Publishing Group UK London},
}

@article{wu2025vision,
  title={Vision-language foundation model for {3D} medical imaging},
  author={Wu, Jing and Wang, Yuli and Zhong, Zhusi and Liao, Weihua and Trayanova, Natalia and Jiao, Zhicheng and Bai, Harrison X},
  journal={npj Artificial Intelligence},
  volume={1},
  number={1},
  pages={17},
  year={2025},
  publisher={Nature Publishing Group UK London},
}

@article{zhou2021models,
  title={Models Genesis},
  author={Zhou, Zongwei and Sodha, Vatsal and Pang, Jiaxuan and Gotway, Michael B and Liang, Jianming},
  journal={Medical Image Analysis},
  volume={67},
  pages={101840},
  year={2021},
  publisher={Elsevier},
}

@inproceedings{xie2022unimiss,
  title={{UniMiSS}: Universal Medical Self-supervised Learning via Breaking Dimensionality Barrier},
  author={Xie, Yutong and Zhang, Jianpeng and Xia, Yong and Wu, Qi},
  booktitle={European Conference on Computer Vision},
  pages={558--575},
  year={2022},
  organization={Springer},
}

@inproceedings{tang2022self,
  title={Self-Supervised Pre-Training of {Swin} Transformers for {3D} Medical Image Analysis},
  author={Tang, Yucheng and Yang, Dong and Li, Wenqi and Roth, Holger R and Landman, Bennett and Xu, Daguang and Nath, Vishwesh and Hatamizadeh, Ali},
  booktitle={Proceedings of the IEEE/CVF Conference on Computer Vision and Pattern Recognition},
  pages={20730--20740},
  year={2022},
}

@article{zhou2023unified,
  title={A Unified Visual Information Preservation Framework for Self-supervised Pre-training in Medical Image Analysis},
  author={Zhou, Hong-Yu and Lu, Chixiang and Chen, Chaoqi and Yang, Sibei and Yu, Yizhou},
  journal={IEEE Transactions on Pattern Analysis and Machine Intelligence},
  volume={45},
  number={7},
  pages={8020--8035},
  year={2023},
  publisher={IEEE},
}

@inproceedings{chen2023masked,
  title={Masked Image Modeling Advances {3D} Medical Image Analysis},
  author={Chen, Zekai and Agarwal, Devansh and Aggarwal, Kshitij and Safta, Wiem and Balan, Mariann Micsinai and Brown, Kevin},
  booktitle={Proceedings of the IEEE/CVF Winter Conference on Applications of Computer Vision},
  pages={1970--1980},
  year={2023},
}

@inproceedings{he2023geometric,
  title={Geometric Visual Similarity Learning in {3D} Medical Image Self-Supervised Pre-Training},
  author={He, Yuting and Yang, Guanyu and Ge, Rongjun and Chen, Yang and Coatrieux, Jean-Louis and Wang, Boyu and Li, Shuo},
  booktitle={Proceedings of the IEEE/CVF Conference on Computer Vision and Pattern Recognition},
  pages={9538--9547},
  year={2023},
}

@article{liu2023m3ae,
  title={{M3AE}: Multimodal Representation Learning for Brain Tumor Segmentation with Missing Modalities},
  author={Liu, Hong and Wei, Dong and Lu, Donghuan and Sun, Jinghan and Wang, Liansheng and Zheng, Yefeng},
  volume={37},
  pages={1657--1665},
  year={2023},
  number={2},
  journal={Proceedings of the AAAI Conference on Artificial Intelligence},
}

@article{zhuang2025mim,
  title={{MiM}: Mask in Mask Self-Supervised Pre-Training for {3D} Medical Image Analysis},
  author={Zhuang, Jiaxin and Wu, Linshan and Wang, Qiong and Fei, Peng and Vardhanabhuti, Varut and Luo, Lin and Chen, Hao},
  journal={IEEE Transactions on Medical Imaging},
  year={2025},
  publisher={IEEE},
  volume={44},
  number={9},
  pages={3727--3740},
}

@article{xing2024hybrid,
  title={Hybrid Masked Image Modeling for {3D} Medical Image Segmentation},
  author={Xing, Zhaohu and Zhu, Lei and Yu, Lequan and Xing, Zhiheng and Wan, Liang},
  journal={IEEE Journal of Biomedical and Health Informatics},
  volume={28},
  number={4},
  pages={2115--2125},
  year={2024},
  publisher={IEEE},
}

@article{wu2025large,
  title={Large-Scale {3D} Medical Image Pre-Training With Geometric Context Priors},
  author={Wu, Linshan and Zhuang, Jiaxin and Chen, Hao},
  journal={IEEE Transactions on Pattern Analysis and Machine Intelligence},
  year={2025},
  publisher={IEEE},
  volume={48},
  number={3},
  pages={3801--3818},
}

@inproceedings{gao2024cross,
  title={Cross-dimensional Medical Self-supervised Representation Learning Based on a Pseudo-{3D} Transformation},
  author={Gao, Fei and Wang, Siwen and Zhang, Fandong and Zhou, Hong-Yu and Wang, Yizhou and Wang, Churan and Yu, Gang and Yu, Yizhou},
  booktitle={International Conference on Medical Image Computing and Computer-Assisted Intervention},
  pages={178--188},
  year={2024},
  organization={Springer},
}

@article{lyu2024masked,
  title={Masked Deformation Modeling for Volumetric Brain {MRI} Self-Supervised Pre-Training},
  author={Lyu, Junyan and Bartlett, Perry F and Nasrallah, Fatima A and Tang, Xiaoying},
  journal={IEEE Transactions on Medical Imaging},
  volume={44},
  number={3},
  pages={1596--1607},
  year={2024},
  publisher={IEEE},
}

@inproceedings{valanarasu2024disruptive,
  title={Disruptive Autoencoders: Leveraging Low-level features for {3D} Medical Image Pre-training},
  author={Valanarasu, Jeya Maria Jose and Tang, Yucheng and Yang, Dong and Xu, Ziyue and Zhao, Can and Li, Wenqi and Patel, Vishal M and Landman, Bennett Allan and Xu, Daguang and He, Yufan and others},
  booktitle={Medical Imaging with Deep Learning},
  pages={1553--1570},
  year={2024},
  organization={PMLR},
  volume={250},
}

@inproceedings{chalcroft2025unified,
  title={Unified {3D} {MRI} Representations via Sequence-Invariant Contrastive Learning},
  author={Chalcroft, Liam and Crinion, Jenny and J. Price, Cathy and Ashburner, John},
  booktitle={International Workshop on Simulation and Synthesis in Medical Imaging},
  pages={63--74},
  year={2025},
  organization={Springer},
}

@article{wang2025improving,
  title={Improving Self-Supervised Medical Image Pre-Training by Early Alignment With Human Eye Gaze Information},
  author={Wang, Sheng and Zhao, Zihao and Shen, Zhenrong and Wang, Bin and Wang, Qian and Shen, Dinggang},
  journal={IEEE Transactions on Medical Imaging},
  volume={44},
  number={10},
  pages={4063--4072},
  year={2025},
  publisher={IEEE},
}

@article{zhu20253d,
  title={{3D} Foundation Model for Generalizable Disease Detection in Head Computed Tomography},
  author={Zhu, Weicheng and Huang, Haoxu and Tang, Huanze and Musthyala, Rushabh and Yu, Boyang and Chen, Long and Vega, Emilio and O'Donnell, Thomas and Dehkharghani, Seena and Frontera, Jennifer A and others},
  journal={arXiv preprint arXiv:2502.02779},
  year={2025}
}

@inproceedings{rui2025multi,
  title={Multi-modal Vision Pre-training for Medical Image Analysis},
  author={Rui, Shaohao and Chen, Lingzhi and Tang, Zhenyu and Wang, Lilong and Liu, Mianxin and Zhang, Shaoting and Wang, Xiaosong},
  booktitle={Proceedings of the Computer Vision and Pattern Recognition Conference},
  pages={5164--5174},
  year={2025},
}

@inproceedings{ye2024continual,
  title={Continual Self-Supervised Learning: Towards Universal Multi-Modal Medical Data Representation Learning},
  author={Ye, Yiwen and Xie, Yutong and Zhang, Jianpeng and Chen, Ziyang and Wu, Qi and Xia, Yong},
  booktitle={Proceedings of the IEEE/CVF Conference on Computer Vision and Pattern Recognition},
  pages={11114--11124},
  year={2024},
}

@article{claessens2025scaling,
  title={Scaling Self-Supervised and Cross-Modal Pretraining for Volumetric {CT} Transformers},
  author={Claessens, Cris and Viviers, Christiaan and D'Amicantonio, Giacomo and Bondarev, Egor and van der Sommen, Fons},
  journal={arXiv preprint arXiv:2511.17209},
  year={2025}
}

@inproceedings{lee2025hu,
  title={{HU}-Based Foreground Masking for {3D} Medical Masked Image Modeling},
  author={Lee, Jin and Dang, Vu and Yu, Gwang-Hyun and Le, Anh and Rahman, Zahid and Jang, Jin-Ho and Lee, Heonzoo and Kim, Kun-Yung and Kim, Jin-Sul and Kim, Jin-Young},
  booktitle={International Workshop on Applications of Medical AI},
  pages={122--131},
  year={2025},
  organization={Springer},
}

@article{xu2025generalizable,
  title={A generalizable {3D} framework and model for self-supervised learning in medical imaging},
  author={Xu, Tony and Hosseini, Sepehr and Anderson, Chris and Rinaldi, Anthony and Krishnan, Rahul G and Martel, Anne L and Goubran, Maged},
  journal={npj Digital Medicine},
  volume={8},
  number={1},
  pages={639},
  year={2025},
  publisher={Nature Publishing Group UK London},
}

@article{saporta2026curia,
  title={{Curia-2}: Scaling Self-Supervised Learning for Radiology Foundation Models},
  author={Saporta, Antoine and Callard, Baptiste and Dancette, Corentin and Khlaut, Julien and Corbi{\`e}re, Charles and Butsanets, Leo and Prat, Amaury and Manceron, Pierre},
  journal={arXiv preprint arXiv:2604.01987},
  year={2026}
}

@article{wang2025towards,
  title={Towards a general-purpose foundation model for functional MRI analysis},
  author={Wang, Cheng and Jiang, Yu and Peng, Zhihao and Li, Chenxin and Bang, Chang-bae and Zhao, Lin and Fu, Wanyi and Lv, Jinglei and Sepulcre, Jorge and Yang, Carl and others},
  journal={Nature Biomedical Engineering},
  pages={1--12},
  year={2026},
  publisher={Nature Publishing Group UK London}
}

@inproceedings{he2022masked,
  title={Masked Autoencoders Are Scalable Vision Learners},
  author={He, Kaiming and Chen, Xinlei and Xie, Saining and Li, Yanghao and Doll{\'a}r, Piotr and Girshick, Ross},
  booktitle={Proceedings of the IEEE/CVF Conference on Computer Vision and Pattern Recognition},
  pages={16000--16009},
  year={2022},
}

@inproceedings{he2020momentum,
  title={Momentum Contrast for Unsupervised Visual Representation Learning},
  author={He, Kaiming and Fan, Haoqi and Wu, Yuxin and Xie, Saining and Girshick, Ross},
  booktitle={Proceedings of the IEEE/CVF Conference on Computer Vision and Pattern Recognition},
  pages={9729--9738},
  year={2020},
}

@inproceedings{wang2022medclip,
  title={{MedCLIP}: Contrastive Learning from Unpaired Medical Images and Text},
  author={Wang, Zifeng and Wu, Zhenbang and Agarwal, Dinesh and Sun, Jimeng},
  booktitle={Proceedings of the 2022 Conference on Empirical Methods in Natural Language Processing},
  pages={3876--3887},
  year={2022},
}

@inproceedings{eslami2023pubmedclip,
  title={{PubMedCLIP}: How Much Does {CLIP} Benefit Visual Question Answering in the Medical Domain?},
  author={Eslami, Sedigheh and Meinel, Christoph and De Melo, Gerard},
  booktitle={Findings of the Association for Computational Linguistics: EACL 2023},
  pages={1181--1193},
  year={2023},
}

@inproceedings{lei2023clip,
  title={{CLIP-Lung}: Textual Knowledge-Guided Lung Nodule Malignancy Prediction},
  author={Lei, Yiming and Li, Zilong and Shen, Yan and Zhang, Junping and Shan, Hongming},
  booktitle={International Conference on Medical Image Computing and Computer-Assisted Intervention},
  pages={403--412},
  year={2023},
  organization={Springer},
}

@article{zhang2023biomedclip,
  title={{BiomedCLIP}: a multimodal biomedical foundation model pretrained from fifteen million scientific image-text pairs},
  author={Zhang, Sheng and Xu, Yanbo and Usuyama, Naoto and Xu, Hanwen and Bagga, Jaspreet and Tinn, Robert and Preston, Sam and Rao, Rajesh and Wei, Mu and Valluri, Naveen and others},
  journal={arXiv preprint arXiv:2303.00915},
  year={2023}
}

@inproceedings{lin2023pmc,
  title={{PMC-CLIP}: Contrastive Language-Image Pre-training Using Biomedical Documents},
  author={Lin, Weixiong and Zhao, Ziheng and Zhang, Xiaoman and Wu, Chaoyi and Zhang, Ya and Wang, Yanfeng and Xie, Weidi},
  booktitle={International Conference on Medical Image Computing and Computer-Assisted Intervention},
  pages={525--536},
  year={2023},
  organization={Springer},
}

@article{khattak2024unimed,
  title={{UniMed-CLIP}: Towards a unified image-text pretraining paradigm for diverse medical imaging modalities},
  author={Khattak, Muhammad Uzair and Kunhimon, Shahina and Naseer, Muzammal and Khan, Salman and Khan, Fahad Shahbaz},
  journal={arXiv preprint arXiv:2412.10372},
  year={2024}
}

@article{nie2025conceptclip,
  title={{ConceptCLIP}: Towards trustworthy medical {AI} via concept-enhanced contrastive langauge-image pre-training},
  author={Nie, Yuxiang and He, Sunan and Bie, Yequan and Wang, Yihui and Chen, Zhixuan and Yang, Shu and Chen, Hao},
  journal={arXiv e-prints},
  pages={arXiv--2501},
  year={2025}
}

@article{lei2025unibrain,
  title={{UniBrain}: Universal Brain {MRI} diagnosis with hierarchical knowledge-enhanced pre-training},
  author={Lei, Jiayu and Dai, Lisong and Jiang, Haoyun and Wu, Chaoyi and Zhang, Xiaoman and Zhang, Yao and Yao, Jiangchao and Xie, Weidi and Zhang, Yanyong and Li, Yuehua and others},
  journal={Computerized Medical Imaging and Graphics},
  volume={122},
  pages={102516},
  year={2025},
  publisher={Elsevier},
}

@inproceedings{hamamci2024ct2rep,
  title={{CT2Rep}: Automated Radiology Report Generation for {3D} Medical Imaging},
  author={Hamamci, Ibrahim Ethem and Er, Sezgin and Menze, Bjoern},
  booktitle={International Conference on Medical Image Computing and Computer-Assisted Intervention},
  pages={476--486},
  year={2024},
  organization={Springer},
}

@article{hamamci2026generalist,
  title={Generalist foundation models from a multimodal dataset for {3D} computed tomography},
  author={Hamamci, Ibrahim Ethem and Er, Sezgin and Wang, Chenyu and Almas, Furkan and Simsek, Ayse Gulnihan and Esirgun, Sevval Nil and Dogan, Irem and Durugol, Omer Faruk and Hou, Benjamin and Shit, Suprosanna and others},
  journal={Nature Biomedical Engineering},
  pages={1--19},
  year={2026},
  publisher={Nature Publishing Group UK London},
}

@article{lin2024ct,
  title={{CT-GLIP}: {3D} grounded language-image pretraining with {CT} scans and radiology reports for full-body scenarios},
  author={Lin, Jingyang and Xia, Yingda and Zhang, Jianpeng and Yan, Ke and Cao, Kai and Lu, Le and Luo, Jiebo and Zhang, Ling},
  journal={arXiv preprint arXiv:2404.15272},
  year={2024}
}

@article{beeche2025pan,
  title={A Pan-Organ Vision-Language Model for Generalizable {3D} {CT} Representations},
  author={Beeche, Cameron and Kim, Joonghyun and Tavolinejad, Hamed and Zhao, Bingxin and Sharma, Rakesh and Duda, Jeffrey and Gee, James and Dako, Farouk and Verma, Anurag and Morse, Colleen and others},
  journal={medRxiv},
  year={2025}
}

@article{li2025towards,
  title={Towards universal text-driven {CT} image segmentation},
  author={Li, Yuheng and Lai, Yuxiang and Thor, Maria and Marshall, Deborah and Buchwald, Zachary and Yu, David S and Yang, Xiaofeng},
  journal={arXiv preprint arXiv:2503.06030},
  year={2025}
}

@article{shui2025large,
  title={Large-scale and fine-grained vision-language pre-training for enhanced {CT} image understanding},
  author={Shui, Zhongyi and Zhang, Jianpeng and Cao, Weiwei and Wang, Sinuo and Guo, Ruizhe and Lu, Le and Yang, Lin and Ye, Xianghua and Liang, Tingbo and Zhang, Qi and others},
  journal={arXiv preprint arXiv:2501.14548},
  year={2025}
}

@article{zhao2025towards,
  title={Towards Scalable Language-Image Pre-training for {3D} Medical Imaging},
  author={Zhao, Chenhui and Lyu, Yiwei and Chowdury, Asadur and Harake, Edward and Kondepudi, Akhil and Rao, Akshay and Hou, Xinhai and Lee, Honglak and Hollon, Todd},
  journal={arXiv preprint arXiv:2505.21862},
  year={2025}
}

@inproceedings{park2025radzero3d,
  title={{RadZero3D}: Bridging Self-Supervised Video Models and Medical Vision-Language Alignment for Zero-Shot Chest {CT} Interpretation},
  author={Park, Jonggwon and Choi, Kyoyun and Yoon, Byungmu and Cho, Hong Geun and Hwang, Bumcheol},
  booktitle={Proceedings of the IEEE/CVF International Conference on Computer Vision},
  pages={6742--6749},
  year={2025},
}

@inproceedings{liu2025t3d,
  title={{T3D}: Advancing {3D} Medical Vision-Language Pre-Training by Learning Multi-View Visual Consistency},
  author={Liu, Che and Ouyang, Cheng and Chen, Yinda and Quilodr{\'a}n-Casas, C{\'e}sar and Ma, Lei and Fu, Jie and Guo, Yike and Shah, Anand and Bai, Wenjia and Arcucci, Rossella},
  booktitle={Proceedings of the IEEE/CVF International Conference on Computer Vision},
  pages={6704--6714},
  year={2025},
}

@inproceedings{cao2025boosting,
  title={Boosting Vision Semantic Density with Anatomy Normality Modeling for Medical Vision-Language Pre-Training},
  author={Cao, Weiwei and Zhang, Jianpeng and Shui, Zhongyi and Wang, Sinuo and Chen, Zeli and Li, Xi and Lu, Le and Ye, Xianghua and Zhang, Qi and Liang, Tingbo and others},
  booktitle={Proceedings of the IEEE/CVF International Conference on Computer Vision},
  pages={23041--23050},
  year={2025},
}

@article{zhang2025velvet,
  title={{VELVET-Med}: Vision and Efficient Language Pre-training for Volumetric Imaging Tasks in Medicine},
  author={Zhang, Ziyang and Yu, Yang and Yang, Xulei and Yeo, Si Yong},
  journal={arXiv preprint arXiv:2508.12108},
  year={2025}
}

@article{li2025medvista3d,
  title={{MedVista3D}: Vision-Language Modeling for Reducing Diagnostic Errors in {3D} {CT} Disease Detection, Understanding and Reporting},
  author={Li, Yuheng and Chen, Yenho and Lai, Yuxiang and Zhong, Jike and Wildman, Vanessa and Yang, Xiaofeng},
  journal={arXiv preprint arXiv:2509.03800},
  year={2025}
}

@inproceedings{li2025more,
  title={More Performant and Scalable: Rethinking Contrastive Vision-Language Pre-training of Radiology in the {LLM} Era},
  author={Li, Yingtai and Lai, Haoran and Zhou, Xiaoqian and Ming, Shuai and Ma, Wenxin and Wei, Wei and Zhou, Shaohua Kevin},
  booktitle={International Conference on Medical Image Computing and Computer-Assisted Intervention},
  pages={348--357},
  year={2025},
  organization={Springer},
}

@inproceedings{yu2025location,
  title={Location-Guided Automated Lesion Captioning in Whole-Body {PET/CT} Images},
  author={Yu, Mingyang and Gao, Yaozong and Shu, Yiran and Chen, Yanbo and Liu, Jingyu and Jiang, Caiwen and Sun, Kaicong and Cui, Zhiming and Zhang, Weifang and Zhan, Yiqiang and others},
  booktitle={International Conference on Medical Image Computing and Computer-Assisted Intervention},
  pages={348--357},
  year={2025},
  organization={Springer},
}

@article{avci2025metadata,
  title={Metadata-aligned {3D} {MRI} representations for contrast understanding and quality control},
  author={Avci, Mehmet Yigit and Borges, Pedro and Fernandez, Virginia and Wright, Paul and Yigitsoy, Mehmet and Ourselin, Sebastien and Cardoso, Jorge},
  journal={arXiv preprint arXiv:2511.00681},
  year={2025}
}

@article{lai2025bridged,
  title={Bridged Semantic Alignment for Zero-Shot {3D} Medical Image Diagnosis},
  author={Lai, Haoran and Jiang, Zihang and Yao, Qingsong and Wang, Rongsheng and He, Zhiyang and Tao, Xiaodong and Lv, Weifu and Wei, Wei and Zhou, Shaohua Kevin},
  journal={IEEE Journal of Biomedical and Health Informatics},
  year={2025},
  publisher={IEEE},
  pages={1--14},
}

@article{mahdizadeh2025scale,
  title={{SCALE-VLP}: Soft-Weighted Contrastive Volumetric Vision-Language Pre-training with Spatial-Knowledge Semantics},
  author={Mahdizadeh, Ailar and Moghadam, Puria Azadi and He, Xiangteng and Mirabbasi, Shahriar and Nasiopoulos, Panos and Sigal, Leonid},
  journal={arXiv preprint arXiv:2511.02996},
  year={2025}
}

@article{ni2024mg,
  title={{MG-3D}: Multi-grained knowledge-enhanced {3D} medical vision-language pre-training},
  author={Ni, Xuefeng and Wu, Linshan and Zhuang, Jiaxin and Wang, Qiong and Wu, Mingxiang and Vardhanabhuti, Varut and Zhang, Lihai and Gao, Hanyu and Chen, Hao},
  journal={arXiv preprint arXiv:2412.05876},
  year={2024}
}

@article{wang2026sigvlp,
  title={{SigVLP}: Sigmoid Volume-Language Pre-Training for Self-Supervised {CT}-Volume Adaptive Representation Learning},
  author={Wang, Jiayi and Reynaud, Hadrien and Hamamci, Ibrahim Ethem and Er, Sezgin and Shit, Suprosanna and Menze, Bjoern and Kainz, Bernhard},
  journal={arXiv preprint arXiv:2602.21735},
  year={2026}
}

@article{ging2026learning,
  title={Learning to Read Where to Look: Disease-Aware Vision-Language Pretraining for {3D} {CT}},
  author={Ging, Simon and Arnold, Philipp and Walter, Sebastian and Alnahas, Hani and Bast, Hannah and Kotter, Elmar and Yang, Jiancheng and Bozorgtabar, Behzad and Brox, Thomas},
  journal={arXiv preprint arXiv:2603.02026},
  year={2026}
}

@inproceedings{radford2021learning,
  title={Learning transferable visual models from natural language supervision},
  author={Radford, Alec and Kim, Jong Wook and Hallacy, Chris and Ramesh, Aditya and Goh, Gabriel and Agarwal, Sandhini and Sastry, Girish and Askell, Amanda and Mishkin, Pamela and Clark, Jack and others},
  booktitle={International Conference on Machine Learning},
  pages={8748--8763},
  year={2021},
  organization={PMLR},
  volume={139},
}

@inproceedings{devlin2019bert,
  title={{BERT}: Pre-Training of Deep Bidirectional Transformers for Language Understanding},
  author={Devlin, Jacob and Chang, Ming-Wei and Lee, Kenton and Toutanova, Kristina},
  booktitle={Proceedings of the 2019 Conference of the North American Chapter of the Association for Computational Linguistics: Human Language Technologies, Volume 1 (Long and Short Papers)},
  pages={4171--4186},
  year={2019},
}

@article{anil2023palm,
  title={Palm 2 technical report},
  author={Anil, Rohan and Dai, Andrew M and Firat, Orhan and Johnson, Melvin and Lepikhin, Dmitry and Passos, Alexandre and Shakeri, Siamak and Taropa, Emanuel and Bailey, Paige and Chen, Zhifeng and others},
  journal={arXiv preprint arXiv:2305.10403},
  year={2023}
}

@article{li2025otter,
  title={{Otter}: A Multi-Modal Model With In-Context Instruction Tuning},
  author={Li, Bo and Zhang, Yuanhan and Chen, Liangyu and Wang, Jinghao and Pu, Fanyi and Cahyono, Joshua Adrian and Yang, Jingkang and Li, Chunyuan and Liu, Ziwei},
  journal={IEEE Transactions on Pattern Analysis and Machine Intelligence},
  year={2025},
  publisher={IEEE},
  volume={47},
  number={9},
  pages={7543--7557},
}

@article{touvron2023llama,
  title={{LLaMA}: Open and efficient foundation language models},
  author={Touvron, Hugo and Lavril, Thibaut and Izacard, Gautier and Martinet, Xavier and Lachaux, Marie-Anne and Lacroix, Timoth{\'e}e and Rozi{\`e}re, Baptiste and Goyal, Naman and Hambro, Eric and Azhar, Faisal and others},
  journal={arXiv preprint arXiv:2302.13971},
  year={2023}
}

@article{bai2025qwen3,
  title={{Qwen3-VL} Technical Report},
  author={Bai, Shuai and Cai, Yuxuan and Chen, Ruizhe and Chen, Keqin and Chen, Xionghui and Cheng, Zesen and Deng, Lianghao and Ding, Wei and Gao, Chang and Ge, Chunjiang and others},
  journal={arXiv preprint arXiv:2511.21631},
  year={2025}
}

@article{touvron2023llama2,
  title={{LLaMA} 2: Open foundation and fine-tuned chat models},
  author={Touvron, Hugo and Martin, Louis and Stone, Kevin and Albert, Peter and Almahairi, Amjad and Babaei, Yasmine and Bashlykov, Nikolay and Batra, Soumya and Bhargava, Prajjwal and Bhosale, Shruti and others},
  journal={arXiv preprint arXiv:2307.09288},
  year={2023}
}

@article{grattafiori2024llama,
  title={The {LLaMA} 3 herd of models},
  author={Grattafiori, Aaron and Dubey, Abhimanyu and Jauhri, Abhinav and Pandey, Abhinav and Kadian, Abhishek and Al-Dahle, Ahmad and Letman, Aiesha and Mathur, Akhil and Schelten, Alan and Vaughan, Alex and others},
  journal={arXiv preprint arXiv:2407.21783},
  year={2024}
}

@article{team2024gemma,
  title={{Gemma 2}: Improving open language models at a practical size},
  author={Team, Gemma and Riviere, Morgane and Pathak, Shreya and Sessa, Pier Giuseppe and Hardin, Cassidy and Bhupatiraju, Surya and Hussenot, L{\'e}onard and Mesnard, Thomas and Shahriari, Bobak and Ram{\'e}, Alexandre and others},
  journal={arXiv preprint arXiv:2408.00118},
  year={2024}
}

@article{awadalla2023openflamingo,
  title={{OpenFlamingo}: An open-source framework for training large autoregressive vision-language models},
  author={Awadalla, Anas and Gao, Irena and Gardner, Josh and Hessel, Jack and Hanafy, Yusuf and Zhu, Wanrong and Marathe, Kalyani and Bitton, Yonatan and Gadre, Samir and Sagawa, Shiori and others},
  journal={arXiv preprint arXiv:2308.01390},
  year={2023}
}

@article{wu2024pmc,
  title={{PMC-LLaMA}: toward building open-source language models for medicine},
  author={Wu, Chaoyi and Lin, Weixiong and Zhang, Xiaoman and Zhang, Ya and Xie, Weidi and Wang, Yanfeng},
  journal={Journal of the American Medical Informatics Association},
  volume={31},
  number={9},
  pages={1833--1843},
  year={2024},
  publisher={Oxford University Press},
}

@article{abdin2024phi,
  title={{Phi-4} technical report},
  author={Abdin, Marah and Aneja, Jyoti and Behl, Harkirat and Bubeck, S{\'e}bastien and Eldan, Ronen and Gunasekar, Suriya and Harrison, Michael and Hewett, Russell J and Javaheripi, Mojan and Kauffmann, Piero and others},
  journal={arXiv preprint arXiv:2412.08905},
  year={2024}
}

@article{zhang2024generalist,
  title={A generalist vision--language foundation model for diverse biomedical tasks},
  author={Zhang, Kai and Zhou, Rong and Adhikarla, Eashan and Yan, Zhiling and Liu, Yixin and Yu, Jun and Liu, Zhengliang and Chen, Xun and Davison, Brian D and Ren, Hui and others},
  journal={Nature Medicine},
  volume={30},
  number={11},
  pages={3129--3141},
  year={2024},
  publisher={Nature Publishing Group US New York},
}

@article{zhang2023pmc,
  title={{PMC-VQA}: Visual instruction tuning for medical visual question answering},
  author={Zhang, Xiaoman and Wu, Chaoyi and Zhao, Ziheng and Lin, Weixiong and Zhang, Ya and Wang, Yanfeng and Xie, Weidi},
  journal={arXiv preprint arXiv:2305.10415},
  year={2023}
}

@inproceedings{li2023llava,
  title={{LLaVA-Med}: Training a Large Language-and-Vision Assistant for Biomedicine in One Day},
  author={Li, Chunyuan and Wong, Cliff and Zhang, Sheng and Usuyama, Naoto and Liu, Haotian and Yang, Jianwei and Naumann, Tristan and Poon, Hoifung and Gao, Jianfeng},
  volume={36},
  pages={28541--28564},
  year={2023},
  booktitle={Advances in Neural Information Processing Systems},
}

@inproceedings{moor2023med,
  title={{Med-Flamingo}: a multimodal medical few-shot learner},
  author={Moor, Michael and Huang, Qian and Wu, Shirley and Yasunaga, Michihiro and Dalmia, Yash and Leskovec, Jure and Zakka, Cyril and Reis, Eduardo Pontes and Rajpurkar, Pranav},
  booktitle={Machine Learning for Health (ML4H)},
  pages={353--367},
  year={2023},
  organization={PMLR},
  volume={225},
}

@article{tu2024towards,
  title={Towards Generalist Biomedical {AI}},
  author={Tu, Tao and Azizi, Shekoofeh and Driess, Danny and Schaekermann, Mike and Amin, Mohamed and Chang, Pi-Chuan and Carroll, Andrew and Lau, Charles and Tanno, Ryutaro and Ktena, Ira and others},
  journal={NEJM AI},
  volume={1},
  number={3},
  pages={AIoa2300138},
  year={2024},
  publisher={Massachusetts Medical Society},
}

@article{liu2023qilin,
  title={{Qilin-Med-VL}: Towards {Chinese} large vision-language model for general healthcare},
  author={Liu, Junling and Wang, Ziming and Ye, Qichen and Chong, Dading and Zhou, Peilin and Hua, Yining},
  journal={arXiv preprint arXiv:2310.17956},
  year={2023}
}

@article{wang2023r2gengpt,
  title={{R2GenGPT}: Radiology Report Generation with frozen {LLMs}},
  author={Wang, Zhanyu and Liu, Lingqiao and Wang, Lei and Zhou, Luping},
  journal={Meta-Radiology},
  volume={1},
  number={3},
  pages={100033},
  year={2023},
  publisher={Elsevier},
}

@inproceedings{chen2024towards,
  title={Towards Injecting Medical Visual Knowledge into Multimodal {LLMs} at Scale},
  author={Chen, Junying and Gui, Chi and Ouyang, Ruyi and Gao, Anningzhe and Chen, Shunian and Chen, Guiming Hardy and Wang, Xidong and Cai, Zhenyang and Ji, Ke and Wan, Xiang and others},
  booktitle={Proceedings of the 2024 Conference on Empirical Methods in Natural Language Processing},
  pages={7346--7370},
  year={2024},
}

@inproceedings{lin2025healthgpt,
  title={{HealthGPT}: A Medical Large Vision-Language Model for Unifying Comprehension and Generation via Heterogeneous Knowledge Adaptation},
  author={Lin, Tianwei and Zhang, Wenqiao and Li, Sijing and Yuan, Yuqian and Yu, Binhe and Li, Haoyuan and He, Wanggui and Jiang, Hao and Li, Mengze and Xiaohui, Song and others},
  booktitle={International Conference on Machine Learning},
  pages={37975--37995},
  year={2025},
  organization={PMLR},
  volume={267},
}

@article{lai2026med,
  title={{Med-R1}: Reinforcement Learning for Generalizable Medical Reasoning in Vision-Language Models},
  author={Lai, Yuxiang and Zhong, Jike and Li, Ming and Zhao, Shitian and Li, Yuheng and Psounis, Konstantinos and Yang, Xiaofeng},
  journal={IEEE Transactions on Medical Imaging},
  year={2026},
  publisher={IEEE},
  volume={45},
  number={6},
  pages={2727--2737},
}

@article{dai2025qoq,
  title={{QoQ-Med}: Building multimodal clinical foundation models with domain-aware {GRPO} training},
  author={Dai, Wei and Chen, Peilin and Ekbote, Chanakya and Liang, Paul Pu},
  journal={arXiv preprint arXiv:2506.00711},
  year={2025}
}

@article{xu2025lingshu,
  title={Lingshu: A generalist foundation model for unified multimodal medical understanding and reasoning},
  author={Xu, Weiwen and Chan, Hou Pong and Li, Long and Aljunied, Mahani and Yuan, Ruifeng and Wang, Jianyu and Xiao, Chenghao and Chen, Guizhen and Liu, Chaoqun and Li, Zhaodonghui and others},
  journal={arXiv preprint arXiv:2506.07044},
  year={2025}
}

@article{sellergren2025medgemma,
  title={{MedGemma} technical report},
  author={Sellergren, Andrew and Kazemzadeh, Sahar and Jaroensri, Tiam and Kiraly, Atilla and Traverse, Madeleine and Kohlberger, Timo and Xu, Shawn and Jamil, Fayaz and Hughes, C{\'\i}an and Lau, Charles and others},
  journal={arXiv preprint arXiv:2507.05201},
  year={2025}
}

@article{ossowski2025octomed,
  title={{OctoMed}: Data Recipes for State-of-the-Art Multimodal Medical Reasoning},
  author={Ossowski, Timothy and Zhang, Sheng and Liu, Qianchu and Qin, Guanghui and Tan, Reuben and Naumann, Tristan and Hu, Junjie and Poon, Hoifung},
  journal={arXiv preprint arXiv:2511.23269},
  year={2025}
}

@article{wu2025towards,
  title={Towards generalist foundation model for radiology by leveraging web-scale {2D}\&{3D} medical data},
  author={Wu, Chaoyi and Zhang, Xiaoman and Zhang, Ya and Hui, Hui and Wang, Yanfeng and Xie, Weidi},
  journal={Nature Communications},
  volume={16},
  number={1},
  pages={7866},
  year={2025},
  publisher={Nature Publishing Group UK London},
}

@article{yang2024advancing,
  title={Advancing multimodal medical capabilities of Gemini},
  author={Yang, Lin and Xu, Shawn and Sellergren, Andrew and Kohlberger, Timo and Zhou, Yuchen and Ktena, Ira and Kiraly, Atilla and Ahmed, Faruk and Hormozdiari, Farhad and Jaroensri, Tiam and others},
  journal={arXiv preprint arXiv:2405.03162},
  year={2024}
}

@inproceedings{shi2025med,
  title={{Med-2E3}: A {2D}-Enhanced {3D} Medical Multimodal Large Language Model},
  author={Shi, Yiming and Zhu, Xun and Wang, Kaiwen and Hu, Ying and Guo, Chenyi and Li, Miao and Wu, Ji},
  booktitle={2025 IEEE International Conference on Bioinformatics and Biomedicine (BIBM)},
  pages={2754--2759},
  year={2025},
  organization={IEEE},
}

@article{jiang2025hulu,
  title={{Hulu-Med}: A transparent generalist model towards holistic medical vision-language understanding},
  author={Jiang, Songtao and Wang, Yuan and Song, Sibo and Hu, Tianxiang and Zhou, Chenyi and Pu, Bin and Zhang, Yan and Yang, Zhibo and Feng, Yang and Zhou, Joey Tianyi and others},
  journal={arXiv preprint arXiv:2510.08668},
  year={2025}
}

@article{shu2025fleming,
  title={{Fleming-VL}: Towards Universal Medical Visual Reasoning with Multimodal {LLMs}},
  author={Shu, Yan and Liu, Chi and Chen, Robin and Li, Derek and Dai, Bryan},
  journal={arXiv preprint arXiv:2511.00916},
  year={2025}
}

@inproceedings{shi2025medm,
  title={{MedM-VL}: What Makes a Good Medical {LVLM}?},
  author={Shi, Yiming and Yang, Shaoshuai and Zhu, Xun and Wang, Haoyu and Fu, Xiangling and Li, Miao and Wu, Ji},
  booktitle={International Workshop on Agentic AI for Medicine},
  pages={290--299},
  year={2025},
  organization={Springer},
}

@article{lei2026versatile,
  title={Versatile Vision-Language Model for {3D} Computed Tomography},
  author={Lei, Jiayu and Fan, Ziqing and Zhang, Yanyong and Xie, Weidi and Zhang, Ya and Wang, Yanfeng},
  volume={40},
  pages={5945--5954},
  year={2026},
  number={8},
  journal={Proceedings of the AAAI Conference on Artificial Intelligence},
}

@article{liu2026medpruner,
  title={{MedPruner}: Training-Free Hierarchical Token Pruning for Efficient {3D} Medical Image Understanding in Vision-Language Models},
  author={Liu, Shengyuan and Ye, Zanting and Lin, Yunrui and Hu, Chen and Geng, Wanting and Han, Xu and Ibragimov, Bulat and Zheng, Yefeng and Yuan, Yixuan},
  journal={arXiv preprint arXiv:2603.11625},
  year={2026}
}

@article{fang2026photon,
  title={Photon: Speedup volume understanding with efficient multimodal large language models},
  author={Fang, Chengyu and Guo, Heng and Jiang, Zheng and He, Chunming and Li, Xiu and Xu, Minfeng},
  journal={arXiv preprint arXiv:2603.25155},
  year={2026}
}

@article{lin2026omnict,
  title={{OmniCT}: Towards a Unified Slice-Volume {LVLM} for Comprehensive {CT} Analysis},
  author={Lin, Tianwei and Qiu, Zhongwei and Zhang, Wenqiao and Liu, Jiang and Xie, Yihan and Gao, Mingjian and Fan, Zhenxuan and Li, Zhaocheng and Li, Sijing and Xie, Zhongle and others},
  journal={arXiv preprint arXiv:2602.16110},
  year={2026}
}

@article{sellergren2026medgemma,
  title={{MedGemma} 1.5 Technical Report},
  author={Sellergren, Andrew and Gao, Chufan and Mahvar, Fereshteh and Kohlberger, Timo and Jamil, Fayaz and Traverse, Madeleine and Tono, Alberto and Sadjad, Bashir and Yang, Lin and Lau, Charles and others},
  journal={arXiv preprint arXiv:2604.05081},
  year={2026}
}

@article{yu2026adapting,
  title={Adapting {2D} Multi-Modal Large Language Model for {3D} {CT} Image Analysis},
  author={Yu, Yang and Xu, Dunyuan and Li, Yaoqian and Li, Xiaomeng and Li, Jinpeng and Heng, Pheng-Ann},
  journal={arXiv preprint arXiv:2604.10233},
  year={2026}
}

@inproceedings{chen2025dia,
  title={Dia-{LLaMA}: Towards Large Language Model-Driven {CT} Report Generation},
  author={Chen, Zhixuan and Luo, Luyang and Bie, Yequan and Chen, Hao},
  booktitle={International Conference on Medical Image Computing and Computer-Assisted Intervention},
  pages={141--151},
  year={2025},
  organization={Springer},
}

@article{bai2024m3d,
  title={{M3D}: Advancing {3D} medical image analysis with multi-modal large language models},
  author={Bai, Fan and Du, Yuxin and Huang, Tiejun and Meng, Max Q-H and Zhao, Bo},
  journal={arXiv preprint arXiv:2404.00578},
  year={2024}
}

@article{blankemeier2026merlin,
  title={Merlin: a computed tomography vision--language foundation model and dataset},
  author={Blankemeier, Louis and Kumar, Ashwin and Cohen, Joseph Paul and Liu, Jiaming and Liu, Longchao and Van Veen, Dave and Gardezi, Syed Jamal Safdar and Yu, Hongkun and Paschali, Magdalini and Chen, Zhihong and others},
  journal={Nature},
  pages={1--11},
  year={2026},
  publisher={Nature Publishing Group UK London},
  volume={652},
  number={8112},
}

@article{li2024towards,
  title={Towards a holistic framework for multimodal large language models in three-dimensional brain {CT} report generation},
  author={Li, Cheng-Yi and Chang, Kao-Jung and Yang, Cheng-Fu and Wu, Hsin-Yu and Chen, Wenting and Bansal, Hritik and Chen, Ling and Yang, Yi-Ping and Chen, Yu-Chun and Chen, Shih-Pin and others},
  journal={arXiv preprint arXiv:2407.02235},
  year={2024}
}

@article{chen20243d,
  title={{3D}-{CT}-{GPT}: Generating {3D} radiology reports through integration of large vision-language models},
  author={Chen, Hao and Zhao, Wei and Li, Yingli and Zhong, Tianyang and Wang, Yisong and Shang, Youlan and Guo, Lei and Han, Junwei and Liu, Tianming and Liu, Jun and others},
  journal={arXiv preprint arXiv:2409.19330},
  year={2024}
}

@article{lai2024e3d,
  title={{E3D-GPT}: enhanced {3D} visual foundation for medical vision-language model},
  author={Lai, Haoran and Jiang, Zihang and Yao, Qingsong and Wang, Rongsheng and He, Zhiyang and Tao, Xiaodong and Wei, Wei and Lv, Weifu and Zhou, S Kevin},
  journal={arXiv preprint arXiv:2410.14200},
  year={2024}
}

@article{chen2025large,
  title={Large Language Model With Region-Guided Referring and Grounding for {CT} Report Generation},
  author={Chen, Zhixuan and Bie, Yequan and Jin, Haibo and Chen, Hao},
  journal={IEEE Transactions on Medical Imaging},
  year={2025},
  publisher={IEEE},
  volume={44},
  number={8},
  pages={3139--3150},
}

@article{lee2024read,
  title={Read like a radiologist: efficient vision-language model for {3D} medical imaging interpretation},
  author={Lee, Changsun and Park, Sangjoon and Shin, Cheong-Il and Choi, Woo Hee and Park, Hyun Jeong and Lee, Jeong Eun and Ye, Jong Chul},
  journal={arXiv preprint arXiv:2412.13558},
  year={2024}
}

@article{zhang2025mepnet,
  title={{MEPNet}: Medical Entity-Balanced Prompting Network for Brain {CT} Report Generation},
  author={Zhang, Xiaodan and Shi, Yanzhao and Ji, Junzhong and Zheng, Chengxin and Qu, Liangqiong},
  volume={39},
  pages={25940--25948},
  year={2025},
  number={24},
  journal={Proceedings of the AAAI Conference on Artificial Intelligence},
}

@article{xin2025med3dvlm,
  title={{Med3DVLM}: An Efficient Vision-Language Model for {3D} Medical Image Analysis},
  author={Xin, Yu and Ates, Gorkem Can and Gong, Kuang and Shao, Wei},
  journal={IEEE Journal of Biomedical and Health Informatics},
  year={2025},
  publisher={IEEE},
  volume={30},
  number={3},
  pages={2524--2536},
}

@article{shi2025hsenet,
  title={{HSENet}: Hybrid Spatial Encoding Network for {3D} Medical Vision-Language Understanding},
  author={Shi, Yanzhao and Zhang, Xiaodan and Ji, Junzhong and Jiang, Haoning and Zheng, Chengxin and Wang, Yinong and Qu, Liangqiong},
  journal={arXiv preprint arXiv:2506.09634},
  year={2025}
}

@article{kyung2025medregion,
  title={{MedRegion-CT}: region-focused multimodal {LLM} for comprehensive {3D} {CT} report generation},
  author={Kyung, Sunggu and Seo, Jinyoung and Lim, Hyunseok and Kim, Dongyeong and Park, Hyungbin and Sung, Jimin and Kim, Jihyun and Jo, Wooyoung and Nam, Yoojin and Kim, Namkug},
  journal={arXiv preprint arXiv:2506.23102},
  year={2025}
}

@article{vepa2025multimodal,
  title={A Multimodal {LLM} Approach for Visual Question Answering on Multiparametric {3D} Brain {MRI}},
  author={Vepa, Arvind Murari and Yu, Yannan and Gan, Jingru and Cuturrufo, Anthony and Li, Weikai and Wang, Wei and Scalzo, Fabien and Sun, Yizhou},
  journal={arXiv preprint arXiv:2509.25889},
  year={2025}
}

@article{maqbool2025petar,
  title={{PETAR}: Localized Findings Generation with Mask-Aware Vision-Language Modeling for {PET} Automated Reporting},
  author={Maqbool, Danyal and Lee, Changhee and Huemann, Zachary and Church, Samuel D and Larson, Matthew E and Perlman, Scott B and Romero, Tomas A and Warner, Joshua D and Lubner, Meghan and Tie, Xin and others},
  journal={arXiv preprint arXiv:2510.27680},
  year={2025}
}

@article{hamamci2025better,
  title={Better Tokens for Better {3D}: Advancing Vision-Language Modeling in {3D} Medical Imaging},
  author={Hamamci, Ibrahim Ethem and Er, Sezgin and Shit, Suprosanna and Reynaud, Hadrien and Yang, Dong and Guo, Pengfei and Edgar, Marc and Xu, Daguang and Kainz, Bernhard and Menze, Bjoern},
  journal={arXiv preprint arXiv:2510.20639},
  year={2025}
}

@article{jiao2025vision,
  title={Vision-Language Models for Automated {3D} {PET/CT} Report Generation},
  author={Jiao, Wenpei and Shang, Kun and Li, Hui and Yan, Ke and Zhang, Jiajin and Yang, Guangjie and Guo, Lijuan and Wan, Yan and Yang, Xing and Jin, Dakai and others},
  journal={arXiv preprint arXiv:2511.20145},
  year={2025}
}

@article{barone2026brain3d,
  title={{Brain3D}: Brain Report Automation via Inflated Vision Transformers in {3D}},
  author={Barone, Mariano and Di Serio, Francesco and Riccio, Giuseppe and Romano, Antonio and Postiglione, Marco and Ferraro, Antonino and Moscato, Vincenzo},
  journal={arXiv preprint arXiv:2602.22098},
  year={2026}
}

@article{lai2026med3d,
  title={{Med3D-R1}: Incentivizing Clinical Reasoning in {3D} Medical Vision-Language Models for Abnormality Diagnosis},
  author={Lai, Haoran and Jiang, Zihang and Zhang, Kun and Yao, Qingsong and Wang, Rongsheng and He, Zhiyang and Tao, Xiaodong and Wei, Wei and Zhou, Shaohua Kevin},
  journal={arXiv preprint arXiv:2602.01200},
  year={2026}
}

@article{shi2026u,
  title={U-{VLM}: Hierarchical Vision Language Modeling for Report Generation},
  author={Shi, Pengcheng and Zhang, Minghui and Song, Kehan and Liu, Jiaqi and Gu, Yun and Zhang, Xinglin},
  journal={arXiv preprint arXiv:2603.00479},
  year={2026}
}

@article{xing2026medvl,
  title={{MedVL-SAM2}: A unified {3D} medical vision-language model for multimodal reasoning and prompt-driven segmentation},
  author={Xing, Yang and Wu, Jiong and Ozdemir, Savas and Zhang, Ying and Yang, Yang and Shao, Wei and Gong, Kuang},
  journal={arXiv preprint arXiv:2601.09879},
  year={2026}
}

@article{zeng2026enhancing,
  title={Enhancing {3D} medical multi-modal large language models with integrated human body priors for computed tomography},
  author={Zeng, Leilei and Liu, Jie and Chen, Wenting and Lyu, Chenyang and Li, Wenxi and Liu, Shaonan and Zhou, Xiande and Shen, Linlin},
  journal={Pattern Recognition},
  pages={113540},
  year={2026},
  publisher={Elsevier},
  volume={179},
}

@article{wang2026enhancing,
  title={Enhancing Fine-Grained Spatial Grounding in {3D} {CT} Report Generation via Discriminative Guidance},
  author={Wang, Chenyu and Dai, Weicheng and Liu, Han and Li, Wenchao and Batmanghelich, Kayhan},
  journal={arXiv preprint arXiv:2604.10437},
  year={2026}
}

@article{mao2025ct,
  title={{CT-Agent}: a multimodal-{LLM} agent for {3D} {CT} radiology question answering},
  author={Mao, Yuren and Xu, Wenyi and Qin, Yuyang and Gao, Yunjun},
  journal={arXiv preprint arXiv:2505.16229},
  year={2025}
}

@article{wang20263dmedagent,
  title={{3DMedAgent}: Unified Perception-to-Understanding for {3D} Medical Analysis},
  author={Wang, Ziyue and Cai, Linghan and Low, Chang Han and Liu, Haofeng and Wu, Junde and Wang, Jingyu and Wang, Rui and Song, Lei and Bian, Jiang and Fu, Jingjing and others},
  journal={arXiv preprint arXiv:2602.18064},
  year={2026}
}

@article{roschewitz2026radagent,
  title={{RadAgent}: A tool-using {AI} agent for stepwise interpretation of chest computed tomography},
  author={Roschewitz, M{\'e}lanie and Styppa, Kenneth and Tao, Yitian and Sohn, Jiwoong and Delbrouck, Jean-Benoit and Gundersen, Benjamin and Deperrois, Nicolas and Bluethgen, Christian and Vogt, Julia and Menze, Bjoern and others},
  journal={arXiv preprint arXiv:2604.15231},
  year={2026}
}

@article{gu2026ct,
  title={{CT-Flow}: Orchestrating {CT} Interpretation Workflow with Model Context Protocol Servers},
  author={Gu, Yannian and Zhang, Xizhuo and Mu, Linjie and Yu, Yongrui and Huang, Zhongzhen and Zhang, Shaoting and Zhang, Xiaofan},
  journal={arXiv preprint arXiv:2603.00123},
  year={2026}
}

@article{yu2025radiologist,
  title={{Radiologist Copilot}: An Agentic Assistant with Orchestrated Tools for Radiology Reporting with Quality Control},
  author={Yu, Yongrui and Huang, Zhongzhen and Mu, Linjie and Zhang, Shaoting and Zhang, Xiaofan},
  journal={arXiv preprint arXiv:2512.02814},
  year={2025}
}

@article{lin2026march,
  title={{MARCH}: Multi-Agent Radiology Clinical Hierarchy for {CT} Report Generation},
  author={Lin, Yi and Ding, Yihao and Wu, Yonghui and Peng, Yifan},
  journal={arXiv preprint arXiv:2604.16175},
  year={2026}
}

@article{xiao2026spineagent,
  title={A multi-agent system for spine {MRI} report generation from multi-sequence imaging},
  author={Xiao, Zhiping and Yang, Junwei and Sun, Gongbo and Zhang, Han and Xu, Hanwen and Yao, Yi and Miller, Zachary D. and King, William E. and Kanani, Mohammed M. and Andre, Jalal B. and Chu, Sammy and Zhang, Ming and Kinahan, Paul E. and Cross, Nathan M. and Wang, Sheng},
  journal={arXiv preprint arXiv:2606.08897},
  year={2026}
}

@article{hosseini2026medtoolica,
  title={{MedToolica}: Finetuning-Free Agentic Compositional Tool Learning for {3D} {CT} Reasoning},
  author={Hosseini, Abdullah and Serag, Ahmed},
  journal={Machine Learning and Knowledge Extraction},
  volume={8},
  number={6},
  pages={162},
  year={2026},
  }

@misc{erdur2026agenticlargelanguagemodels,
      title={Agentic Large Language Models for Training-Free Neuro-Radiological Image Analysis}, 
      author={Ayhan Can Erdur and Daniel Scholz and Jiazhen Pan and Benedikt Wiestler and Daniel Rueckert and Jan C. Peeken},
      year={2026},
      eprint={2604.16729},
      archivePrefix={arXiv},
      primaryClass={cs.CV},
      url={https://arxiv.org/abs/2604.16729}, 
}

@inproceedings{kim2024mdagents,
  title={{MDAgents}: An Adaptive Collaboration of {LLMs} for Medical Decision-Making},
  author={Kim, Yubin and Park, Chanwoo and Jeong, Hyewon and Chan, Yik S and Xu, Xuhai and McDuff, Daniel and Lee, Hyeonhoon and Ghassemi, Marzyeh and Breazeal, Cynthia and Park, Hae W},
  volume={37},
  pages={79410--79452},
  year={2024},
  booktitle={Advances in Neural Information Processing Systems},
}

@inproceedings{li2024mmedagent,
  title={{MMedAgent}: Learning to Use Medical Tools with Multi-modal Agent},
  author={Li, Binxu and Yan, Tiankai and Pan, Yuanting and Luo, Jie and Ji, Ruiyang and Ding, Jiayuan and Xu, Zhe and Liu, Shilong and Dong, Haoyu and Lin, Zihao and others},
  booktitle={Findings of the Association for Computational Linguistics: EMNLP 2024},
  pages={8745--8760},
  year={2024},
}

@article{wang2025medagent,
  title={{MedAgent-Pro}: Towards evidence-based multi-modal medical diagnosis via reasoning agentic workflow},
  author={Wang, Ziyue and Wu, Junde and Cai, Linghan and Low, Chang Han and Yang, Xihong and Li, Qiaxuan and Jin, Yueming},
  journal={arXiv preprint arXiv:2503.18968},
  year={2025}
}

@article{qu2026baai,
  title={{BAAI Cardiac Agent}: An intelligent multimodal agent for automated reasoning and diagnosis of cardiovascular diseases from cardiac magnetic resonance imaging},
  author={Qu, Taiping and Zhang, Hongkai and Zhang, Lantian and Zhao, Can and Zhang, Nan and Wang, Hui and Zhou, Zhen and Zou, Mingye and Bo, Kairui and Zhao, Pengfei and others},
  journal={arXiv preprint arXiv:2604.04078},
  year={2026}
}

@article{sajua2025agentmri,
  title={{AgentMRI}: A Vison Language Model-Powered {AI} System for Self-regulating {MRI} Reconstruction with Multiple Degradations},
  author={Sajua, Gulfam Ahmed and Akhib, Marjan and Chang, Yuchou},
  journal={Journal of Imaging Informatics in Medicine},
  pages={1422--1440},
  year={2025},
  publisher={Springer},
  volume={39},
  number={2},
}

@article{zhong2025vision,
  title={Vision-language model for report generation and outcome prediction in {CT} pulmonary angiogram},
  author={Zhong, Zhusi and Wang, Yuli and Wu, Jing and Hsu, Wen-Chi and Somasundaram, Vin and Bi, Lulu and Kulkarni, Shreyas and Ma, Zhuoqi and Collins, Scott and Baird, Grayson and others},
  journal={npj Digital Medicine},
  volume={8},
  number={1},
  pages={432},
  year={2025},
  publisher={Nature Publishing Group UK London},
}

@article{hoopes2024voxelprompt,
  title={{VoxelPrompt}: A Vision Agent for End-to-End Medical Image Analysis},
  author={Hoopes, Andrew and Dey, Neel and Butoi, Victor Ion and Guttag, John V and Dalca, Adrian V},
  journal={arXiv preprint arXiv:2410.08397},
  year={2024}
}

@article{liu2026medsam,
  title={{MedSAM-Agent}: Empowering Interactive Medical Image Segmentation with Multi-turn Agentic Reinforcement Learning},
  author={Liu, Shengyuan and Bao, Liuxin and Yang, Qi and Geng, Wanting and Zheng, Boyun and Li, Chenxin and Chen, Wenting and Peng, Houwen and Yuan, Yixuan},
  journal={arXiv preprint arXiv:2602.03320},
  year={2026}
}

@article{huang2026medsegagent,
  title={{MedSegAgent}: A Universal and Scalable Multi-Agent System for Instructive Medical Image Segmentation},
  author={Huang, Ziyan and Wang, Haoyu and Ye, Jin and Ji, Yuanfeng and Hu, Xiaowei and Liu, Lihao and Yang, Zhikai and Li, Wei and Hu, Ming and Su, Yanzhou and others},
  journal={IEEE Journal of Biomedical and Health Informatics},
  year={2026},
  publisher={IEEE},
  pages={1--12},
}

@article{shen2026medopenclaw,
  title={{MedOpenClaw}: Auditable Medical Imaging Agents Reasoning over Uncurated Full Studies},
  author={Shen, Weixiang and Hu, Yanzhu and Liu, Che and Wu, Junde and Zhu, Jiayuan and Shen, Chengzhi and Xu, Min and Jin, Yueming and Wiestler, Benedikt and Rueckert, Daniel and others},
  journal={arXiv preprint arXiv:2603.24649},
  year={2026}
}

@article{tzanis2026dosimetron,
  title={{DosimeTron}: Automating Personalized {Monte Carlo} Radiation Dosimetry in {PET/CT} with Agentic {AI}},
  author={Tzanis, Eleftherios and Klontzas, Michail E and Tzortzakakis, Antonios},
  journal={arXiv preprint arXiv:2604.06280},
  year={2026}
}

@article{zhong2026neuroagent,
  title={{NeuroAgent}: {LLM} Agents for Multimodal Neuroimaging Analysis and Research},
  author={Zhong, Lujia and Xia, Yihao and Zhang, Jianwei and Huang, Shuo and Yue, Jiaxin and Xia, Mingyang and Shi, Yonggang},
  journal={arXiv preprint arXiv:2605.06584},
  year={2026}
}

@article{han2026nexus,
  title={Towards a Virtual Neuroscientist: Autonomous Neuroimaging Analysis via Multi-Agent Collaboration},
  author={Han, Keqi and Li, Xiang and Zhao, Songlin and Su, Yao and Yuan, Yixuan and He, Lifang and Yang, Carl},
  journal={arXiv preprint arXiv:2605.09366},
  year={2026}
}

@article{wang2025feasibility,
  title={A feasibility study of automating radiotherapy planning with large language model agents},
  author={Wang, Qingxin and Wang, Zhongqiu and Li, Minghua and Ni, Xinye and Tan, Rong and Zhang, Wenwen and Wubulaishan, Maitudi and Wang, Wei and Yuan, Zhiyong and Zhang, Zhen and others},
  journal={Physics in Medicine \& Biology},
  volume={70},
  number={7},
  pages={075007},
  year={2025},
  publisher={IOP Publishing},
}

@article{saha2026picking,
  title={Picking the Right Specialist: Attentive Neural Process-based Selection of Task-Specialized Models as Tools for Agentic Healthcare Systems},
  author={Saha, Pramit and Strong, Joshua and Alsharid, Mohammad and Mishra, Divyanshu and Noble, J Alison},
  journal={arXiv preprint arXiv:2602.14901},
  year={2026}
}

@article{fallahpour2025medrax,
  title={{MedRAX}: Medical reasoning agent for chest x-ray},
  author={Fallahpour, Adibvafa and Ma, Jun and Munim, Alif and Lyu, Hongwei and Wang, Bo},
  journal={arXiv preprint arXiv:2502.02673},
  year={2025}
}

@article{huai2026tool,
  title={Which Tool Response Should I Trust? Tool-Expertise-Aware Chest X-ray Agent with Multimodal Agentic Learning},
  author={Huai, Zheang and Yang, Honglong and Li, Xiaomeng},
  journal={arXiv preprint arXiv:2602.21517},
  year={2026}
}

@article{lu2026medvistagym,
  title={{MEDVISTAGYM}: A Scalable Training Environment for Thinking with Medical Images via Tool-Integrated Reinforcement Learning},
  author={Lu, Meng and Lu, Yuxing and Zhuang, Yuchen and Mullins, Megan and Xie, Yang and Xiao, Guanghua and Fleming, Charles and Shi, Wenqi and Wang, Xuan},
  journal={arXiv preprint arXiv:2601.07107},
  year={2026}
}

@article{chen2026theraagent,
  title={{TheraAgent}: Multi-Agent Framework with Self-Evolving Memory and Evidence-Calibrated Reasoning for {PET} Theranostics},
  author={Chen, Zhihao and Wang, Jiahui and Chen, Yizhou and Ji, Xiaozhong and Hu, Xiaobin and Hong, Jimin and Bosbach, Wolfram Andreas and Rominger, Axel and Afshar-Oromieh, Ali and Shan, Hongming and others},
  journal={arXiv preprint arXiv:2603.13676},
  year={2026}
}

@article{jabal2026agentic,
  title={Agentic Automation of {BT-RADS} Scoring: End-to-End Multi-Agent System for Standardized Brain Tumor Follow-up Assessment},
  author={Jabal, Mohamed Sobhi and Zhang, Jikai and LaBella, Dominic and Houk, Jessica L and Zhang, Dylan and Rudie, Jeffrey D and Magudia, Kirti and Mazurowski, Maciej A and Calabrese, Evan},
  journal={arXiv preprint arXiv:2603.21494},
  year={2026}
}

@article{yang2026lungnoduleagent,
  title={{LungNoduleAgent}: A Collaborative Multi-Agent System for Precision Diagnosis of Lung Nodules},
  author={Yang, Cheng and Jin, Hui and Yu, Xinlei and Wang, Zhipeng and Liu, Yaoqun and Fan, Fenglei and Lei, Dajiang and Jia, Gangyong and Wang, Changmiao and Ge, Ruiquan},
  volume={40},
  pages={29793--29801},
  year={2026},
  number={35},
  journal={Proceedings of the AAAI Conference on Artificial Intelligence},
}

@article{aiersilan2026neuro,
  title={{Neuro-Oracle}: A Trajectory-Aware Agentic {RAG} Framework for Interpretable Epilepsy Surgical Prognosis},
  author={Aiersilan, Aizierjiang and Koubeissi, Mohamad},
  journal={arXiv preprint arXiv:2604.14216},
  year={2026}
}

@article{vahistha2026agent,
  title={{Agent-MIRA}: {AI}-orchestrated medical imaging agent for {PET} image retrieval and assistance},
  author={Vahistha, Rajat and Brosda, Sandra and Aoude, Lauren G and Ng, Jessica and Kundu, Parveen and Barbour, Andrew P and Vegh, Viktor},
  journal={Computerized Medical Imaging and Graphics},
  volume={125},
  pages={102725},
  year={2026},
  publisher={Elsevier},
}

@article{nusrat2025autonomous,
  title={Autonomous radiotherapy treatment planning using {DOLA}: A privacy-preserving, {LLM}-based optimization agent},
  author={Nusrat, Humza and Luo, Bing and Hall, Ryan and Kim, Joshua and Bagher-Ebadian, Hassan and Doemer, Anthony and Movsas, Benjamin and Thind, Kundan},
  journal={arXiv preprint arXiv:2503.17553},
  year={2025}
}

@article{zaiss2026agentic,
  title={Agentic {MR} sequence development: leveraging {LLMs} with {MR} skills for automatic physics-informed sequence development},
  author={Zaiss, Moritz and Aly, Amr and Endres, Jonathan and Dornstetter, Tobias and Weinm{\"u}ller, Simon and Maier, Andreas},
  journal={arXiv preprint arXiv:2604.13282},
  year={2026}
}

@article{pambudi2025bridging,
  title={Bridging Clinical Narratives and {ACR} Appropriateness Guidelines: A Multi-Agent {RAG} System for Medical Imaging Decisions},
  author={Pambudi, Satrio and Menolascina, Filippo},
  journal={arXiv preprint arXiv:2510.04969},
  year={2025}
}

@article{tziakouri2025reinforcement,
  title={Reinforcement Learning for Clinical Reasoning: Aligning {LLMs} with {ACR} Imaging Appropriateness Criteria},
  author={Tziakouri, Anni and Menolascina, Filippo},
  journal={arXiv preprint arXiv:2510.05194},
  year={2025}
}

@inproceedings{kang2025scan,
  title={{Scan-Do Attitude}: Towards Autonomous {CT} Protocol Management Using a Large Language Model Agent},
  author={Kang, Xingjian and Vorberg, Linda and Maier, Andreas and Katzmann, Alexander and Taubmann, Oliver},
  booktitle={International Workshop on Agentic AI for Medicine},
  pages={46--54},
  year={2025},
  organization={Springer},
}

@article{zheng2024well,
  title={How well can modern {LLMs} act as agent cores in radiology environments?},
  author={Zheng, Qiaoyu and Wu, Chaoyi and Qiu, Pengcheng and Dai, Lisong and Zhang, Ya and Wang, Yanfeng and Xie, Weidi},
  journal={arXiv preprint arXiv:2412.09529},
  year={2024}
}

@incollection{Andrearczyk2023Hecktor,
  title={Overview of the {HECKTOR} Challenge at {MICCAI} 2022: Automatic Head and Neck Tumor Segmentation and Outcome Prediction in {PET/CT}},
  author={Andrearczyk, Vincent and Oreiller, Valentin and Abobakr, Moamen and Akhavanallaf, Azadeh and Balermpas, Panagiotis and Boughdad, Sarah and Capriotti, Leo and Castelli, Joel and Cheze Le Rest, Catherine and Decazes, Pierre and others},
  booktitle={{3D} Head and Neck Tumor Segmentation in {PET/CT} Challenge},
  pages={1--30},
  year={2022},
  publisher={Springer},
}

@article{Armato2011LidcIdri,
  title={The {Lung Image Database Consortium} ({LIDC}) and {Image Database Resource Initiative} ({IDRI}): A Completed Reference Database of Lung Nodules on {CT} Scans},
  author={Armato III, Samuel G and McLennan, Geoffrey and Bidaut, Luc and McNitt-Gray, Michael F and Meyer, Charles R and Reeves, Anthony P and Zhao, Binsheng and Aberle, Denise R and Henschke, Claudia I and Hoffman, Eric A and others},
  journal={Medical Physics},
  volume={38},
  number={2},
  pages={915--931},
  year={2011},
  publisher={Wiley Online Library},
}

@article{Bercea2025Nova,
  title={Nova: A benchmark for anomaly localization and clinical reasoning in brain {MRI}},
  author={Bercea, Cosmin I and Li, Jun and Raffler, Philipp and Riedel, Evamaria O and Schmitzer, Lena and Kurz, Angela and Bitzer, Felix and Ro{\ss}m{\"u}ller, Paula and Canisius, Julian and Beyrle, Mirjam L and others},
  journal={arXiv preprint arXiv:2505.14064},
  year={2025}
}

@article{Chen2025DeepTumorVQA,
  title={Are vision language models ready for clinical diagnosis? a {3D} medical benchmark for tumor-centric visual question answering},
  author={Chen, Yixiong and Xiao, Wenjie and Bassi, Pedro RAS and Zhou, Xinze and Er, Sezgin and Hamamci, Ibrahim Ethem and Zhou, Zongwei and Yuille, Alan},
  journal={arXiv preprint arXiv:2505.18915},
  year={2025}
}

@inproceedings{Gai2025ThreeDRad,
  title={{3D-RAD}: A Comprehensive {3D} Radiology Med-{VQA} Dataset with Multi-Temporal Analysis and Diverse Diagnostic Tasks},
  author={Gai, Xiaotang and Liu, Jiaxiang and Li, Yichen and Meng, Zijie and Wu, Jian and Liu, Zuozhu},
  volume={38},
  year={2025},
  booktitle={Advances in Neural Information Processing Systems},
}

@inproceedings{Huang2023Inspect,
  title={{INSPECT}: A Multimodal Dataset for Patient Outcome Prediction of Pulmonary Embolisms},
  author={Huang, Shih-Cheng and Huo, Zepeng and Steinberg, Ethan and Chiang, Chia-Chun and Langlotz, Curtis and Lungren, Matthew and Yeung, Serena and Shah, Nigam and Fries, Jason},
  volume={36},
  pages={17742--17772},
  year={2023},
  booktitle={Advances in Neural Information Processing Systems},
}

@inproceedings{Ji2022Amos,
  title={{AMOS}: A Large-Scale Abdominal Multi-Organ Benchmark for Versatile Medical Image Segmentation},
  author={Ji, Yuanfeng and Bai, Haotian and Ge, Chongjian and Yang, Jie and Zhu, Ye and Zhang, Ruimao and Li, Zhen and Zhanng, Lingyan and Ma, Wanling and Wan, Xiang and others},
  volume={35},
  pages={36722--36732},
  year={2022},
  booktitle={Advances in Neural Information Processing Systems},
}

@article{Jiang2026LoV3D,
  title={{LoV3D}: Grounding Cognitive Prognosis Reasoning in Longitudinal {3D} Brain {MRI} via Regional Volume Assessments},
  author={Jiang, Zhaoyang and Fu, Zhizhong and McAllister, David and Kim, Yunsoo and Wu, Honghan},
  journal={arXiv preprint arXiv:2603.12071},
  year={2026}
}

@article{Khademi2025AutoRadLung,
  title={{AutoRad-Lung}: A radiomic-guided prompting autoregressive vision-language model for lung nodule malignancy prediction},
  author={Khademi, Sadaf and Shabanpour, Mehran and Taleei, Reza and Oikonomou, Anastasia and Mohammadi, Arash},
  journal={arXiv preprint arXiv:2503.20662},
  year={2025}
}

@article{LaMontagne2019Oasis3,
  title={{OASIS-3}: longitudinal neuroimaging, clinical, and cognitive dataset for normal aging and {Alzheimer} disease},
  author={LaMontagne, Pamela J and Benzinger, Tammie LS and Morris, John C and Keefe, Sarah and Hornbeck, Russ and Xiong, Chengjie and Grant, Elizabeth and Hassenstab, Jason and Moulder, Krista and Vlassenko, Andrei G and others},
  journal={medRxiv},
  pages={2019--12},
  year={2019},
  publisher={Cold Spring Harbor Laboratory Press}
}

@article{Li2025AbdomenAtlas,
  title={{AbdomenAtlas}: A large-scale, detailed-annotated, \& multi-center dataset for efficient transfer learning and open algorithmic benchmarking},
  author={Li, Wenxuan and Qu, Chongyu and Chen, Xiaoxi and Bassi, Pedro RAS and Shi, Yijia and Lai, Yuxiang and Yu, Qian and Xue, Huimin and Chen, Yixiong and Lin, Xiaorui and others},
  journal={Medical Image Analysis},
  volume={97},
  pages={103285},
  year={2024},
  publisher={Elsevier},
}

@inproceedings{Liu2025Argus,
  title={Argus: Benchmarking and Enhancing Vision-Language Models for {3D} Radiology Report Generation},
  author={Liu, Che and Wan, Zhongwei and Wang, Yuqi and Shen, Hui and Wang, Haozhe and Zheng, Kangyu and Zhang, Mi and Arcucci, Rossella},
  booktitle={Findings of the Association for Computational Linguistics: ACL 2025},
  pages={16448--16460},
  year={2025},
}

@article{Menze2015Brats,
  title={The Multimodal Brain Tumor Image Segmentation Benchmark ({BRATS})},
  author={Menze, Bjoern H and Jakab, Andras and Bauer, Stefan and Kalpathy-Cramer, Jayashree and Farahani, Keyvan and Kirby, Justin and Burren, Yuliya and Porz, Nicole and Slotboom, Johannes and Wiest, Roland and others},
  journal={IEEE Transactions on Medical Imaging},
  volume={34},
  number={10},
  pages={1993--2024},
  year={2014},
  publisher={IEEE},
}

@inproceedings{Nguyen2025ViPETReportGen,
  title={Toward a Vision-Language Foundation Model for Medical Data: Multimodal Dataset and Benchmarks for {Vietnamese} {PET/CT} Report Generation},
  author={Nguyen, Tien and Nguyen, Dac and Nguyen, Trung Thanh and Thao Nguyen, Truong and Pham, Hieu and Barthelemy, Johan and Quan, Tran Minh and Nguyen, Quoc Viet Hung and Nguyen, Thanh Tam and Son, Mai and others},
  volume={38},
  year={2025},
  booktitle={Advances in Neural Information Processing Systems},
}

@article{Setio2017Luna16,
  title={Validation, comparison, and combination of algorithms for automatic detection of pulmonary nodules in computed tomography images: The {LUNA16} challenge},
  author={Setio, Arnaud Arindra Adiyoso and Traverso, Alberto and De Bel, Thomas and Berens, Moira SN and Van Den Bogaard, Cas and Cerello, Piergiorgio and Chen, Hao and Dou, Qi and Fantacci, Maria Evelina and Geurts, Bram and others},
  journal={Medical Image Analysis},
  volume={42},
  pages={1--13},
  year={2017},
  publisher={Elsevier},
}

@article{Wasserthal2023TotalSegmentator,
  title={{TotalSegmentator}: Robust Segmentation of 104 Anatomic Structures in {CT} Images},
  author={Wasserthal, Jakob and Breit, Hanns-Christian and Meyer, Manfred T and Pradella, Maurice and Hinck, Daniel and Sauter, Alexander W and Heye, Tobias and Boll, Daniel T and Cyriac, Joshy and Yang, Shan and others},
  journal={Radiology: Artificial Intelligence},
  volume={5},
  number={5},
  pages={e230024},
  year={2023},
  publisher={Radiological Society of North America},
}

@article{Yan2018DeepLesion,
  title={{DeepLesion}: automated mining of large-scale lesion annotations and universal lesion detection with deep learning},
  author={Yan, Ke and Wang, Xiaosong and Lu, Le and Summers, Ronald M},
  journal={Journal of Medical Imaging},
  volume={5},
  number={3},
  pages={036501--036501},
  year={2018},
  publisher={Society of Photo-Optical Instrumentation Engineers},
}

@article{Zhang2025RadGenomeChestCT,
  title={Development of a large-scale grounded vision language dataset for chest {CT} analysis},
  author={Zhang, Xiaoman and Wu, Chaoyi and Zhao, Ziheng and Lei, Jiayu and Tian, Weiwei and Zhang, Ya and Xie, Weidi and Wang, Yanfeng},
  journal={Scientific Data},
  volume={12},
  number={1},
  pages={1636},
  year={2025},
  publisher={Nature Publishing Group UK London},
}

@article{Tankel2026InformCT,
  title={{INFORM-CT}: INtegrating {LLMs} and {VLMs} FOR Incidental Findings Management in Abdominal {CT}},
  author={Tankel, Idan and Mazor, Nir and Brada, Rafi and LeBedis, Christina and Ben-Yosef, Guy},
  journal={arXiv preprint arXiv:2512.14732},
  year={2025}
}

@article{Antonelli2022MSD,
  title={The Medical Segmentation Decathlon},
  author={Antonelli, Michela and Reinke, Annika and Bakas, Spyridon and Farahani, Keyvan and Kopp-Schneider, Annette and Landman, Bennett A and Litjens, Geert and Menze, Bjoern and Ronneberger, Olaf and Summers, Ronald M and others},
  journal={Nature Communications},
  volume={13},
  number={1},
  pages={4128},
  year={2022},
  publisher={Nature Publishing Group UK London},
}

@article{Lei2024AutoRGBrain,
  title={{AutoRG-Brain}: Grounded report generation for brain {MRI}},
  author={Lei, Jiayu and Zhang, Xiaoman and Wu, Chaoyi and Dai, Lisong and Zhang, Ya and Zhang, Yanyong and Wang, Yanfeng and Xie, Weidi and Li, Yuehua},
  journal={arXiv preprint arXiv:2407.16684},
  year={2024}
}

@inproceedings{Lu2026GastricX,
  title={{Gastric-X}: A Multimodal Multi-Phase Benchmark Dataset for Advancing Vision-Language Models in Gastric Cancer Analysis},
  author={Li, Yuanzhe and Chen, Hao and Yin, Rui and Ba, Juyan and Zhang, Yu and Lu, Sheng},
  booktitle={Proceedings of the IEEE/CVF Conference on Computer Vision and Pattern Recognition},
  pages={2490--2501},
  year={2026}
}

@article{Baharoon2025ReXGroundingCT,
  title={{ReXGroundingCT}: A {3D} Chest {CT} Dataset for Segmentation of Findings from Free-Text Reports},
  author={Baharoon, Mohammed and Luo, Luyang and Moritz, Michael and Kumar, Abhinav and Kim, Sung Eun and Zhang, Xiaoman and Zhu, Miao and Alabbad, Mahmoud H and Alhazmi, Maha S and Mistry, Neel P and others},
  journal={NEJM AI},
  pages={AIdbp2501220},
  year={2026},
  publisher={Massachusetts Medical Society},
  volume={3},
  number={7},
}

@article{Trinh2026SpatialMed,
  title={Beyond Medical Diagnostics: How Medical Multimodal Large Language Models Think in Space},
  author={Trinh, Quoc-Huy and Ding, Xi and Liu, Yang and Qin, Zhenyue and Li, Xingjian and Durak, Gorkem and Aktas, Halil Ertugrul and Keles, Elif and Bagci, Ulas and Xu, Min},
  journal={arXiv preprint arXiv:2603.13800},
  year={2026}
}

@article{monon2026ctspatialvqa,
  title={Lost in Volume: The {CT-SpatialVQA} Benchmark for Evaluating Semantic-Spatial Understanding of {3D} Medical Vision-Language Models},
  author={Monon, Mashrafi and Rahman, Umaima and Hanif, Asif and Saeed, Numan and Yaqub, Mohammad},
  journal={arXiv preprint arXiv:2605.08787},
  year={2026}
}

@article{nguyen2026medstepbench,
  title={{Med-StepBench}: A Hierarchical Reasoning Framework for Evaluating Hallucinations in Medical Vision-Language Models},
  author={Nguyen, Minh Khoi and Le, Dai Lam and Jafari, Amir Reza and Nguyen, Tuan Dung and Son, Mai Hong and Thong, Mai Huy and Nguyen, Quang Huy and Nguyen, Thanh Trung and Farahbakhsh, Reza and Crespi, Noel and others},
  journal={arXiv preprint arXiv:2605.10002},
  year={2026}
}

@article{maksudov2026abra,
  title={{ABRA}: Agent Benchmark for Radiology Applications},
  author={Maksudov, Bulat and Kurenkov, Vladislav and Curran, Kathleen M and Mileo, Alessandra},
  journal={arXiv preprint arXiv:2605.11224},
  year={2026}
}

@article{liu2026oncologyvqa,
  title={Automated Report-Derived Oncology {VQA} Benchmark for Evaluating Vision-Language Models on {3D} Medical Imaging},
  author={Liu, Bo and Gu, Hanxue and Li, Xiangru and Zhu, Zheren and Ellison, Jacob and Wang, Kang and Lupo, Janine M and Yang, Yang and Lin, Hui},
  journal={arXiv preprint arXiv:2606.02809},
  year={2026}
}

@article{ashraf2026medcta,
  title={{MedCTA}: A Benchmark for Clinical Tool Agents},
  author={Ashraf, Tajamul and Jeong, Hyewon and Thoker, Fida Mohammad and Ghanem, Bernard},
  journal={arXiv preprint arXiv:2606.11702},
  year={2026}
}

@article{shi2026reportqa,
  title={{ReportQA}: {QA}-Based Radiology Report Evaluation},
  author={Shi, Yiming and Yang, Shaoshuai and Chen, Xi and Li, Haolin and Zhang, Hengyu and Jiang, Che and Wang, Kaiwen and Zhu, Xun and Xie, Dong and Wang, Fei and others},
  journal={arXiv preprint arXiv:2606.15037},
  year={2026}
}

@article{malik2026cortex,
  title={{CORTEX}: A Structured Reasoning Benchmark for Trustworthy {3D} Chest {CT} {MLLMs}},
  author={Malik, Hashmat Shadab and Hashmi, Anees Ur Rehman and Saeed, Numan and Naseer, Muzammal and Khan, Salman and Lippert, Christoph},
  journal={arXiv preprint arXiv:2606.27264},
  year={2026}
}

@article{liu2025medsam3,
  title={{MedSAM3}: Delving into Segment Anything with Medical Concepts},
  author={Liu, Anglin and Xue, Rundong and Cao, Xu R and Shen, Yifan and Lu, Yi and Li, Xiang and Chen, Qianqian and Chen, Jintai},
  journal={arXiv preprint arXiv:2511.19046},
  year={2025}
}

@article{jiang2026ibisagent,
  title={{IBISAgent}: Reinforcing Pixel-Level Visual Reasoning in {MLLMs} for Universal Biomedical Object Referring and Segmentation},
  author={Jiang, Yankai and Li, Qiaoru and Xu, Binlu and Sun, Haoran and Ding, Chao and Dong, Junting and Cai, Yuxiang and Zhang, Xuhong and Yin, Jianwei},
  journal={arXiv preprint arXiv:2601.03054},
  year={2026}
}

@article{acosta2022multimodal,
  title={Multimodal biomedical {AI}},
  author={Acosta, Juli{\'a}n N and Falcone, Guido J and Rajpurkar, Pranav and Topol, Eric J},
  journal={Nature Medicine},
  volume={28},
  number={9},
  pages={1773--1784},
  year={2022},
  publisher={Nature Publishing Group US New York},
}

@article{moor2023foundation,
  title={Foundation models for generalist medical artificial intelligence},
  author={Moor, Michael and Banerjee, Oishi and Abad, Zahra Shakeri Hossein and Krumholz, Harlan M and Leskovec, Jure and Topol, Eric J and Rajpurkar, Pranav},
  journal={Nature},
  volume={616},
  number={7956},
  pages={259--265},
  year={2023},
  publisher={Nature Publishing Group UK London},
}

@article{rao2025multimodal,
  title={Multimodal generative {AI} for medical image interpretation},
  author={Rao, Vishwanatha M and Hla, Michael and Moor, Michael and Adithan, Subathra and Kwak, Stephen and Topol, Eric J and Rajpurkar, Pranav},
  journal={Nature},
  volume={639},
  number={8056},
  pages={888--896},
  year={2025},
  publisher={Nature Publishing Group UK London},
}

@article{sounderajah2025stard,
  title={The {STARD-AI} reporting guideline for diagnostic accuracy studies using artificial intelligence},
  author={Sounderajah, Viknesh and Guni, Ahmad and Liu, Xiaoxuan and Collins, Gary S and Karthikesalingam, Alan and Markar, Sheraz R and Golub, Robert M and Denniston, Alastair K and Shetty, Shravya and Moher, David and others},
  journal={Nature Medicine},
  volume={31},
  number={10},
  pages={3283--3289},
  year={2025},
  publisher={Nature Publishing Group US New York},
}

@article{vasey2022decide,
  title={Reporting guideline for the early stage clinical evaluation of decision support systems driven by artificial intelligence: {DECIDE-AI}},
  author={Vasey, Baptiste and Nagendran, Myura and Campbell, Bruce and Clifton, David A and Collins, Gary S and Denaxas, Spiros and Denniston, Alastair K and Faes, Livia and Geerts, Bart and Ibrahim, Mudathir and others},
  journal={BMJ},
  volume={377},
  year={2022},
  publisher={British Medical Journal Publishing Group},
  pages={e070904},
}

@article{liu2020consort,
  title={Reporting guidelines for clinical trial reports for interventions involving artificial intelligence: the {CONSORT-AI} extension},
  author={Liu, Xiaoxuan and Rivera, Samantha Cruz and Moher, David and Calvert, Melanie J and Denniston, Alastair K and Ashrafian, Hutan and Beam, Andrew L and Chan, An-Wen and Collins, Gary S and Darzi, Ara and Deeks, Jonathan J and others},
  journal={The Lancet Digital Health},
  volume={2},
  number={10},
  pages={e537--e548},
  year={2020},
  publisher={Elsevier},
}

@article{vandesande2024warrant,
  title={To warrant clinical adoption {AI} models require a multi-faceted implementation evaluation},
  author={van de Sande, Davy and Chung, Eline Fung Fen and Oosterhoff, Jacobien and van Bommel, Jasper and Gommers, Diederik and van Genderen, Michel E},
  journal={npj Digital Medicine},
  volume={7},
  number={1},
  pages={58},
  year={2024},
  publisher={Nature Publishing Group UK London},
}

@article{norgeot2020minimum,
  title={Minimum information about clinical artificial intelligence modeling: the {MI-CLAIM} checklist},
  author={Norgeot, Beau and Quer, Giorgio and Beaulieu-Jones, Brett K and Torkamani, Ali and Dias, Raquel and Gianfrancesco, Milena and Arnaout, Rima and Kohane, Isaac S and Saria, Suchi and Topol, Eric and others},
  journal={Nature Medicine},
  volume={26},
  number={9},
  pages={1320--1324},
  year={2020},
  publisher={Nature Publishing Group US New York},
}

@article{tejani2024claim,
  title={Checklist for Artificial Intelligence in Medical Imaging ({CLAIM}): 2024 Update},
  author={Tejani, Ali S and Klontzas, Michail E and Gatti, Anthony A and Mongan, John T and Moy, Linda and Park, Seong Ho and Kahn Jr, Charles E and CLAIM 2024 Update Panel},
  journal={Radiology: Artificial Intelligence},
  volume={6},
  number={4},
  pages={e240300},
  year={2024},
  publisher={Radiological Society of North America},
}

@article{lekadir2025future,
  title={{FUTURE-AI}: international consensus guideline for trustworthy and deployable artificial intelligence in healthcare},
  author={Lekadir, Karim and Frangi, Alejandro F and Porras, Antonio R and Glocker, Ben and Cintas, Celia and Langlotz, Curtis P and Weicken, Eva and Asselbergs, Folkert W and Prior, Fred and Collins, Gary S and others},
  journal={BMJ},
  volume={388},
  year={2025},
  publisher={British Medical Journal Publishing Group},
  pages={e081554},
}

@article{finlayson2021clinician,
  title={The Clinician and Dataset Shift in Artificial Intelligence},
  author={Finlayson, Samuel G and Subbaswamy, Adarsh and Singh, Karandeep and Bowers, John and Kupke, Annabel and Zittrain, Jonathan and Kohane, Isaac S and Saria, Suchi},
  journal={New England Journal of Medicine},
  volume={385},
  number={3},
  pages={283--286},
  year={2021},
  publisher={Mass Medical Soc},
}

@article{obermeyer2019dissecting,
  title={Dissecting racial bias in an algorithm used to manage the health of populations},
  author={Obermeyer, Ziad and Powers, Brian and Vogeli, Christine and Mullainathan, Sendhil},
  journal={Science},
  volume={366},
  number={6464},
  pages={447--453},
  year={2019},
  publisher={American Association for the Advancement of Science},
}

@article{alber2025medical,
  title={Medical large language models are vulnerable to data-poisoning attacks},
  author={Alber, Daniel Alexander and Yang, Zihao and Alyakin, Anton and Yang, Eunice and Rai, Sumedha and Valliani, Aly A and Zhang, Jeff and Rosenbaum, Gabriel R and Amend-Thomas, Ashley K and Kurland, David B and others},
  journal={Nature Medicine},
  volume={31},
  number={2},
  pages={618--626},
  year={2025},
  publisher={Nature Publishing Group US New York},
}

@article{mahajan2025cognitive,
  title={Cognitive bias in clinical large language models},
  author={Mahajan, Arjun and Obermeyer, Ziad and Daneshjou, Roxana and Lester, Jenna and Powell, Dylan},
  journal={npj Digital Medicine},
  volume={8},
  number={1},
  pages={428},
  year={2025},
  publisher={Nature Publishing Group UK London},
}

@article{orlando2026medscribe,
  title={{MedScribe}: Clinically Grounded {CT} Reporting through Agentic Workflows},
  author={Orlando, Giuseppe A and Papotti, Paolo and Zuluaga, Maria A and Humbert, Olivier and Lorenzi, Marco},
  journal={arXiv preprint arXiv:2605.01779},
  year={2026}
}

@article{alim2026gaze,
  title={{GAZE}: Grounded Agentic Zero-shot Evaluation with Viewer-Level Tools and Literature Retrieval on Rare Brain {MRI}},
  author={Alim, Duaa and Alim, Mogtaba and Chalcroft, Liam},
  journal={arXiv preprint arXiv:2605.00876},
  year={2026}
}

@article{wind2026safety,
  title={Safety and accuracy follow different scaling laws in clinical large language models},
  author={Wind, Sebastian and Nguyen, Tri-Thien and Sopa, Jeta and Lotfinia, Mahshad and Bickelhaup, Sebastian and Uder, Michael and K{\"o}stler, Harald and Wellein, Gerhard and Nebelung, Sven and Truhn, Daniel and others},
  journal={arXiv preprint arXiv:2605.04039},
  year={2026}
}

@article{choi2026petct,
  title={End-to-End {PET/CT} Interpretation and Quantification with an {LLM}-Orchestrated {AI} Agent: A Real-World Pilot Study},
  author={Choi, Hongyoon and Bae, Sungwoo and Na, Kwon Joong},
  journal={Journal of Nuclear Medicine},
  year={2026},
  publisher={Society of Nuclear Medicine},
  pages={jnumed.126.272362},
}

@article{moukheiber2026sgmri,
  title={Beyond a Single Frame: Multi-Frame Spatially Grounded Reasoning Across Volumetric {MRI}},
  author={Moukheiber, Lama and Yeung, Caleb M and Xue, Haotian and Helbling, Alec and Zhao, Zelin and Chen, Yongxin},
  journal={arXiv preprint arXiv:2604.15808},
  year={2026}
}

@article{zuo2026artifact,
  title={An Artifact-based Agent Framework for Adaptive and Reproducible Medical Image Processing},
  author={Zuo, Lianrui and Liu, Yihao and Rudravaram, Gaurav and Ramadass, Karthik and Krishnan, Aravind R and Phillips, Michael D and Bodien, Yelena G and Patel, Mayur B and Trujillo, Paula and Martinez, Yency Forero and others},
  journal={arXiv preprint arXiv:2604.21936},
  year={2026}
}

@article{shen2026neuroclaw,
  title={Neuroclaw technical report},
  author={Wang, Cheng and He, Zhibin and Peng, Zhihao and Liu, Shengyuan and Hu, Yufan and Carl, Yang and Lifang, He and Sun, Lichao and Li, Xiang and Yuan, Yixuan},
  journal={arXiv preprint arXiv:2604.24696},
  year={2026}
}

@article{wang2026martp,
  title={{MARTP}: a multi-agent simulation framework for automated radiation therapy planning based on {LLMs}},
  author={Wang, Dongzhao and Hu, Zeyun and Li, Yang and Xu, Dachuan and Yang, Ruijie and Zhuge, Changjing},
  journal={Physics in Medicine \& Biology},
  year={2026},
  volume={71},
  number={12},
  pages={125037},
}

@article{nusrat2026sage,
  title={Automated stereotactic radiosurgery planning using a human-in-the-loop reasoning large language model agent},
  author={Nusrat, Humza and Francisco, Luke and Luo, Bing and Bagher-Ebadian, Hassan and Kim, Joshua and Chin-Snyder, Karen and Siddiqui, Salim and Shah, Mira and Mellon, Eric and Ghassemi, Mohammad and others},
  journal={arXiv preprint arXiv:2512.20586},
  year={2025}
}

@article{Yang2023MedMNIST,
  title={{MedMNIST} v2 - A large-scale lightweight benchmark for {2D} and {3D} biomedical image classification},
  author={Yang, Jiancheng and Shi, Rui and Wei, Donglai and Liu, Zequan and Zhao, Lin and Ke, Bilian and Pfister, Hanspeter and Ni, Bingbing},
  journal={Scientific Data},
  volume={10},
  number={1},
  pages={41},
  year={2023},
  publisher={Nature Publishing Group UK London},
}

@article{Akinci2025TotalSegmentatorMRI,
  title={{TotalSegmentator} {MRI}: robust sequence-independent segmentation of multiple anatomic structures in {MRI}},
  author={D'Antonoli, Tugba Akinci and Berger, Lucas K and Indrakanti, Ashraya K and Vishwanathan, Nathan and Wei{\ss}, Jakob and Jung, Matthias and Berkarda, Zeynep and Rau, Alexander and Reisert, Marco and K{\"u}stner, Thomas and others},
  journal={arXiv preprint arXiv:2405.19492},
  year={2024}
}

@article{Wang2025Triad,
  title={Triad: Vision foundation model for {3D} magnetic resonance imaging},
  author={Wang, Shansong and Safari, Mojtaba and Li, Qiang and Chang, Chih-Wei and Qiu, Richard LJ and Roper, Justin and Yu, David S and Yang, Xiaofeng},
  journal={Research Square},
  pages={rs--3},
  year={2025}
}

@inproceedings{Luo2024RAOS,
  title={Rethinking Abdominal Organ Segmentation ({RAOS}) in the Clinical Scenario: A Robustness Evaluation Benchmark with Challenging Cases},
  author={Luo, Xiangde and Li, Zihan and Zhang, Shaoting and Liao, Wenjun and Wang, Guotai},
  booktitle={International Conference on Medical Image Computing and Computer-Assisted Intervention},
  pages={531--541},
  year={2024},
  organization={Springer},
}

@article{Podobnik2024HaNSeg,
  title={{HaN-Seg}: The head and neck organ-at-risk {CT} and {MR} segmentation challenge},
  author={Podobnik, Ga{\v{s}}per and Ibragimov, Bulat and Tappeiner, Elias and Lee, Chanwoong and Kim, Jin Sung and Mesbah, Zacharia and Modzelewski, Romain and Ma, Yihao and Yang, Fan and Rudecki, Miko{\l}aj and others},
  journal={Radiotherapy and Oncology},
  volume={198},
  pages={110410},
  year={2024},
  publisher={Elsevier},
}

@article{Cardenas2020AAPMRTMAC,
  title={Head and neck cancer patient images for determining auto-segmentation accuracy in {T2}-weighted magnetic resonance imaging through expert manual segmentations},
  author={Cardenas, Carlos E and Mohamed, Abdallah SR and Yang, Jinzhong and Gooding, Mark and Veeraraghavan, Harini and Kalpathy-Cramer, Jayashree and Ng, Sweet Ping and Ding, Yao and Wang, Jihong and Lai, Stephen Y and others},
  journal={Medical Physics},
  volume={47},
  number={5},
  pages={2317--2322},
  year={2020},
  publisher={Wiley Online Library},
}

@article{hu2026simulating,
  title={Simulating the Real World: A Unified Survey of Multimodal Generative Models},
  author={Hu, Yuqi and Wang, Longguang and Liu, Xian and Chen, Ling-Hao and Guo, Yuwei and Shi, Yukai and Liu, Ce and Rao, Anyi and Wang, Zeyu and Xiong, Hui},
  journal={IEEE Transactions on Pattern Analysis and Machine Intelligence},
  year={2026},
  publisher={IEEE},
  pages={1--20},
}

@article{bluethgen2025agenticradiology,
  title={Agentic Systems in Radiology: Design, Applications, Evaluation, and Challenges},
  author={Bluethgen, Christian and Van Veen, Dave and Truhn, Daniel and Kather, Jakob Nikolas and Moor, Michael and Polacin, Malgorzata and Chaudhari, Akshay and Frauenfelder, Thomas and Langlotz, Curtis P. and Krauthammer, Michael and Nooralahzadeh, Farhad},
  journal={arXiv preprint arXiv:2510.09404},
  year={2025}
}

@article{lievin2026towards,
  title={Towards conversational {AI} for disease management},
  author={Palepu, Anil and Li{\'e}vin, Valentin and Weng, Wei-Hung and Saab, Khaled and Stutz, David and Cheng, Yong and Kulkarni, Kavita and Mahdavi, S Sara and Barral, Jo{\"e}lle and Webster, Dale R and others},
  journal={arXiv preprint arXiv:2503.06074},
  year={2025}
}

@article{qwen2026agentworld,
  title={{Qwen-AgentWorld}: Language World Models for General Agents},
  author={Zuo, Yuxin and Xiao, Zikai and Sheng, Li and Huang, Fei and Tu, Jianhong and Liu, Yuxuan and Tang, Tianyi and Hu, Xiaomeng and Su, Yang and Lan, Qingfeng and others},
  journal={arXiv preprint arXiv:2606.24597},
  year={2026}
}

@article{sun2026experiencemakesskillfulenabling,
  title={Experience Makes Skillful: Enabling Generalizable Medical Agent Reasoning via Self-Evolving Skill Memory},
  author={Sun, Haoran and Li, Wenjie and Zhang, Yujie and Lin, Zekai and Zhang, Fanrui and Chen, Kaitao and He, Xingqi and Li, Yichen and Liu, Mianxin and Liu, Lei and others},
  journal={arXiv preprint arXiv:2606.09365},
  year={2026}
}

@article{jiangmedvr,
  title={{MedVR}: Annotation-free medical visual reasoning via agentic reinforcement learning},
  author={Jiang, Zheng and Guo, Heng and Fang, Chengyu and Xiao, Changchen and Hu, Xinyang and Sun, Lifeng and Xu, Minfeng},
  journal={arXiv preprint arXiv:2604.08203},
  year={2026}
}

@article{ridnik2024code,
  title={Code generation with {AlphaCodium}: From prompt engineering to flow engineering},
  author={Ridnik, Tal and Kredo, Dedy and Friedman, Itamar},
  journal={arXiv preprint arXiv:2401.08500},
  year={2024}
}

@article{dong2022shape,
  title={Shape Tracking and Feedback Control of Cardiac Catheter Using {MRI}-Guided Robotic Platform--Validation With Pulmonary Vein Isolation Simulator in {MRI}},
  author={Dong, Ziyang and Wang, Xiaomei and Fang, Ge and He, Zhuoliang and Ho, Justin Di-Lang and Cheung, Chim-Lee and Tang, Wai Lun and Xie, Xiaochen and Liang, Liyuan and Chang, Hing-Chiu and others},
  journal={IEEE Transactions on Robotics},
  volume={38},
  number={5},
  pages={2781--2798},
  year={2022},
  publisher={IEEE},
}

@article{granados2026evolving,
  title={Evolving surgical teams in the age of artificial intelligence and robotics},
  author={Granados, Alejandro and Khanna, Raghav and Fischer, Nikola and Raison, Nicholas and Ciabattini, Margarita and Robertshaw, Harry and Boels, Maxence and Malik, Mohsan and Granados, Veronica and Vercauteren, Tom and others},
  journal={Frontiers in Science},
  volume={4},
  pages={1783803},
  year={2026},
  publisher={Frontiers Media SA},
}

@article{chen2026ai,
  title={{AI}-Driven Revolution of Medical Robotics Across Surgical Innovation, Rehabilitation Intelligence, and Multimodal Healthcare Delivery},
  author={Chen, Fanxuan and Chen, Haoman and Yu, Tao and Wang, Ruoyun and Wang, Yi and Zhang, Xian and Li, Jiachen and Liu, Kaishuo and Hai, Darong and Bao, Xueying and others},
  journal={MedComm},
  volume={7},
  number={3},
  pages={e70597},
  year={2026},
  publisher={Wiley Online Library},
}

@article{wang2026can,
  title={How can reasoning capability empower the {AI} copilot robot in endoscopic surgery},
  author={Wang, Guankun and Bai, Long and Ren, Hongliang},
  journal={npj Digital Medicine},
  volume={9},
  number={1},
  pages={447},
  year={2026},
  publisher={Nature Publishing Group},
}

@article{zhang2026seer,
  title={Skill-Evolving Grounded Reasoning for Free-Text Promptable {3D} Medical Image Segmentation},
  author={Zhang, Tongrui and Wang, Chenhui and Li, Yongming and Chen, Zhihao and Zhan, Xufeng and Shan, Hongming},
  journal={arXiv preprint arXiv:2603.08215},
  year={2026}
}

@article{ferber2026towards,
  title={Towards autonomous medical artificial intelligence agents},
  author={Ferber, Dyke and Hilgers, Lars and H{\"o}per, Christiane and Kinny-K{\"o}ster, Benedict and Eckardt, Jan-Niklas and Egger-Heidrich, Katharina and Bill, Marius and Schneider, Martin MK and Clusmann, Jan and Kadric, Lejla and others},
  journal={Nature},
  pages={1--10},
  year={2026},
}

@article{saab2026advancing,
  title={Advancing conversational diagnostic {AI} with multimodal reasoning},
  author={Saab, Khaled and Park, Chunjong and Strother, Tim and Freyberg, Jan and Barrett, David GT and Cheng, Yong and Weng, Wei-Hung and Stutz, David and Tomasev, Nenad and Palepu, Anil and others},
  journal={Nature Medicine},
  pages={1--11},
  year={2026},
  publisher={Nature Publishing Group US New York},
  volume={32},
  number={5},
}

@article{woznitza2026ai,
  title={{AI}-based chest X-ray prioritization in the lung cancer diagnostic pathway: the {LungIMPACT} randomized controlled trial},
  author={Woznitza, Nick and Smith, Lesley and Rawlinson, Janette and Au-Yong, Iain and George, Bindu and Djearaman, Madava G and Nair, Arjun and Lee, Richard W and Navani, Neal and Ndwandwe, Siyabonga and others},
  journal={Nature Medicine},
  pages={1--8},
  year={2026},
  publisher={Nature Publishing Group US New York},
  volume={32},
  number={5},
}

@article{ding2025clarity,
  title={{CLARITY}: Medical World Model for Guiding Treatment Decisions by Modeling Context-Aware Disease Trajectories in Latent Space},
  author={Ding, Tianxingjian and Zou, Yuanhao and Chen, Chen and Shah, Mubarak and Tian, Yu},
  journal={arXiv preprint arXiv:2512.08029},
  year={2025}
}

@inproceedings{yang2025medical,
  title={Medical World Model},
  author={Yang, Yijun and Wang, Zhao-Yang and Liu, Qiuping and Sun, Shuwen and Wang, Kang and Chellappa, Rama and Zhou, Zongwei and Yuille, Alan and Zhu, Lei and Zhang, Yu-Dong and others},
  booktitle={Proceedings of the IEEE/CVF International Conference on Computer Vision},
  pages={8319--8329},
  year={2025},
}

@article{liu2026medical,
  title={Medical world models: representing medical states, modelling clinical dynamics and guiding intervention policies},
  author={Liu, Ke and Li, Mengxuan and Bao, Yanyi and Zhang, Tianyun and Chu, Chong and Bu, Jiajun and Wang, Haishuai},
  journal={arXiv preprint arXiv:2606.16721},
  year={2026}
}

@article{wang2026brain,
  title={{Brain-WM}: Brain Glioblastoma World Model},
  author={Wang, Chenhui and Zheng, Boyun and Bao, Liuxin and Peng, Zhihao and Woo, Peter YM and Shan, Hongming and Yuan, Yixuan},
  journal={arXiv preprint arXiv:2603.07562},
  year={2026}
}

@article{gatopoulos2026coralbay,
  title={{CoralBay}: A Self-Supervised {CT} Foundation Model},
  author={Gatopoulos, Ioannis and K{\"a}nzig, Nicolas and Ot{\'a}lora, Sebastian and Tang, Fei},
  journal={arXiv preprint arXiv:2606.03888},
  year={2026}
}

@article{tao2026renalclip,
  title={A disease-centric vision-language foundation model for precision oncology in kidney cancer},
  author={Tao, Yuhui and Zhao, Zhongwei and Wang, Zilong and Luo, Xufang and Chen, Feng and Wang, Kang and Wu, Chuanfu and Zhang, Xue and Zhang, Shaoting and Yao, Jiaxi and others},
  journal={Nature Communications},
  year={2026},
  publisher={Nature Publishing Group},
}

@article{shi2026diseasecentric,
  title={Disease-Centric Vision-Language Pretraining with Hybrid Visual Encoding for {3D} Computed Tomography},
  author={Shi, Bowen and Cao, Weiwei and Yuan, Ruifeng and Chang, Wanxing and Dai, Wenrui and Xiong, Hongkai and Zhang, Ling and Zhang, Jianpeng},
  journal={arXiv preprint arXiv:2606.25546},
  year={2026}
}

@article{wang2026asap,
  title={{ASAP}: Advancing Medical Volumetric Representation Learning with Anatomy-aware Semantically-adaptive Pre-training},
  author={Wang, Rongsheng and Tang, Fenghe and Jiang, Zihang and Li, Yingtai and Zhang, Xu and Lai, Haoran and Ma, Wenxin and Wei, Wei and He, Zhiyang and Tao, Xiaodong and others},
  journal={arXiv preprint arXiv:2606.00602},
  year={2026}
}

@article{park2026glint,
  title={{GLINT}: Sparsely Gated Vision-Language Alignment for Fine-Grained Radiology Representations},
  author={Park, Jonggwon and Lee, Seongeun and Park, Junhyun and Yun, Hannah and Kim, Hyunwoong and Jeong, Sohyun and Kang, Hyewon and Yoon, Byungmu and Choi, Kyoyun},
  journal={arXiv preprint arXiv:2606.03180},
  year={2026}
}

@article{khlaut2026jolia,
  title={Jolia: Concept-Level Vision-Language Alignment for {3D} {CT} Contrastive Learning},
  author={Khlaut, Julien and Corbi{\`e}re, Charles and Callard, Baptiste and Prat, Amaury and Butsanets, Leo and Saporta, Antoine and Danielou, Th{\'e}o and Machado, Leo and Floch, Korentin Le and Boeken, Tom and others},
  journal={arXiv preprint arXiv:2606.24570},
  year={2026}
}

@article{jiang2026gleve,
  title={{GLeVE}: Graph-Guided Lesion Grounding with Proposal Verification in {3D} {CT}},
  author={Jiang, Shuo and Hong, Yuhao and Jiang, Chunbo and Chen, Weihong and Chen, Huangwei and Zhu, Shenghao and Wu, Beining and Liu, Mingxuan and Zhu, Zhu and Qin, Feiwei and others},
  journal={arXiv preprint arXiv:2605.22619},
  year={2026}
}

@article{li2026mri2rep,
  title={{MRI2Rep}: Autoregressive Structured Report Generation for {3D} Liver {MRI}},
  author={Li, Xinran and Yang, Junlin and Shewarega, Annabella and Zhou, Zongwei and Chapiro, Julius and Duncan, James S and Staib, Lawrence H},
  journal={arXiv preprint arXiv:2606.25279},
  year={2026}
}

@article{chen2026unireason,
  title={{UniReason-Med}: A Shared Grounded Reasoning Interface for {2D}-to-{3D} Transfer in Medical {VQA}},
  author={Chen, Mengzhuo and Shu, Yan and Liu, Chi and Piao, Hongming and Wang, Xidong and Li, Derek and Dai, Bryan},
  journal={arXiv preprint arXiv:2606.11740},
  year={2026}
}

@article{sharma2026rad3dprefix,
  title={Revisiting {LLM} Adaptation for {3D} {CT} Report Generation: A Study of Scaling and Diagnostic Priors},
  author={Sharma, Vanshali and Bejar, Andrea M and Aktas, Halil Ertugrul and Trinh, Quoc-Huy and Jha, Debesh and Durak, Gorkem and Bagci, Ulas},
  journal={arXiv preprint arXiv:2606.17213},
  year={2026}
}

@article{mayelasserre2026templatecollapse,
  title={Generating Reports or Repeating Templates? Measuring and Mitigating Template Collapse in {3D} {CT} Report Generation},
  author={Maye-Lasserre, Tom and Li, Yitong and Jian, Bailiang and Ghahremani, Morteza and Wiestler, Benedikt and Wachinger, Christian},
  journal={arXiv preprint arXiv:2605.30984},
  year={2026}
}

@article{lin2026tifgrpo,
  title={Regulating Anatomy-Aware Rewards via Trajectory-Integral Feedback for Volumetric Computed Tomography Analysis},
  author={Lin, Tianwei and Qiu, Zhongwei and Cao, Jie and Liu, Jiang and Yan, Wenjie and Zhang, Bo and Zhong, Yu and Zhang, Wenqiao and Xia, Yingda and Zhang, Ling},
  journal={arXiv preprint arXiv:2605.20277},
  year={2026}
}

@article{li2026emrl,
  title={{E-MRL}: Cross-view Aligned Evidence-driven Multimodal Reinforcement Learning for Reliable {3D} Tumor Analysis},
  author={Li, Sijing and Qiu, Zhongwei and Wang, Zhuoya and Yun, Boxiang and Yi, Zhenyu and Xu, Jianwei and Zhang, Wenqiao and Xia, Yingda and Zhang, Ling},
  journal={arXiv preprint arXiv:2606.23888},
  year={2026}
}

@article{hu2025landscape,
  title={The Landscape of Medical Agents: A Survey},
  author={Hu, Xiaobin and Qian, Yunhang and Yu, Jiaquan and Liu, Jingjing and Tang, Peng and Ji, Xiaozhong and Xu, Chengming and Liu, Jiawei and Yan, Xiaoxiao and Yu, Xinlei and others},
  journal={Authorea Preprints},
  year={2025},
  publisher={Authorea}
}

@article{chen2019med3d,
  title={{Med3D}: Transfer learning for {3D} medical image analysis},
  author={Chen, Sihong and Ma, Kai and Zheng, Yefeng},
  journal={arXiv preprint arXiv:1904.00625},
  year={2019}
}

@inproceedings{zhuang2019self,
  title={Self-supervised Feature Learning for {3D} Medical Images by Playing a Rubik's Cube},
  author={Zhuang, Xinrui and Li, Yuexiang and Hu, Yifan and Ma, Kai and Yang, Yujiu and Zheng, Yefeng},
  booktitle={International Conference on Medical Image Computing and Computer-Assisted Intervention},
  pages={420--428},
  year={2019},
  organization={Springer},
}

@article{dosovitskiy2020image,
  title={An image is worth 16x16 words: Transformers for image recognition at scale},
  author={Dosovitskiy, Alexey and Beyer, Lucas and Kolesnikov, Alexander and Weissenborn, Dirk and Zhai, Xiaohua and Unterthiner, Thomas and Dehghani, Mostafa and Minderer, Matthias and Heigold, Georg and Gelly, Sylvain and others},
  journal={arXiv preprint arXiv:2010.11929},
  year={2020}
}

@article{eisenhauer2009new,
  title={New response evaluation criteria in solid tumours: Revised {RECIST} guideline (version 1.1)},
  author={Eisenhauer, Elizabeth A and Therasse, Patrick and Bogaerts, Jan and Schwartz, Lawrence H and Sargent, Danielle and Ford, Robert and Dancey, Janet and Arbuck, Stephen and Gwyther, Steve and Mooney, Margaret and others},
  journal={European Journal of Cancer},
  volume={45},
  number={2},
  pages={228--247},
  year={2009},
  publisher={Elsevier},
}

@article{wahl2009recist,
  title={From {RECIST} to {PERCIST}: Evolving Considerations for {PET} Response Criteria in Solid Tumors},
  author={Wahl, Richard L and Jacene, Heather and Kasamon, Yvette and Lodge, Martin A},
  journal={Journal of Nuclear Medicine},
  volume={50},
  number={Suppl 1},
  pages={122S--150S},
  year={2009},
  publisher={Society of Nuclear Medicine},
}

@article{wen2010updated,
  title={Updated Response Assessment Criteria for High-Grade Gliomas: Response Assessment in Neuro-Oncology Working Group},
  author={Wen, Patrick Y and Macdonald, David R and Reardon, David A and Cloughesy, Timothy F and Sorensen, A Gregory and Galanis, Evanthia and DeGroot, John and Wick, Wolfgang and Gilbert, Mark R and Lassman, Andrew B and others},
  journal={Journal of Clinical Oncology},
  volume={28},
  number={11},
  pages={1963--1972},
  year={2010},
  publisher={American Society of Clinical Oncology},
}

@article{liu2026towards,
  title={Towards Autonomous and Auditable Medical Imaging Model Development},
  author={Liu, Shengyuan and Jiang, Jia-Xuan and Zheng, Boyun and Wang, Cheng and Wang, Zipei and Pan, Wentao and Wu, Hongtao and Peng, Houwen and Gu, Yu and Sun, Lichao and others},
  journal={arXiv preprint arXiv:2607.10522},
  year={2026}
}

@article{fayaz2026implementation,
  title={What Is Implementation Science: And Why It Matters for Bridging the Artificial Intelligence Innovation-to-Application Gap in Medical Imaging},
  author={Fayaz-Bakhsh, Ahmad and Tania, Janice and Lutfi, Syaheerah Lebai and Jha, Abhinav K and Rahmim, Arman},
  journal={{PET Clinics}},
  volume={21},
  number={1},
  pages={1--16},
  year={2026},
  publisher={Elsevier}
}

@article{abdollahi2026computational,
  title={Computational Oncology and the Augmented Oncologist: How Implementation-Ready {AI} and Digital Twins Will Transform Education, Research, and Practice in Precision Oncology—Insights from Theranostics},
  author={Abdollahi, Hamid and Fayaz-Bakhsh, Ahmad and Klyuzhin, Ivan and Soltani, Madjid and Saboury, Babak and Rahmim, Arman},
  journal={Frontiers in Biomedical Technologies},
  year={2026},
  note={In press}
}

@article{liu2026radsight,
  title={RadSight: Towards Perceptually Reliable Multimodal Radiology Image Understanding},
  author={Liu, Jianqin and Cao, Weiwei and Chang, Wanxing and Yuan, Ruifeng and Shi, Bowen and Zheng, Zhilin and Zhang, Xianjie and Zhang, Ling and Wang, Peng and Zhang, Jianpeng},
  journal={arXiv preprint arXiv:2607.22293},
  year={2026}
}

@article{yuan2026clinfusion,
  title={ClinFusion: A Vision-Centric Multimodal LLM System for Holistic Medical Understanding},
  author={Yuan, Hangjie and Qian, Yichen and Tang, Zhiwei and Xu, Xianzhe and Wu, Lirong and Yang, Sicheng and Wang, Jinwang and Wang, Pengju and Zeng, Zhitao and Han, Yizeng and others},
  journal={arXiv preprint arXiv:2607.24743},
  year={2026}
}

@article{zhang2026vision,
  title={A Vision-language Framework for Comparative Reasoning in Radiology},
  author={Zhang, Tengfei and Zhao, Ziheng and Zhang, Xiaoman and Dai, Lisong and Qiu, Pengcheng and Zhang, Ya and Wang, Yanfeng and Xie, Weidi},
  journal={arXiv preprint arXiv:2606.06407},
  year={2026}
}

@article{dong2026policy,
  title={Policy-Driven CT-Agent: Modeling Phase-Aware Diagnostic Control for Clinically Consistent CT Reasoning},
  author={Dong, Yanmeng and Li, Han and Li, Yujia and Liu, Jingsong and Ma, Xun and Hu, Yanzhu and Xu, Zhengyang and Li, Zhicheng and Navab, Nassir and Zhou, Shaohua Kevin},
  journal={arXiv preprint arXiv:2607.10748},
  year={2026}
}

@article{zhu2020rubik,
  title={{Rubik’s Cube}+: A self-supervised feature learning framework for {3d} medical image analysis},
  author={Zhu, Jiuwen and Li, Yuexiang and Hu, Yifan and Ma, Kai and Zhou, S Kevin and Zheng, Yefeng},
  journal={Medical image analysis},
  volume={64},
  pages={101746},
  year={2020},
  publisher={Elsevier}
}

\end{document}